\documentclass{article}

\PassOptionsToPackage{numbers, compress,sort}{natbib}

\usepackage[final]{neurips_2021}
\usepackage{microtype}
\usepackage{graphicx}
\usepackage{caption}
\usepackage{subcaption}

\usepackage{booktabs} 

\usepackage[utf8]{inputenc} 
\usepackage[T1]{fontenc}    
\usepackage{hyperref}       
\usepackage{url}            
\usepackage{booktabs}       
\usepackage{amsfonts}       
\usepackage{nicefrac}       
\usepackage{microtype}      
\usepackage{xcolor}         
\usepackage{amsthm}
\usepackage{mathrsfs}
\usepackage{amsmath}
\usepackage{amssymb}
\usepackage{mathtools}
\usepackage{soul}
\usepackage[ruled,longend, algo2e]{algorithm2e}
\usepackage{algorithm}
\usepackage{algorithmic}
\usepackage{hhline}
\usepackage{multirow}
\usepackage{stmaryrd}
\usepackage{enumitem,pifont}
\usepackage{etoolbox}
\usepackage{wrapfig}

\newcommand{\E}{\mathbb{E}}
\newcommand{\bas}[1]{\begin{align*}#1\end{align*}}
\newcommand{\phiv}{\boldsymbol{\phi}}
\newcommand{\ba}[1]{\begin{align}#1\end{align}}

\newcommand{\cdotv}{\boldsymbol{\cdot}}

\usepackage{stackengine}

\makeatletter
\newcommand{\distas}[1]{\mathbin{\overset{#1}{\kern\z@\sim}}}%

\newcommand{\beqs}{\vspace{0mm}\begin{eqnarray}}
\newcommand{\eeqs}{\vspace{0mm}\end{eqnarray}}
\newcommand{\barr}{\begin{array}}
\newcommand{\earr}{\end{array}}

\newcommand{\hv}[0]{{\boldsymbol{h}} }

\newcommand{\xv}{\boldsymbol{x}}
\newcommand{\yv}{\boldsymbol{y}}

\newcommand{\epsilonv}{\boldsymbol{\epsilon}}

\newcommand{\etav}[0]{\boldsymbol{\eta}}

\newcommand{\thetav}{\boldsymbol{\theta}}

\newcommand{\given}{\,|\,}
\newtheorem{theorem}{Theorem}
\newtheorem{lemma}[theorem]{Lemma}

\definecolor{codegreen}{rgb}{0,0.3,0.6}
\definecolor{codegray}{rgb}{0.5,0.5,0.5}
\definecolor{codepurple}{rgb}{0.58,0,0.82}
\definecolor{backcolour}{rgb}{0.95,0.95,0.92}
\definecolor{orange}{rgb}{1,0.5,0}
\definecolor{mydarkblue}{rgb}{0,0.08,0.45}

\newcommand{\RN}[1]{%
	\textup{\lowercase\expandafter{\it \romannumeral#1}}%
}

\usepackage{listings}
\lstdefinestyle{mystyle}{
    basicstyle=\tiny,
    commentstyle=\color{codegreen},
    keywordstyle=\color{magenta},
    numberstyle=\tiny\color{codegray},
    stringstyle=\color{codepurple},
    basicstyle=\fontsize{8.5}{9}\selectfont\ttfamily,
    breakatwhitespace=false,         
    breaklines=true,                 
    captionpos=b,                    
    keepspaces=true,                 
    numbers=none,
    citecolor=mydarkblue,
    numbersep=5pt,                  
    showspaces=false,                
    showstringspaces=false,
}

\usepackage[multiple]{footmisc}

\newcommand{\mz}[1]{{\color{red}{[mz:#1]}}}
\newcommand{\hz}[1]{{\color{cyan}{[hj:#1]}}}

\title{Exploiting Chain Rule and Bayes' Theorem to \\Compare Probability Distributions}

\author{%
  Huangjie Zheng\\
  Department of Statistics \& Data Science\\
  The University of Texas at Austin\\
Austin, TX 78712 \\
  \texttt{huangjie.zheng@utexas.edu} \\
   \And
  Mingyuan Zhou \\
  McCombs School of Business\\
  The University of Texas at Austin \\
  Austin, TX 78712 \\
   \texttt{mingyuan.zhou@mccombs.utexas.edu} \\
}

\begin{document}

\maketitle

\begin{abstract}
To measure the difference between two probability distributions, referred to as the source and target, respectively, we exploit both the chain rule and Bayes' theorem to construct conditional transport (CT), which is constituted by both a forward component and a backward one. The forward CT is the expected cost of moving a source data point to a target one, with their joint distribution defined by the product of the source probability density function (PDF) and a source-dependent conditional distribution, which is related to the target PDF via Bayes' theorem. The backward CT is defined by reversing the direction. The CT cost can be approximated by replacing the source and target PDFs with their discrete empirical distributions supported on mini-batches, making it amenable to implicit distributions and stochastic gradient descent-based optimization. When applied to train a generative model, CT is shown to strike a good balance between mode-covering and mode-seeking behaviors and strongly resist mode collapse. On a wide variety of benchmark datasets for generative modeling, substituting the default statistical distance of an existing generative adversarial network with CT is shown to consistently improve the performance. 
PyTorch code is provided.

\end{abstract}

\section{Introduction}

Measuring the difference between two probability distributions is a fundamental problem in statistics and machine learning \citep{cover1999elements,bishop2006pattern,murphy2012machine}. A variety of statistical distances, such as the Kullback--Leibler (KL) divergence \citep{kullback1951information}, Jensen--Shannon (JS) divergence \citep{lin1991divergence}, maximum mean discrepancy (MMD) \citep{gretton2006kernel}, 
and Wasserstein distance \citep{kantorovich2006translocation}, have been proposed %
to quantify the difference. They have been widely used for generative modeling with different mode covering/seeking behaviors 
\citep{kingma2013auto,goodfellow2014generative,binkowski2018demystifying,arjovsky2017wasserstein,genevay2018learning, balaji2019entropic}.
The KL divergence, 
directly related to both maximum likelihood estimation and variational inference \citep{wainwright2008graphical,hoffman2013stochastic,blei2017variational}, 
requires the two probability distributions to share the same support and is often inapplicable if either is an implicit distribution whose probability density function (PDF) is unknown \citep{mohamed2016learning,huszar2017variational,tran2017hierarchical,yin2018semi}. 
Variational auto-encoders (VAEs) \citep{kingma2013auto}, the KL divergence based deep generative models, are stable to train, but often exhibit mode-covering behaviors in its generated data, producing  
blurred images.
The JS divergence is directly related to the min-max loss of a generative adversarial net (GAN) when the discriminator is optimal \citep{goodfellow2014generative}, while the Wasserstein-1 distance is directly related to the min-max loss of a Wasserstein GAN \citep{arjovsky2017wasserstein}, whose critic is optimized under the 1-Lipschitz constraint. However, it is difficult to maintain a good balance between the updates of the generator and discriminator/critic, making (Wasserstein) GANs notoriously brittle to train. MMD \citep{gretton2006kernel} is an RKHS-based statistical distance behind MMD-GANs \citep{li2015generative,li2017mmd,binkowski2018demystifying}, which have also shown promising results in generative modeling when trained with a min-max loss.
Different from VAEs, these GAN-based models often exhibit mode dropping and face the danger of mode collapse if not well tuned during the training. 

 
In this paper, we introduce conditional transport (CT) as a new method to quantify the difference between two probability distributions, which will be referred to as
 the source distribution $p_{X}(\xv)$ and target distribution $p_Y(\yv)$, respectively.
The construction of CT is motivated by the following observation: the difference between $p_{X}(\xv)$ and $p_Y(\yv)$ can be reflected by  the expected  difference of two dependent random variables $\xv$ and $\yv$, whose joint distribution $\pi(\xv,\yv)$ is constrained by both $p_X(\xv)$ and $p_{Y}(\yv)$ in a certain way. Denoting $c(\xv,\yv)\ge 0$ as a cost function to measure the difference between points $\xv$ and $\yv$, such as $c(\xv,\yv)=\|\xv-\yv\|_2^2$,  the expected difference is expressed as $\E_{\pi(\xv,\yv)}[c(\xv,\yv)]$. A basic way to constrain $\pi(\xv,\yv)$ with both $p_X(\xv)$ and $p_Y(\yv)$ is to let $\pi(\xv,\yv)=p_X(\xv)p_Y(\yv)$, which means drawing $\xv$ and $\yv$ independently from $p_X(\xv)$ and $p_{Y}(\yv)$, respectively; this expected difference $\E_{p_X(\xv)p_{Y}(\yv)}[c(\xv,\yv)]$ is closely related to the energy distance~\cite{bellemare2017cramer}. Another constraining method is to require both $\int \pi(\xv,\yv) d\yv =p_{X}(\xv)$ and $\int \pi(\xv,\yv) d\xv =p_{Y}(\yv)$, under which ${\min_{\pi}}\{\E_{\pi(\xv,\yv)}[c(\xv,\yv)]\}$ becomes the Wasserstein distance \citep{kantorovich2006translocation,
villani2008optimal,santambrogio2015optimal,COTFNT}. 

A key insight of this paper is that by exploiting the chain rule and Bayes' theorem, there exist two additional ways to constrain $\pi(\xv,\yv)$ with both $p_X(\xv)$ and $p_{Y}(\yv)$: 1) %
A forward CT that can be viewed as moving the source to target distribution; 2) A backward CT that reverses the  direction.
Our intuition is that given a source (target) point, it is more likely to be moved to a target (source) point closer to it. More specifically,
if the target distribution does not provide good coverage of the source density, then there will exist source data points that lie in low-density regions of the target, 
making the expected cost of the forward CT high. Therefore, we expect that minimizing the forward CT will encourage the target distribution to exhibit a \emph{mode-covering} behavior with respect to (\textit{w.r.t.}) the source PDF. Reversing the direction, we expect that minimizing the backward CT will encourage the target distribution to exhibit a \emph{mode-seeking} behavior \textit{w.r.t.} the source PDF. Minimizing the combination of both 
is expected to strike a good balance between these two distinct behaviors.

To demonstrate the use of 
CT, 
we apply it to train implicit (or explicit)
distributions to model %
both 1D and 2D toy data, MNIST digits, and natural images. The implicit distribution is defined by a deep generative model (DGM) that is simple to sample from. We provide empirical evidence to show how to control the mode-covering versus mode-seeking behaviors by adjusting the ratio of the forward CT versus backward CT. To train a DGM for natural images, 
we focus on adapting existing GANs, with minimal changes to their settings except for substituting the statistical distances in their loss functions with CT. We leave tailoring the network architectures and
settings to CT for future study. 
%
 %
Modifying the loss functions of various existing 
DGMs
with CT,
our experiments show consistent improvements in not only quantitative performance and generation quality, but also learning stability. Our code is available at \url{https://github.com/JegZheng/CT-pytorch}.

%
%

%

%
\section{
Chain rule and Bayes' theorem based
conditional transport} 
%

Exploiting the chain rule and Bayes' theorem, we can constrain $\pi(\xv,\yv)$ with both $p_X(\xv)$ and $p_{Y}(\yv)$ in two different ways, leading to the forward CT and backward CT, respectively.
To define the forward CT, we use the chain rule to factorize the joint distribution as 
$$\pi(\xv,\yv) =p_X(\xv) \pi_Y(\yv\given \xv),$$ where $\pi_Y(\yv\given \xv)$ is a conditional distribution of $\yv$ given $\xv$. This construction ensures $\int \pi(\xv,\yv) d\yv =p_{X}(\xv)$ but not $\int \pi(\xv,\yv) d\xv = p_{Y}(\yv)$. 
Denote 
$d_{\phiv}(\hv_1,\hv_2)\in \mathbb{R}$ as a function parameterized by $\phiv$, which measures the difference between two vectors $\hv_1,\hv_2\in \mathbb{R}^H$ of dimension $H$.  
While allowing $\int \pi(\xv,\yv) d\xv \neq p_{Y}(\yv)$, to appropriately constraint $\pi(\xv,\yv)$
by $p_Y(\yv)$, 
we treat $p_Y(\yv)$ as the prior distribution, view $e^{-d_{\phiv}(\xv,\yv)}$ as an unnormalized likelihood term, and follow Bayes' theorem to define 
\ba{
\pi_Y(\yv\given \xv) = e^{-d_{\phiv}(\xv,\yv)}p_Y(\yv)/Q(\xv),~~ \textstyle Q(\xv):=\int e^{-d_{\phiv}(\xv,\yv)}p_Y(\yv) \text{d}\yv,\label{eq:forwardCT_conditional}
}
where $Q(\xv)$ is a normalization term that ensures $\int \pi_Y(\yv\given \xv)d\yv=1$. We refer to $\pi_Y(\yv\given \xv)$ as the forward ``navigator,''  which specifies how likely a given $\xv$ will be mapped to a target point $\yv\sim p_{Y}(\yv)$. We now define the cost of the forward CT as
\ba{
\textstyle &\mathcal C 
(
X\rightarrow Y)=
%
%
%
\E_{\xv\sim p_X(\xv)}\E_{\yv\sim \pi_{Y}(\cdotv\given \xv) }[c(\xv,\yv)]\label{eq:OT_E_ygivenx}.
}
In the forward CT, we expect large $c(\xv,\yv)$ to typically co-occur with small $\pi_Y(\yv\given \xv)$ as long as $p_{Y}(\yv)$ provides a good coverage of the density of $\xv$. Thus minimizing the forward CT cost is expected to encourage $p_{Y}(\yv)$ to exhibit a mode-covering behavior \textit{w.r.t.} $p_{X}(\xv)$. Such kind of behavior is also expected when minimizing the forward KL divergence as  $\mathrm{KL}(p_X||p_Y)=\mathbb{E}_{\xv\sim p_X}\big[\ln \frac{p_X(\xv)}{p_Y(\xv)}\big]$, which calls for $p_Y(\xv)>0$ whenever $p_{X}(\xv)>0$.

Reversing the direction, we construct the backward CT, where the joint is factorized as 
$\pi(\xv,\yv) =p_Y(\yv) \pi_X(\xv\given \yv)$ and 
the backward navigator is defined as \ba{
\pi_X(\xv\given \yv)=e^{-d_{\phiv}(\xv,\yv)} p_X(\xv)/Q(\yv),~~~\textstyle Q(\yv):=\int e^{-d_{\phiv}(\xv,\yv)}p_X(\xv) \text{d}\xv.\label{eq:backwardCT_conditional}
}
This 
ensures $\int \pi(\xv,\yv) d\xv =p_{Y}(\yv)$; while allowing $\int \pi(\xv,\yv) d\yv \neq p_{X}(\xv)$, it constrains $\pi(\xv,\yv) $ by treating $p_X(\xv)$  as the prior to construct $\pi_X(\xv\given \yv)$. The backward CT cost is now defined as
\ba{
\textstyle \mathcal C 
(
X\leftarrow Y)
&= %
\E_{\yv\sim p_{Y}(\yv)}\E_{\xv\sim \pi_{X}(\cdotv\given \yv) }[c(\xv,\yv)]%
.
\label{eq:OT_E_xgiveny}
}
In the backward CT, we expect large $c(\xv,\yv)$  to typically co-occur with small ${\pi}_X(\xv\given \yv)$   as long as $p_X(\xv)$ has good coverage of the density of $\yv$. Thus minimizing the backward CT cost is expected to encourage $p_{Y}(\yv)$ to exhibit a mode-seeking behavior \textit{w.r.t.} $p_{X}(\xv)$. Such kind of behavior is also expected when minimizing the reverse KL  divergence as  $\mathrm{KL}(p_Y||p_X)=\mathbb{E}_{\xv\sim p_Y}\big[\ln \frac{p_Y(\xv)}{p_X(\xv)}\big]$, which allows $p_Y(\xv)=0$ when $p_X(\xv)>0$ and it is fine for $p_Y$ to just fit some portion of $p_X$.

In comparison to the forward and revers KLs, the proposed forward and backward CT are more broadly applicable as they don't require $p_X$ and $p_Y$ to share the same distribution support and have analytic PDFs. 
For the cases where the KLs can be evaluated, 
we introduce 
$$\mbox{D}(X,Y)= \mbox{KL}(p_X||p_Y)-\mbox{KL}(p_Y||p_X)$$ as a formal way to quantify the mode-seeking and mode-covering behavior of $p_Y$ w.r.t. $p_X$, with  $\mbox{D}(X,Y)>0$ implying mode seeking and with $D(X,Y)< 0$ implying mode covering.

Combining both the forward and backward CTs, we now define the CT cost as
\ba{
 %
 %
 \mathcal C_{\rho}(X, Y) 
 :=\textstyle 
 \rho\mathcal C(X\rightarrow Y)+ 
 (1-\rho)\mathcal C(X\leftarrow Y), 
 \label{eq:CT_divergence}
}
%
where $\rho\in[0,1]$ is a parameter that can be adjusted to encourage $p_{Y}(\yv)$ to exhibit \textit{w.r.t.} $p_{X}(\xv)$  mode-seeking ($\rho=0$), mode-covering ($\rho=1$), or a balance of two distinct behaviors ($\rho\in(0,1)$). 
By definition we have
 $\mathcal C_{\rho}(X, Y)\ge 0$, where the equality can be achieved 
when $p_X= p_Y$ and the navigator parameter $\phiv$ is optimized such that $e^{-d_{\phiv}(\xv,\yv)}$ is equal to one if and only if $\xv=\yv$ and zero otherwise. We also have $\mathcal C_{\rho=0.5}(X, Y)=\mathcal C_{\rho=0.5}(Y, X)$.  We fix $\rho=0.5$ unless specified otherwise.

\subsection{Conjugacy based analytic conditional distributions}\label{sec:theoretical}

Estimating the forward and backward CTs %
involves %
$\pi_Y(\yv\given \xv)$ and $\pi_X(\xv\given \yv)$, respectively. Both conditional distributions, however, are generally intractable to evaluate and sample from, unless 
$p_X(\xv)$ and $p_Y(\yv)$ are 
conjugate priors for likelihoods proportional to
$e^{-d(\xv,\yv)}$, $i.e.$, $\pi_X(\xv\given \yv)$ and $\pi_Y(\yv\given \xv)$ are in the same probability distribution family as $p_X(\xv)$ and $p_Y(\yv)$, respectively. For example, if $d(\xv,\yv)=\|\xv-\yv\|_2^2$ and both $p_{X}(\xv)$ and $p_{Y}(\yv)$ are multivariate normal distributions, then both $\pi_X(\xv\given \yv)$ and $\pi_Y(\yv\given \xv)$ will follow multivariate normal distributions. 

To be more specific, we provide a univariate normal based example, with $x,y,\phi,\theta\in \mathbb{R}$ and 
\ba{\textstyle &p_X(x)=\mathcal{N}(0,1),~p_Y(y)= \mathcal N(0, e^\theta),~d_{\phi}(x,y) = {(x-y)^2}/({2 e^{\phi}}),~ c(x,y)=(x-y)^2.\label{eq:1Dnormal}}
Here we have $\mbox{D}(X,Y)=  \mbox{KL}[\mathcal N(0,1)|| \mathcal N(0, e^\theta)] - \mbox{KL}[ \mathcal N(0, e^\theta) || \mathcal N(0,1)] = \theta - sinh(\theta)$, which is positive when $\theta<0$, implying mode-seeking, and negative when $\theta>0$, implying mode-covering.
As shown in Appendix~\ref{appendix: 1d_gaussian}, we have analytic forms of the 
forward and backward navigators as
$$\pi_{Y}(y\given x) = \mathcal{N}
(\sigma(\theta-\phi)x,\sigma(\theta-\phi)e^\phi),~~~
\pi_{X}(x\given y) = \mathcal{N}( %
\sigma(-\phi)y,\sigma(\phi)
),$$ 
where $\sigma(a)=1/(1+e^{-a})$ denotes the sigmoid function, and forward and backward CT costs as
$$
\mathcal C(X\rightarrow Y) = %
\sigma(\phi-\theta)(e^\theta+\sigma(\phi-\theta)),~~~
\mathcal C(X\leftarrow Y) =
\sigma(\phi)(1+\sigma(\phi)e^\theta).
$$
%
%
%

\begin{figure*}[t]
\centering
\includegraphics[width=\textwidth]{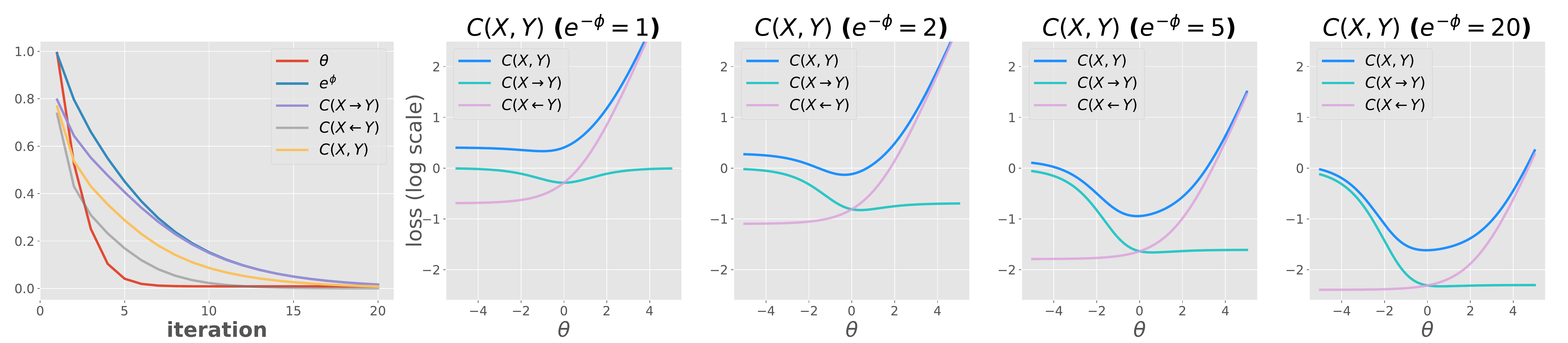}\vspace{-2mm}
\caption{ \small
Illustration of minimizing the CT cost $\mathcal C_{\phi,\theta}(X, Y)$ between $\mathcal N(0,1)$ and $\mathcal N(0, e^\theta)$. \textit{Left}:
Evolution of CT cost, its parameters, and forward and backward costs;
\textit{Right}: 4 CT cost curves against $\theta$ as $e^\phi$ is being optimized to a small value to jointly show the optimized $\phi$ provides better learning dynamic for the learning of $\theta$.
}\label{fig:1d_gaussian}\vspace{-4.5mm}
\end{figure*}

As a proof of concept, we illustrate the optimization under CT using the above example,
for which $\theta=0$ is the optimal solution that makes $p_X=p_Y$. 
Thus when applying gradient descent to minimize the CT cost $\mathcal C_{\rho=0.5}(X, Y)$, we expect the generator parameter $\theta\rightarrow 0$ with proper learning dynamic, as long as the learning of the navigator parameter $\phi$ is appropriately controlled. 
This is confirmed by Fig.\,\ref{fig:1d_gaussian}, which shows that %
as the navigator $\phi$ gets optimized by minimizing CT cost, it is more obvious that $\theta$ will minimize the CT cost at zero. This suggests that the navigator parameter $\phi$ mainly plays the role in assisting the learning of $\theta$. 
%
%
%
%
%
%
%
%
%
%
%
%
%
%
The right four subplots describe the log-scale curves of forward cost, backward cost and bi-directional CT costs  \textit{w.r.t.} $\theta$ as $\phi$ gets optimized to four different values. 
It is worth noting that the forward cost is minimized at $e^\theta>1$, which implies a mode-covering behavior, and the backward cost is minimized at $e^\theta\rightarrow 0$, which implies a mode-seeking behavior, while the bi-directional cost is minimized at around the optimal solution $e^{\theta}=1$; the forward CT cost exhibits a flattened curve on the right hand side of its minimum, adding to which the backward CT cost  not only moves that minimum left, making it closer to $\theta=0$, but also raises the whole curve on the right hand side, making the optimum of $\theta$ become easier to reach via gradient descent. 

To apply CT in a general setting where the analytical forms of the distributions are unknown, there is no conjugacy, or we only have access to random samples from the %
distributions, below we show we can
 approximate the CT cost 
by replacing both $p_X(\xv)$ and $p_{Y}(\yv)$ with their 
corresponding discrete empirical distributions supported on mini-batches.
%
%
Minimizing this approximate CT cost,
amenable to mini-batch 
SGD based optimization, is found to be effective 
in driving the target (model) distribution $p_Y$ towards the source (data) distribution $p_X$, with the ability to control the mode-seeking and mode-covering behaviors of $p_Y$ \textit{w.r.t.} $p_X$.

\subsection{Approximate CT given empirical samples}\label{sec:empirical}

Below we use generative modeling as an example to show how to apply the CT cost in a general setting that only requires access to random samples of %
both $\xv$ and $\yv$. 
Denote $\xv$ as a data taking its value in $%
\mathbb{R}^V$. %
In practice, we observe a finite set $\mathcal{X}=\{\xv_i\}_{i=1}^{|\mathcal{X}|}$, consisting of $|\mathcal{X}|$ data samples assumed to be $iid$ drawn from $p_X(\xv)$.
Given $\mathcal{X}$, the usual task is to learn a distribution to approximate $p_X(\xv)$, 
for which we consider %
a deep generative model (DGM) defined as 
$
\yv=G_{\thetav}(\epsilonv),~\epsilonv\sim p(\epsilonv), %
$
where $G_{\thetav}$ is a generator that transforms noise $\epsilonv\sim p(\epsilonv)$ via a deep neural network parameterized by $\thetav$ to generate random sample $\yv\in \mathbb{R}^V$. While the PDF %
of the generator, denoted as $p_{Y}(\yv;\thetav)$, is often intractable to evaluate, it is straightforward to draw $\yv\sim p_{Y}(\yv;\thetav)$ with $G_{\thetav}$. %
While knowing neither $p_X(\xv)$ nor $p_{Y}(\yv;\thetav)$, we can obtain discrete empirical 
distributions $p_{\hat X_N}$ and $ p_{\hat Y_M}$ supported on mini-batches $\xv_{1:N}$ and $\yv_{1:M}$,
%
as defined below, %
to guide the optimization of $G_{\thetav}$ in an iterative manner.
With $N$ random observations sampled without replacement from $\mathcal{X}$, we define 
\ba{\textstyle
%
p_{\hat X_N}(\xv)
=\frac{1}{N}\sum_{i=1}^N \delta{(\xv-\xv_i)},~~\{\xv_1,\ldots,\xv_N\}\subseteq \mathcal{X}
\label{eq:x}
}
as an empirical distribution for $\xv$. %
Similarly, with $M$ random samples of the generator, we define 
\ba{\textstyle
%
p_{\hat Y_M}(\yv)
=\frac{1}{M}\sum_{j=1}^M \delta{(\yv-\yv_j)},~\yv_j=G_{\thetav}(\epsilonv_j),~\epsilonv_j\, {\scriptstyle \stackrel{iid}{\sim}}\, p(\epsilonv)~~.
\label{eq:y}
}

Substituting $p_{Y}(\yv;\thetav)$ in \eqref{eq:OT_E_ygivenx} with
$p_{\hat Y_M}(\yv)$,
the continuous forward navigator becomes a discrete one as
\ba{
\textstyle\hat{\pi}_{Y}(\yv\given \xv)
&=\textstyle\sum_{j=1}^M \hat{\pi}_M(\yv_j \given \xv,\phiv) \delta_{\yv_j},~~\hat{\pi}_M(\yv_j \given \xv,\phiv) 
:=\textstyle\frac{e^{-d_{\phiv}(\xv,\yv_j) }}{\sum_{j'=1}^M e^{ -d_{\phiv}(\xv,\yv_{j'}) }}.\label{eq:pi_hat_Y}
}
Thus given $p_{\hat Y_M}$,
the cost of a forward CT can be approximated as
\ba{
\mathcal C_{\phiv,\thetav}( 
X\rightarrow \hat Y_M
) &
=\E_{\yv_{1:M}\, {\scriptstyle \stackrel{iid}{\sim}}\, p_{Y}(\yv;\thetav)}\E_{\xv\sim p_X(\xv)}\left[ \textstyle 
\textstyle \sum_{j=1}^M c(\xv,\yv_j)
\hat{\pi}_M(\yv_j\given \xv,\phiv)\right],\label{eq:x2nu}
}
which can be interpreted as the expected cost of following the forward %
navigator to stochastically transport a random source point $\xv$ %
to one of the $M$ %
randomly instantiated ``anchors'' of the target %
distribution. Similar to previous analysis, we expect this approximate forward CT to stay small as long as $p_{Y}(\yv;\thetav)$ exhibits a mode covering behavior \textit{w.r.t.} $p_{X}(\xv)$.

%
Similarly, %
we can approximate the backward navigator and CT cost as
%
\ba{
&\hat{\pi}_{X}(\xv\given \yv)
=\textstyle\sum_{i=1}^N \hat{\pi}_N(\xv_i \given \yv,\phiv) \delta_{\xv_i},~~
\hat{\pi}_N(\xv_i \given \yv,\phiv) 
:=\textstyle\frac{e^{-d_{\phiv}(\xv_i,\yv) }}{\sum_{i'=1}^N e^{ -d_{\phiv}(\xv_{i'},\yv) }},\notag\\
%
%
%
&\mathcal C_{\phiv,\thetav}( 
\hat X_N \leftarrow Y
)  
=\E_{\xv_{1:M}\, {\scriptstyle \stackrel{iid}{\sim}}\, p_{X}(\xv)}\E_{\yv\sim p_{Y}(\yv;\thetav)}\left[ 
%
\textstyle \sum_{i=1}^N c(\xv_i,\yv)
\hat{\pi}_N(\xv_i\given \yv,\phiv)\right]. \label{eq:y2mu}
}
Similar to previous analysis, we expect this approximate backward CT to stay small as long as $p_{Y}(\yv;\thetav)$ exhibits a mode-seeking behavior \textit{w.r.t.} $p_{X}(\xv)$.

%
Combining 
\eqref{eq:x2nu} and \eqref{eq:y2mu}, 
we define the approximate CT cost as
\ba{
%
%
 \mathcal{C}_{\phiv,\thetav,\rho}(\hat X_N, \hat Y_M )=
%
\rho
\mathcal C_{\phiv,\thetav}( X\rightarrow \hat Y_M) +
(1-\rho)
\mathcal C_{\phiv,\thetav}( \hat X_N\leftarrow Y) ,%
\label{eq:C_NM}
}
 an unbiased sample estimate of which, given mini-batches $\xv_{1:N}$ and $\yv_{1:M}$, can be expressed as
\ba{
\mathcal{L}_{\phiv,\thetav,\rho}(\xv_{1:N},\yv_{1:M})&=\textstyle\sum_{i=1}^N\sum_{j=1}^Mc(\xv_i,\yv_j)\left(
\textstyle\frac{\rho}{N}\hat{\pi}_M(\yv_j\given \xv_i,\phiv)+
\textstyle\frac{1-\rho}{M}\hat{\pi}_N(\xv_i\given \yv_j,\phiv)\right)\notag\\
&\!\!\!\!\!\!\!\!\!\!\!\!\!=\textstyle\sum_{i=1}^N\sum_{j=1}^Mc(\xv_i,\yv_j)\left(
\textstyle\frac{\rho}{N}
\textstyle\frac{e^{-d_{\phiv}(\xv_i,\yv_{j}) }}{\sum_{j'=1}^M e^{ -d_{\phiv}(\xv_{i},\yv_{j'}) }}
+
\textstyle\frac{1-\rho}{M}\textstyle\frac{e^{-d_{\phiv}(\xv_i,\yv_j) }}{\sum_{i'=1}^N e^{ -d_{\phiv}(\xv_{i'},\yv_j) }}\right)
\label{eq:BOT_sample}.
}

\begin{lemma}\label{lemma:limit}
Approximate CT in \eqref{eq:C_NM} is asymptotic as %
$\lim_{N,M\rightarrow \infty} %
\mathcal{C}_{\phiv,\thetav,\rho}(\hat X_N, \hat Y_M )
=\mathcal C_{\phiv,\thetav,\rho}(X,Y).$%
\end{lemma}

\subsection{Cooperatively-trained or adversarially-trained feature encoder}
\label{sec:feature_extraction}
To apply CT for generative modeling of high-dimensional data, such as natural images, 
we need to define an appropriate cost function $c(\xv,\yv)$ to measure the difference between two random points. A naive choice is some distance between their raw feature vectors, such as
$
c(\xv,\yv) = \|\xv-\yv\|_2^2
$, which, however, is known to often poorly reflect the difference between high-dimensional data residing on low-dimensional manifolds. For this reason, with cosine similarity \citep{salimans2018improving} as %
$ 
{\textstyle\cos(\hv_1,\hv_2) :=\frac 
{\hv_1^T\hv_2}{{\sqrt{\hv_1^T\hv_1}}{\sqrt{\hv_2^T\hv_2}}}}$, 
we further introduce a feature encoder 
$\mathcal T_{\etav}(\cdotv)$, parameterized by $\etav$, %
to help redefine the point-to-point cost  and both navigators  %
as
\ba{
\textstyle c_{\etav}(\xv,\yv) %
= 1 -
\cos(\mathcal{T}_{\etav}(\xv),\mathcal{T}_{\etav}(\yv) ),~~~
d_{\phiv}\left(\frac{\mathcal{T}_{\etav}(\xv)}{\|\mathcal{T}_{\etav}(\xv)\|},\frac{\mathcal{T}_{\etav}(\yv)}{\|\mathcal{T}_{\etav}(\yv)\|}\right).
\label{eq:d_eta}
}
%
To apply the CT cost to train a DGM, we find that the feature encoder $\mathcal{T}_{\etav}(\cdotv)$ can be learned in two different ways: 1) Cooperatively-trained: Training them cooperatively by alternating between two different losses: training the generator under a fixed $\mathcal{T}_{\etav}(\cdotv)$ with the CT loss, and training $\mathcal{T}_{\etav}(\cdotv)$ under a fixed generator with a different loss, such as the GAN discriminator loss, WGAN critic loss, and MMD-GAN \cite{binkowski2018demystifying} critic loss. 2) Adversarially-trained: Viewing the feature encoder as a critic and training it to maximize the CT cost, by not only inflating the point-to-point cost, but also distorting the feature space used to construct the forward and backward navigators' conditional distributions. 

To be more specific, below we present the details for the adversarial way to train $\mathcal{T}_{\etav}$. 
Given training data  $\mathcal X$,
to train the generator $G_{\thetav}$, forward navigator $\pi_{\phiv}(\yv\given \xv)$, backward navigator $\pi_{\phiv}(\xv\given \yv)$, and encoder $\mathcal T_{\etav}$,
we view the encoder as a critic and propose to solve
a min-max problem as
\ba{\!\!\!
\min_{\phiv,\thetav}\max_{\etav} \E_{\xv_{1:N}\subseteq \mathcal{X},
~\epsilonv_{1:M}\, {\scriptstyle \stackrel{iid}{\sim}}\, p(\epsilonv)}[\mathcal{L}_{\phiv,\thetav,\rho,\etav}(\xv_{1:N},\{G_{\thetav}(\epsilonv_{j})\}_{j=1}^M)],
\label{eq:min-max}
}
where $\mathcal{L}_{\phiv,\thetav,\rho,\etav}$ is defined the same as in 
\eqref{eq:BOT_sample}, except that we replace $c(\xv_i,\yv_j)$ and $d_{\phiv}(\cdotv,\cdotv)$  with
their corresponding ones
shown in \eqref{eq:d_eta} and  use reparameterization in \eqref{eq:y} to draw $\yv_{1:M}:=\{G_{\thetav}(\epsilonv_{j})\}_{j=1}^M$. %
With SGD, we update $\phiv$ and $\thetav$  using $\nabla_{\phiv,\thetav} \mathcal{L}_{\phiv,\thetav,\rho,\etav}(\xv_{1:N},\{G_{\thetav}(\epsilonv_{j})\}_{j=1}^M))$ and, if the feature encoder is 
adversarially-trained, update $\etav$  using
$
-\nabla_{\etav} \mathcal{L}_{\phiv,\thetav,\rho,\etav}(\xv_{1:N},\{G_{\thetav}(\epsilonv_{j})\}_{j=1}^M))
$.

We find by experiments that both ways to learn the encoder work well, with the adversarial one generally providing better performance. It is worth noting that in (Wasserstein) GANs, while the adversarially-trained discriminator/critic  plays a similar role as a feature encoder, the learning dynamics between the discriminator/critic and generator need to be carefully tuned to maintain training stability and prevent trivial solutions ($e.g.$, mode collapse). By contrast, the feature encoder of the CT cost based DGM can be stably trained in two different ways. Its update does not need to be well synchronized with the generator and can be stopped at any time of the training.

\section{Related work}

In practice,  variational auto-encoders \citep{kingma2013auto}, the KL divergence based deep generative models, are stable to train, but often exhibit mode-covering behaviors and generate blurred images \citep{chen2016variational,zhao2017infovae,zheng2018degeneration,alemifixing,zheng2019understanding}. By contrast, both GANs and Wasserstein GANs can generate photo-realistic images, but they often suffer from stability and mode collapse issues,  requiring the update of the discriminator/critic to be well synchronized with that of the generator.  
This paper 
introduces
conditional transport (CT)
as
a new method to quantify the difference between two probability distributions. Deep generative models trained under CT not only allow the balance between mode-covering and mode-seeking behaviors to be adjusted, but also allow the encoder to be pretrained or frozen at any time during cooperative/adversarial training.  
%

As the JS divergence requires the two distributions to have the same support,
the Wasserstein distance is often considered as 
more appealing for generative modeling as it allows the two %
distributions to have non-overlapping support \citep{villani2008optimal,santambrogio2015optimal,COTFNT}. However, while GANs and Wasserstein GANs in theory are connected to the JS divergence and Wasserstein distance, respectively, several recent works show that they should not be naively understood as the minimizers of 
their corresponding statistical distances, and the role played by their min-max training dynamics should not be overlooked \citep{kodali2017convergence,fedus2018many,stanczuk2021wasserstein}.
In particular, \citet{fedus2018many} show that even when the gradient of the JS divergence does not exist and hence GANs are predicted to fail from the perspective of divergence minimization, the discriminator is able to provide useful learning signal.
\citet{stanczuk2021wasserstein} show 
that the dual form based Wasserstein GAN loss does not provide a meaningful approximation of the Wasserstein distance; while primal form based  methods could better approximate the true Wasserstein distance, they in general clearly underperform Wasserstein GANs in terms of the  generation quality for high-dimensional data, such as natural images, and require an inner loop to compute the transport plan for each mini-batch, leading to high computational cost \citep{genevay2018learning,iohara2019generative,mallasto2019well,pinetz2019estimation,stanczuk2021wasserstein}. See previous works for discussions on the approximation error and gradient bias when estimating the Wasserstein distance with mini-batches 
\cite{bottou2017geometrical,bellemare2017cramer,
binkowski2018demystifying,bernton2019parameter}.

MMD-GAN \citep{li2015generative,li2017mmd,binkowski2018demystifying} that calculates the MMD statistics in the latent space of a feature encoder is the most similar to the CT cost in terms of the actual loss function used for optimization. In particular, both the MMD-GAN loss and CT loss, given mini-batches $\xv_{1:N}$ and $\yv_{1:M}$, involve computing the differences of all $NM$ pairs $(\xv_i,\yv_j)$. Different from MMD-GAN, there is no need in CT to choose a kernel and tune its parameters. We provide below an ablation study to evaluate both 1) MMD generator + CT encoder and 2) MMD encoder + CT generator, which shows 1) performs on par with MMD, while 2) performs clearly better than MMD and on par with CT.
\section{Experimental results}\label{sec:experiments}

\begin{figure}[t]
\centering
\includegraphics[width=.9\textwidth]
        {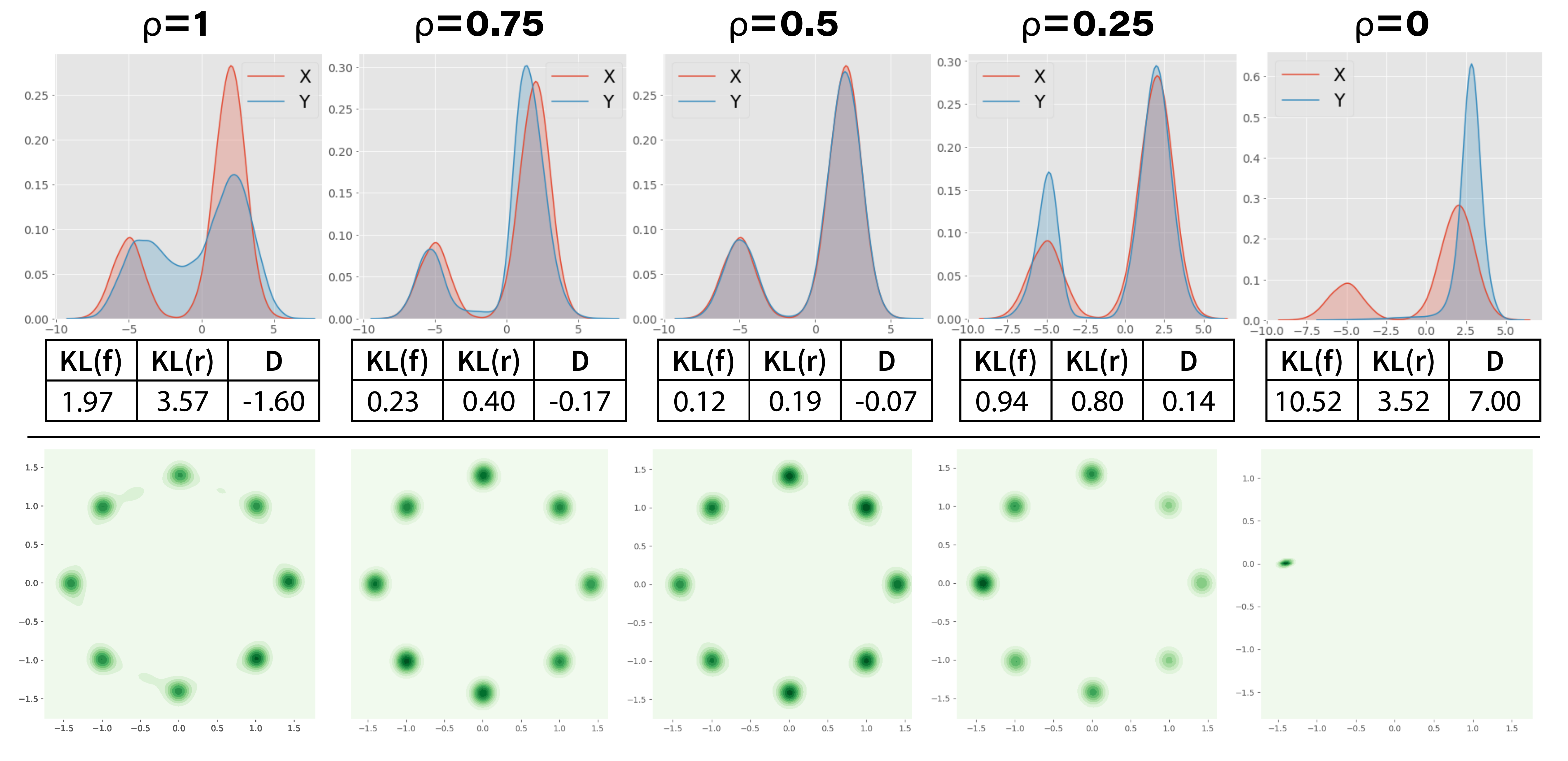}\vspace{-4mm}
\caption{\small Forward and backward analysis: (\textit{top)} Fitting 1D bi-modal Gaussian. Quantitative results of estimated forward KL (KL[$\hat p_X||\hat p_Y$]), reverse KL (KL[$\hat p_Y||\hat p_X$]), and the difference between the forward and reverse KL (D=KL[$\hat p_X||\hat p_Y$]-KL[$\hat p_Y||\hat p_X$]) are shown below each sub-figure. (\textit{bottom}) 2D 8-Gaussian mixture by interpolating between the forward CT ($\rho=1$) and backward CT ($\rho=0$). 
}\label{fig:ACT_fb}\vspace{-4mm}
\end{figure}

\textbf{Forward and backward analysis: }
To empirically verify our previous analysis of the mode covering (seeking) behavior of 
the forward (backward) CT,  we train a DGM with %
\eqref{eq:C_NM}
and show the corresponding interpolation weight from the forward CT cost to the backward one, which means 
CT$_\rho$ reduces from forward CT ($\rho = 1$), to the CT in \eqref{eq:C_NM} ($\rho\in(0,1)$), and to backward CT ($\rho = 0$).
We consider the squared Euclidean ($i.e.$ $\mathcal L_2^2$) distance to define both cost $c(\xv,\yv)=\|\xv-\yv\|_2^2$ and $d_{\phiv}(\xv,\yv)= \|\mathcal T_{\phiv}(\xv)-\mathcal T_{\phiv}(\yv)\|_2^2$,
where $\mathcal T_{\phiv}$ denotes a neural network parameterized by $\phiv$.
We consider a 1D example of a bimodal Gaussian mixture $p_X(x) = \frac{1}{4}\mathcal{N}(x;-5,1) + \frac{3}{4}\mathcal{N}(x;2,1)$ and a 2D example of 8-modal Gaussian mixture with equal component weight  as in \citet{gulrajani2017improved}. We use an empirical sample set $\mathcal X$, consisting of $|\mathcal X|=5,000$ samples 
from both 1D and 2D cases, and illustrate  in Fig.\,\ref{fig:ACT_fb} the KDE of {5000} generated samples $y_j=G_{\thetav}(\epsilonv_j)$ after 5000 training epochs. {For the 1D case, we take 200 grids in  $[-10,10]$ to approximate the empirical distribution of $\hat p_X$ and $\hat p_Y$, and report the corresponding forward KL (KL[$\hat p_X||\hat p_Y$]), reverse KL (KL[$\hat p_Y||\hat p_X$]), and their difference $\mathrm{D}(X,Y) = \text{KL}[\hat p_X||\hat p_Y] - \text{KL}[\hat p_X||\hat p_Y]$ below each corresponding sub-figure in Fig.\,\ref{fig:ACT_fb}.}

%
%
Comparing the results of different $\rho$ in Fig.\,\ref{fig:ACT_fb}, it suggests that minimizing the forward CT cost only encourages the generator to exhibit mode-covering behaviors, while minimizing the backward CT cost only encourages mode-seeking behaviors.
Combining both costs provides a user-controllable balance between mode covering and seeking, leading to satisfactory fitting performance, as shown in Columns 2-4.
Note that for a fair comparison, we stop the fitting at the same iteration; in practice, we find if training with more iterations, both $\rho = 0.75$ and $\rho = 0.25$ can achieve comparable results as $\rho=0.5$ in this example. 
Allowing the mode covering and seeking behaviors to be controlled by adjusting $\rho$ is an attractive property of CT$_\rho$. 

\begin{figure}[t]
\centering
        \includegraphics[width=.9\textwidth]{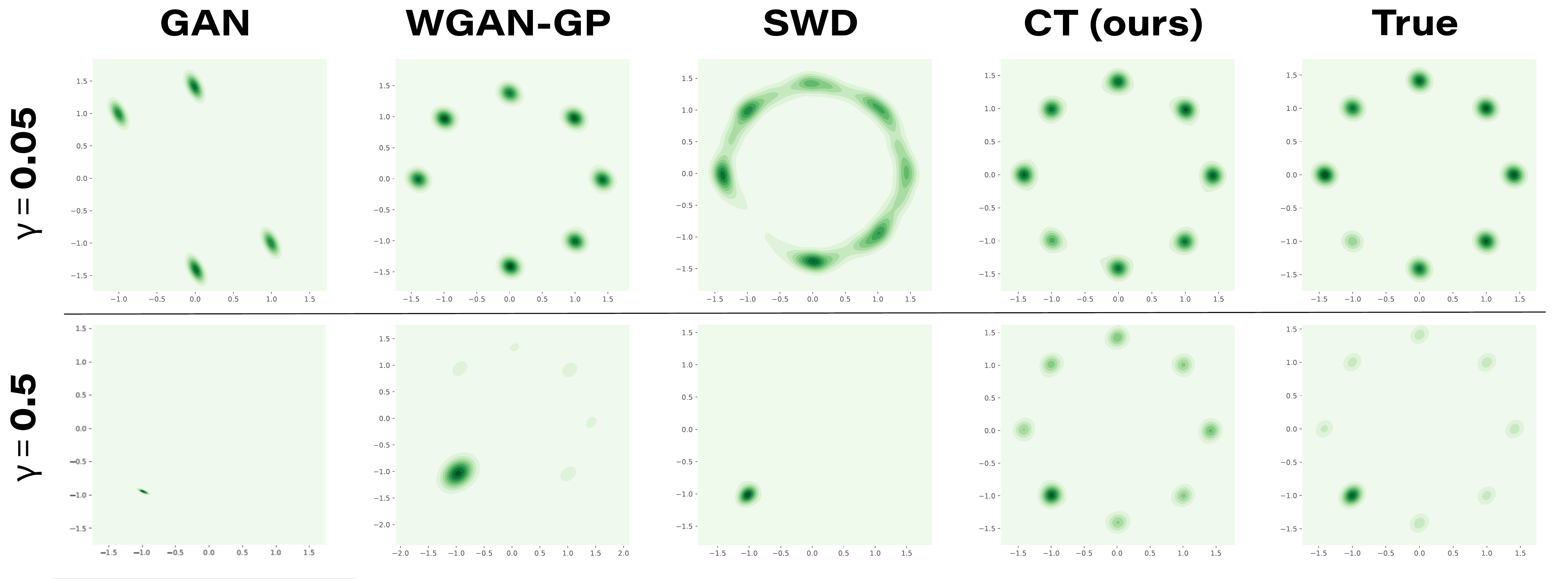}
        \vspace{-4mm}
 \caption{\small Experiments on the resistance to model collapse: Comparison of the generation quality on 8-Gaussian mixture data: one of the 8 modes has weight $\gamma$ and the rest modes have equal weight as $\frac{1-\gamma}{7}$. %
 }\vspace{-5mm}
 \label{fig:2d_biased_8gaussian}
\end{figure}

\textbf{Resistance to mode collapse: } We continue to use a 8-Gaussian mixture to empirically evaluate how well a DGM resists mode collapse. %
Unlike the data in Fig.\,\ref{fig:ACT_fb}, where 8 modes are equally weighted, here the mode at the left lower corner is set to have weight $\gamma$, while the other modes are set to have the same weight of %
$\frac{1-\gamma}{7}$. We set $\mathcal{X}$ with 5000 samples and the mini-batch size as $N=100$. %
When $\gamma$ is lowered to 0.05, %
its corresponding mode is shown to be missed by
GAN, WGAN, and SWD-based DGM, while well kept by the CT-based DGM.
As an explanation, GANs are known to be susceptible to mode collapse; WGAN and SWD-based DGMs are sensitive to the mini-batch size, as when $\gamma$ equals to a small value, the samples from this mode will appear in the mini-batches less frequently than those from any other mode, amplifying their missing mode problem. Similarly, when $\gamma$ is increased to 0.5, the other modes are likely to be missed by the baseline DGMs, while the CT-based DGM does not miss any modes. The resistance of CT to mode dropping can be attributed to its forward component's 
mode-covering property. The backward's mode-seeking property further helps distinguish the density of each mode component to avoid making components of equal weight.

\textbf{CT for 2D toy data and robustness in adversarial feature extraction:} To test CT with more general cases, we further conduct experiments on 4 representative 2D datasets for generative modeling evaluation \cite{gulrajani2017improved}: 8-Gaussian mixture, Swiss Roll, Half Moons, and 25-Gaussian mixture. We apply the vanilla GAN \citep{goodfellow2014generative} and Wasserstein GAN with gradient penalty (WGAN-GP) \citep{gulrajani2017improved} as two representatives of min-max
DGMs that require solving a min-max loss. We then apply the generators trained under the sliced Wasserstein distance (SWD) \citep{deshpande2018generative} and CT cost as two representatives of min-max-free DGMs. Moreover, we include CT with an adversarial feature encoder trained with \eqref{eq:d_eta} to test the robustness of adversary and compare with the baselines in solving the min-max loss.

On each 2D data, we train these DGMs as one would normally do during the first $5k$ epochs. We then only train the generator and freeze all the other learnable model parameters, which means we freeze the discriminator in GAN, critic in WGAN, the navigator parameter $\phiv$ of the CT cost, and both $(\phiv,\etav)$ of CT with an adversarial feature encoder, for another $5k$ epochs. Figs.\,\ref{fig:2d_8gaussian}-\ref{fig:2d_25gaussian} in Appendix~\ref{appendix:2d_toy} illustrate this training process on each dataset, where %
for both min-max baseline DGMs, the models collapse after the first $5k$ epochs, while the training for SWD remains stable and that for CT continues to improve. Compared to SWD, our method covers all data density modes and moves the generator much closer to the true data density. Notably, for CT with an adversarially trained feature encoder, although it 
has switched from
solving a min-max loss to freezing the feature encoder after 5$k$ epochs, the frozen feature encoder continues to guide the DGM to finish the training in the last $5k$ epochs, which shows the robustness of the CT cost.

\begin{figure}[t]
\centering
\begin{subfigure}[t]{0.20\textwidth}
    \includegraphics[width=\textwidth]
    {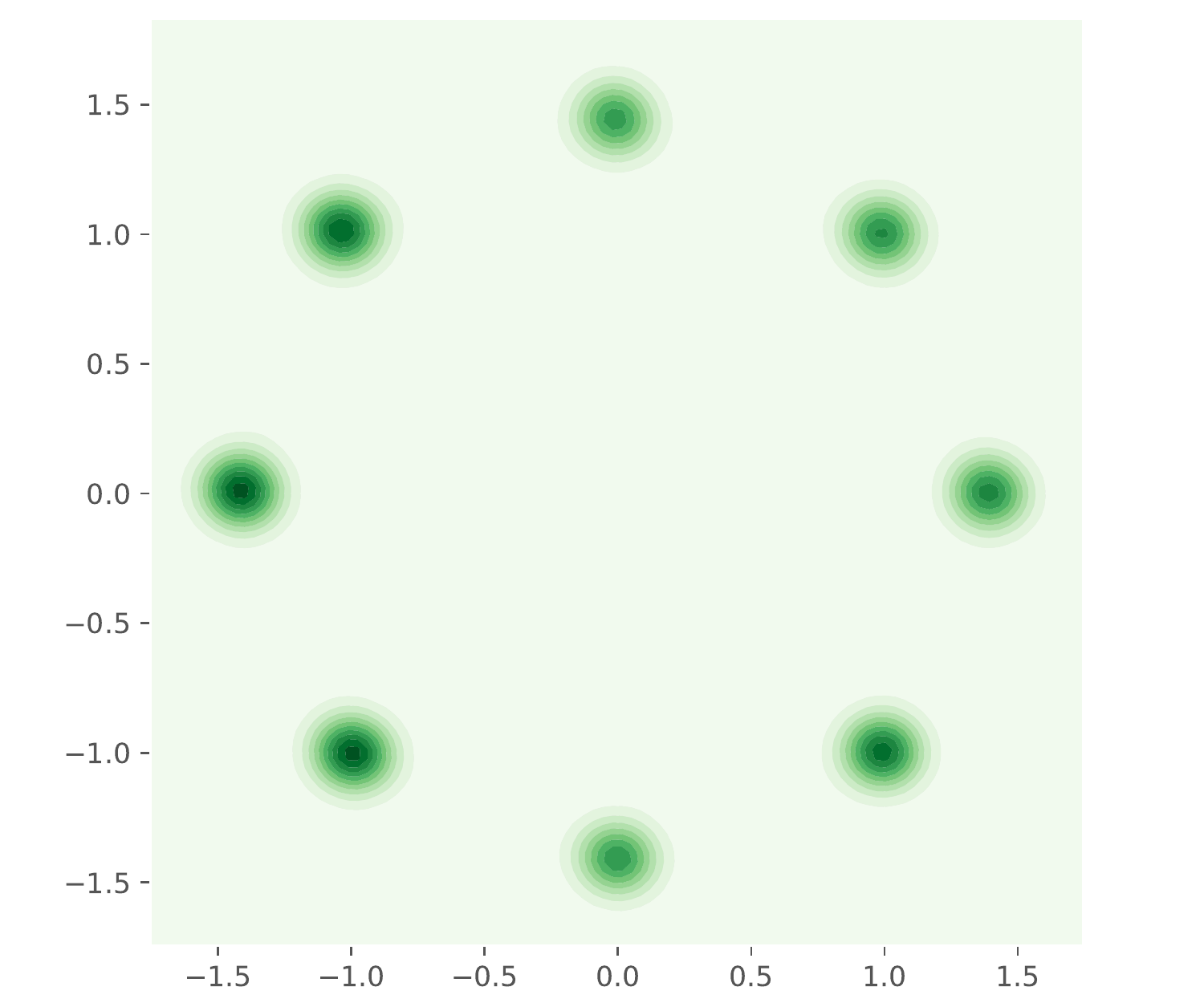}\vspace{-2mm}
\caption{\small Adv CT.
}\label{fig:minmax_CT}
\end{subfigure}\vrule\hfill
\begin{subfigure}[t]{0.18\textwidth}
    \includegraphics[width=\textwidth]
    {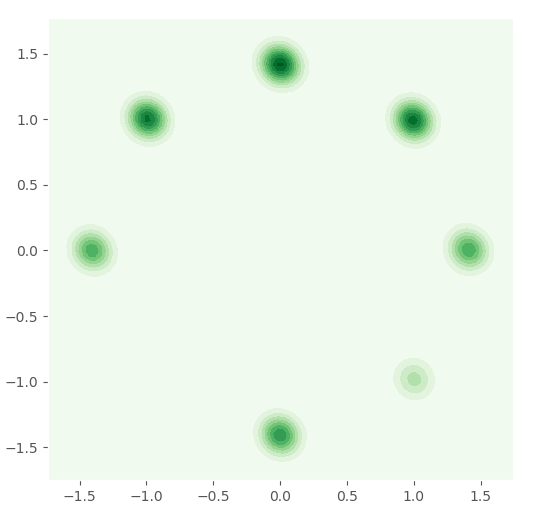}\vspace{-2mm}
\caption{\small GAN.
}\label{fig:min_gan}
\end{subfigure}\hfill
\begin{subfigure}[t]{0.20\textwidth}
    \includegraphics[width=\textwidth]
    {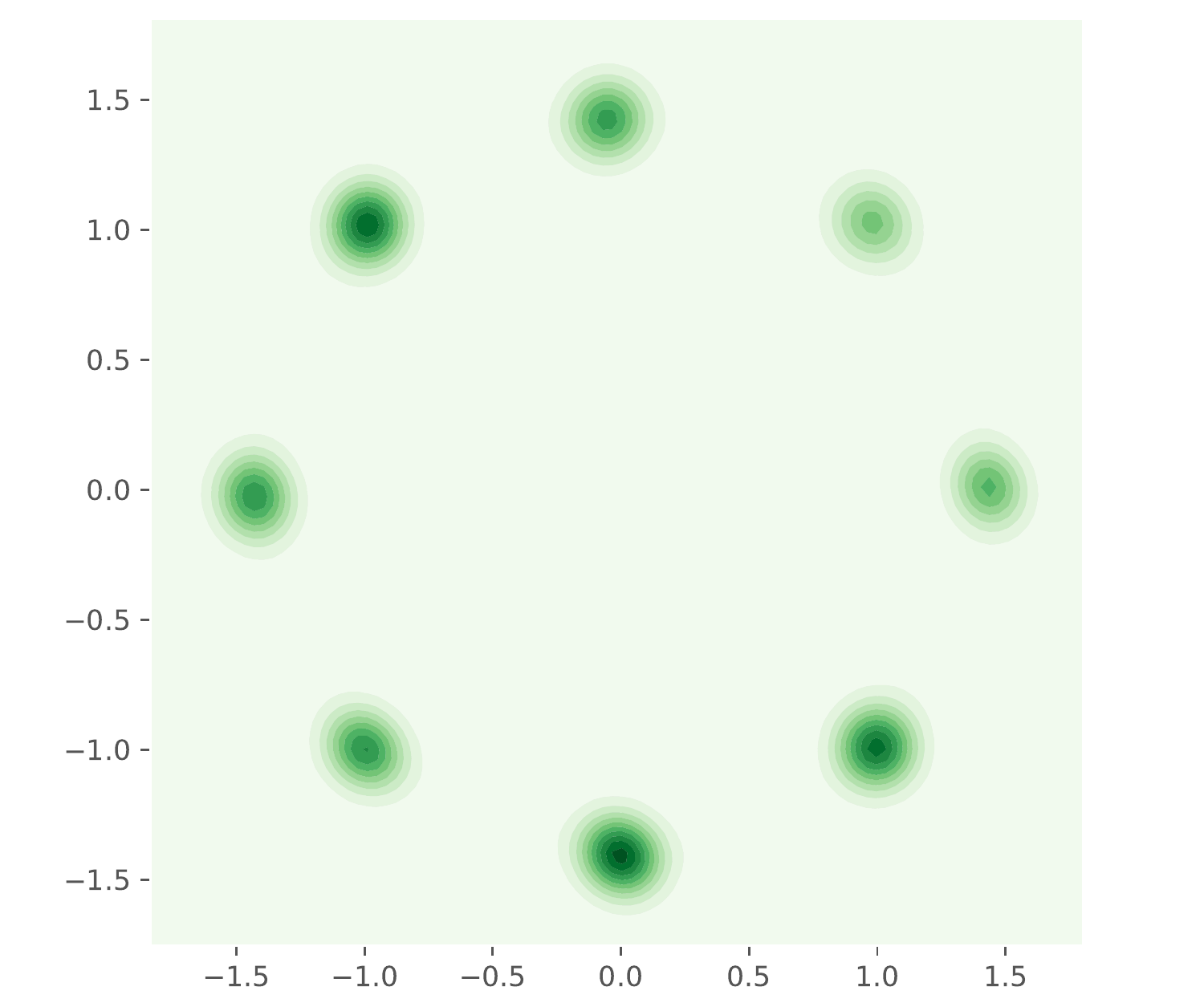}\vspace{-2mm}
\caption{\small $\mathcal{L}_D$ + CT.
}\label{fig:min_CT_max_d}
\end{subfigure}\vrule\hfill
\begin{subfigure}[t]{0.18\textwidth}
\includegraphics[width=\textwidth]
{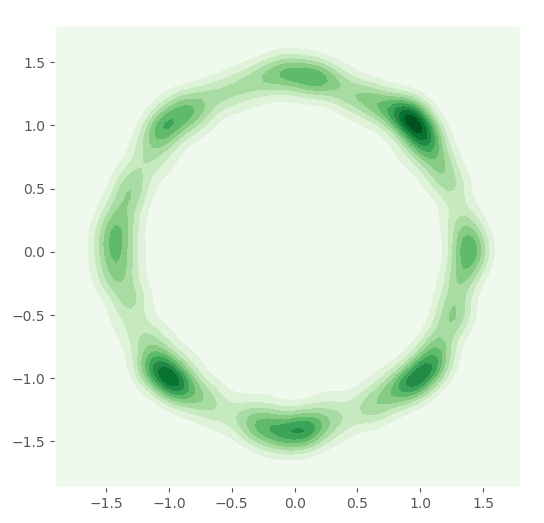}\vspace{-2mm}
\caption{\small SWD.
}\label{fig:min_slice}
\end{subfigure}\hfill
\begin{subfigure}[t]{0.20\textwidth}
\includegraphics[width=\textwidth]
{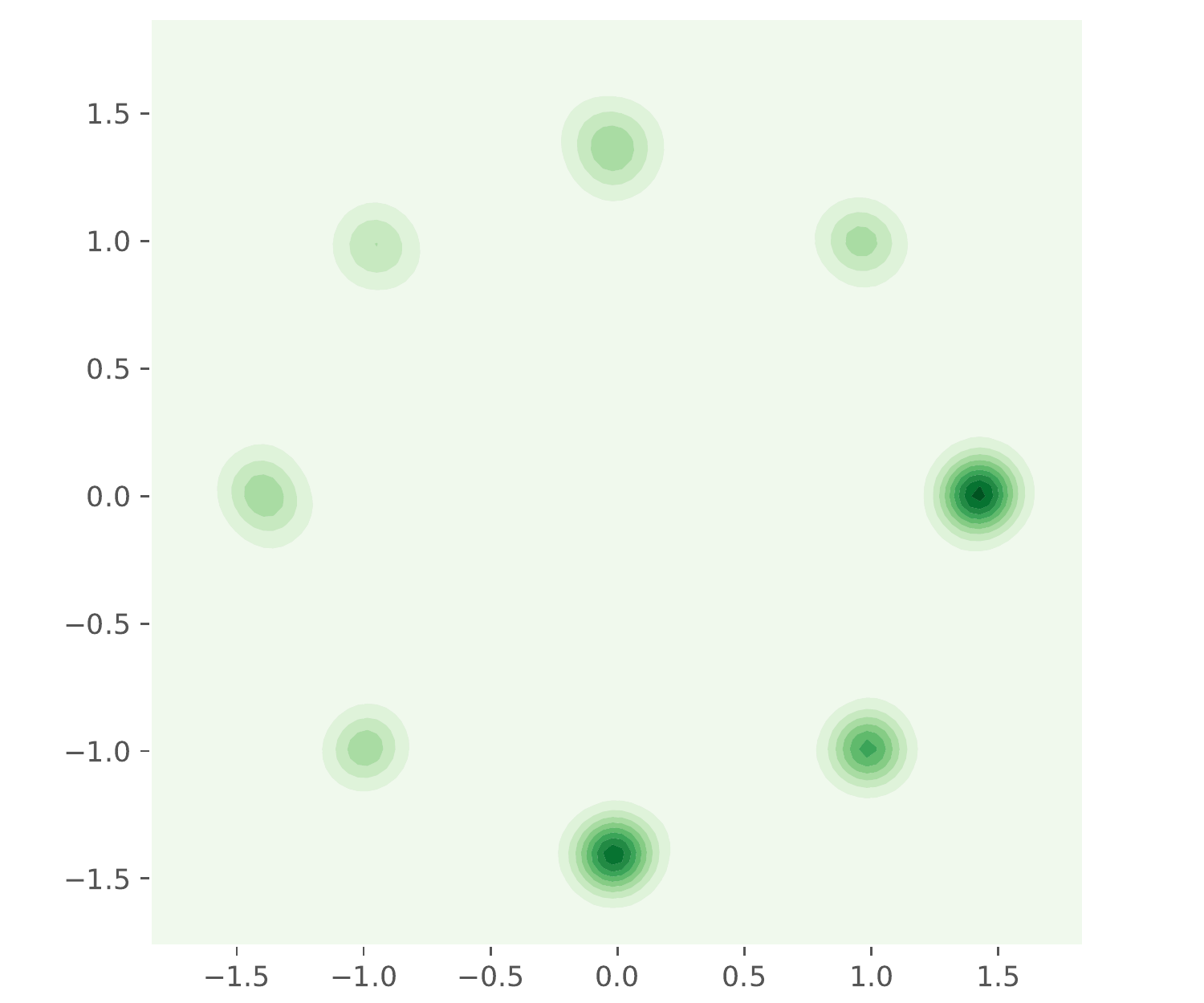}\vspace{-2mm}
\caption{\small Slicing + CT.
}\label{fig:min_slicedCT}
\end{subfigure}
\vspace{-1mm}
\caption{\small Ablation of fitting results by minimizing CT in different spaces: (a) CT calculated with adversarially trained encoder. (b-c) GAN \textit{vs.} CT with feature space cooperatively trained with discriminator loss. (d-f) Sliced Wasserstein distance and CT in the sliced space. 
}\label{fig:ablation_coop}\vspace{-1mm}
\end{figure}

\begin{table}[t]
\vspace{-4mm}
  \begin{minipage}[t]{.34\columnwidth}
    \centering
    \caption{\small FID comparison with different cooperative training on CIFAR-10 (lower FID is preferred). }
    \label{tab:comparison_co-train}
    \renewcommand{\arraystretch}{1.2}
    \setlength{\tabcolsep}{1.0mm}{ 
    \begin{tabular}{|c|c|}
    \hline
    Critic space & FID $\downarrow$ \\ \hline
    Discriminator &  29.7 \\
    Slicing & 32.4 \\ \hline
    Adversarial CT & \textbf{22.1}
    \\ \hline 
    \end{tabular}
    }
  \end{minipage}~~~~~~~\hfill
  \begin{minipage}[t]{.63\columnwidth}
    \centering
    \caption{\small FID Comparison with using MMD (Rational quadratic kernel/distance kernel) and CT loss in training critic/generator on CIFAR-10 (lower FID is preferred). }
    \label{tab:comparison_MMD}
    \renewcommand{\arraystretch}{1.2}
    \setlength{\tabcolsep}{1.0mm}{ 
    \begin{tabular}{|c|c|c|c|c|c|c|c|}
    \hline
     \multicolumn{2}{|c|}{\multirow{2}{*}{MMD-rq}} & \multicolumn{2}{c|}{Generator loss} & \multicolumn{2}{|c|}{\multirow{2}{*}{MMD-dist}} & \multicolumn{2}{c|}{Generator loss}\\ \cline{3-4} \cline{7-8}
     \multicolumn{2}{|c|}{} & MMD & CT & \multicolumn{2}{|c|}{} & MMD & CT \\ \hline
    {{Critic}} & MMD & 39.9 & 24.1 & {{Critic}} & MMD & 40.3 & \textbf{28.8}  \\ \cline{2-4} \cline{6-8}
    {loss} & CT & 41.4 & \textbf{23.9} & {loss} & CT & 30.9 & {29.4}  \\\hline
    \end{tabular}
    }
  \end{minipage}~~~~~
  \vspace{-3mm}
\end{table}

\textbf{Ablation of cooperatively-trained and adversarially-trained CT:} As previous experiments show the adversarially-trained feature encoder could provide a valid feature space for CT cost, we further study the performance of the encoders cooperatively trained with other losses. Here we leverage,  as two alternatives, the space of an encoder trained with the discriminator loss in GANs and the empirical Wasserstein distance in sliced 1D spaces \cite{wu2019sliced}. We test these settings on both 8-Gaussian, as shown Fig.\,\ref{fig:ablation_coop}, and CIFAR-10 data, as shown in Table~\ref{tab:comparison_co-train}. It is confirmed these encoders are able to cooperatively work with CT, in general producing less appealing results with those trained by maximizing CT. From this view, although CT is able to provide guidance for the generators in the feature space learned with various options, maximizing CT is still preferred to ensure the efficiency. Moreover, as observed in Figs.\,\ref{fig:min_gan}-\ref{fig:min_slicedCT}, CT clearly improves the fitting with sliced Wasserstein distance.  To explain why CT helps improve in the sliced space, we further provide a toy example in 1D to study the properties of CT and empirical Wasserstein distance in Appendix~\ref{appendix:WvsCT}.


\textbf{Ablation of MMD and CT:}
As MMD also compares the pair-wise sample relations in a mini-batch, 
we study if MMD and CT can benefit each other. The feature space of MMD-GAN can be considered as $\mathcal{T}_{\etav} \circ k$, where $k$ is the rational quadratic or distance kernel in {\citet{binkowski2018demystifying}}. Here we evaluate the combinations of MMD/CT as the generator/encoder criterion to train DGMs. 
On $\text{CIFAR-10}$, shown in Table~\ref{tab:comparison_MMD}, combining MMD and CT generally has improvement over MMD alone in FID. It is interesting to notice that for MMD-GAN, learning its generator with the CT cost shows more obvious improvement than learning its feature encoder with the CT cost. We speculate the estimation of MMD relies on a supremum of its witness function, which needs to be maximized \textit{w.r.t} $\mathcal{T}_{\etav} \circ k$ and cannot be guaranteed by maximizing CT \textit{w.r.t} $\mathcal{T}_{\etav}$. In the case of MMD-dist, using CT for witness function updates shows a more clear improvement, probably because CT has a similar form as MMD when using the distance kernel. From this view, CT and MMD are naturally able to be combined to compare the distributional difference with pair-wise sample relations. Different from MMD, CT does not involve the choice of kernel and its navigators assist to improve the comparison efficiency.
Below we show on more image datasets, CT is compatible with many existing models, and achieve good results to show improvements on a variety of data with different scale.

\textbf{Adversarially-trained CT for natural images: } 
We conduct a variety of experiments on natural images to evaluate the performance and reveal the properties of DGMs optimized under the CT cost. 
We consider three widely-used image datasets, including CIFAR-10 \citep{cifar10}, CelebA \citep{celeba}, and LSUN-bedroom \citep{lsun} for general evaluation, as well as CelebA-HQ \cite{karras2018progressive}, FFHQ \cite{karras2019style} for evaluation in high-resolution. We compare the results of DGMs optimized with the CT cost against DGMs %
trained with their original criterion including DCGAN \citep{radford2015unsupervised}, Sliced Wasserstein Generative model (SWG) \cite{deshpande2018generative}, MMD-GAN \citep{binkowski2018demystifying}, SNGAN \citep{miyato2018spectral}, and StyleGAN2 \cite{Karras2019stylegan2}. For fair comparison, we leverage the best configurations reported in their corresponding paper or Github page. The detailed setups can be found in Appendix~\ref{app:experiment detail}. For evaluation metric, we 
consider
the commonly used Fr\'echet inception distance (FID, lower is preferred) \cite{heusel2017gans} on all datasets and Inception Score (IS, higher is preferred) \cite{salimans2016improved} on CIFAR-10.
Both FID and IS are calculated using a pre-trained inception model 
\citep{szegedy2016rethinking}.

\begin{table}[]
\centering
\caption{\small Results of CT with different deep generative models on CIFAR-10, CelebA and LSUN. Base model results are quoted from corresponding paper or github page. }\label{tab:fid}
\resizebox{.85\columnwidth}{!}{%
\begin{tabular}{ccccc}
\toprule
\multirow{2}{*}{Method} & \multicolumn{3}{c}{Fr\'echet Inception Distance (FID  $\downarrow$)} & Inception Score ($\uparrow$) \\ \cmidrule(lr){2-4} \cmidrule(lr){5-5}
 & CIFAR-10 & CelebA & LSUN-bedroom & CIFAR-10 \\
\cmidrule(lr){1-1} \cmidrule(lr){2-4} \cmidrule(lr){5-5}
DCGAN \citep{radford2015unsupervised} & 30.2$\pm$0.9 & 52.5$\pm$2.2 & 61.7$\pm$2.9 & 6.2$\pm$0.1 \\
%
CT-DCGAN
& 
\textbf{22.1$\pm$1.1} & 
\textbf{29.4$\pm$2.0} & 
\textbf{32.6$\pm$2.5} & 
\textbf{7.5$\pm$0.1} \\
\cmidrule(lr){1-1} \cmidrule(lr){2-4} \cmidrule(lr){5-5}
SWG \citep{deshpande2018generative} & 33.7$\pm$1.5 & 21.9$\pm$2.0 & 67.9$\pm$2.7 & - \\
CT-SWG & \textbf{25.9$\pm$ 0.9} &  \textbf{18.8 $\pm$ 1.2} & \textbf{39.0 $\pm$ 2.1} & 6.9 $\pm$ 0.1 \\
\cmidrule(lr){1-1} \cmidrule(lr){2-4} \cmidrule(lr){5-5}
MMD-GAN \citep{binkowski2018demystifying} &  39.9$\pm$0.3 & 20.6$\pm$0.3 & \textbf{32.0$\pm$0.3}  & 6.5$\pm$0.1 \\
CT-MMD-GAN &
\textbf{23.9 $\pm$ 0.4} & 
\textbf{13.8 $\pm$ 0.4} & 
38.3 $\pm$ 0.3 &
\textbf{7.4 $\pm$ 0.1} \\
\cmidrule(lr){1-1} \cmidrule(lr){2-4} \cmidrule(lr){5-5}
SNGAN \citep{miyato2018spectral} & 21.5$\pm$1.3 & 21.7$\pm$1.5 & 31.1$\pm$2.1 & {8.2$\pm$0.1} \\
CT-SNGAN
& 
{\textbf{17.2$\pm$1.0}} & 
{\textbf{9.2$\pm$1.0}} & 
{\textbf{16.8$\pm$2.1}} & 
{\textbf{8.8$\pm$0.1}} \\
\cmidrule(lr){1-1} \cmidrule(lr){2-4} \cmidrule(lr){5-5}
StyleGAN2 \citep{Karras2019stylegan2} & 5.8 & 5.2  & \textbf{2.9}  & {10.0 } \\
CT-StyleGAN2&
\textbf{2.9 $\pm$ 0.5} & 
\textbf{4.0 $\pm$ 0.7} & 
6.3 $\pm$ 0.2 &
\textbf{10.1 $\pm$ 0.1} \\
\bottomrule
\end{tabular}%
}\vspace{-2mm}
\end{table}

The summary of FID and IS on previously mentioned model is reported in Table~\ref{tab:fid}. We observe that trained with CT cost, all the models have improvements with different margin in most cases, suggesting that CT is compatible with standard GANs, SWG, MMD-GANs, WGANs and generally helps improve generation quality, especially for data with richer modalities like CIFAR-10. CT is also compatible with advanced model architecture like StyleGAN2, confirming that a better feature space could make CT more efficient to guide the generator and produce better results.

\begin{figure}[!t]
\centering
\includegraphics[width=.315\textwidth]{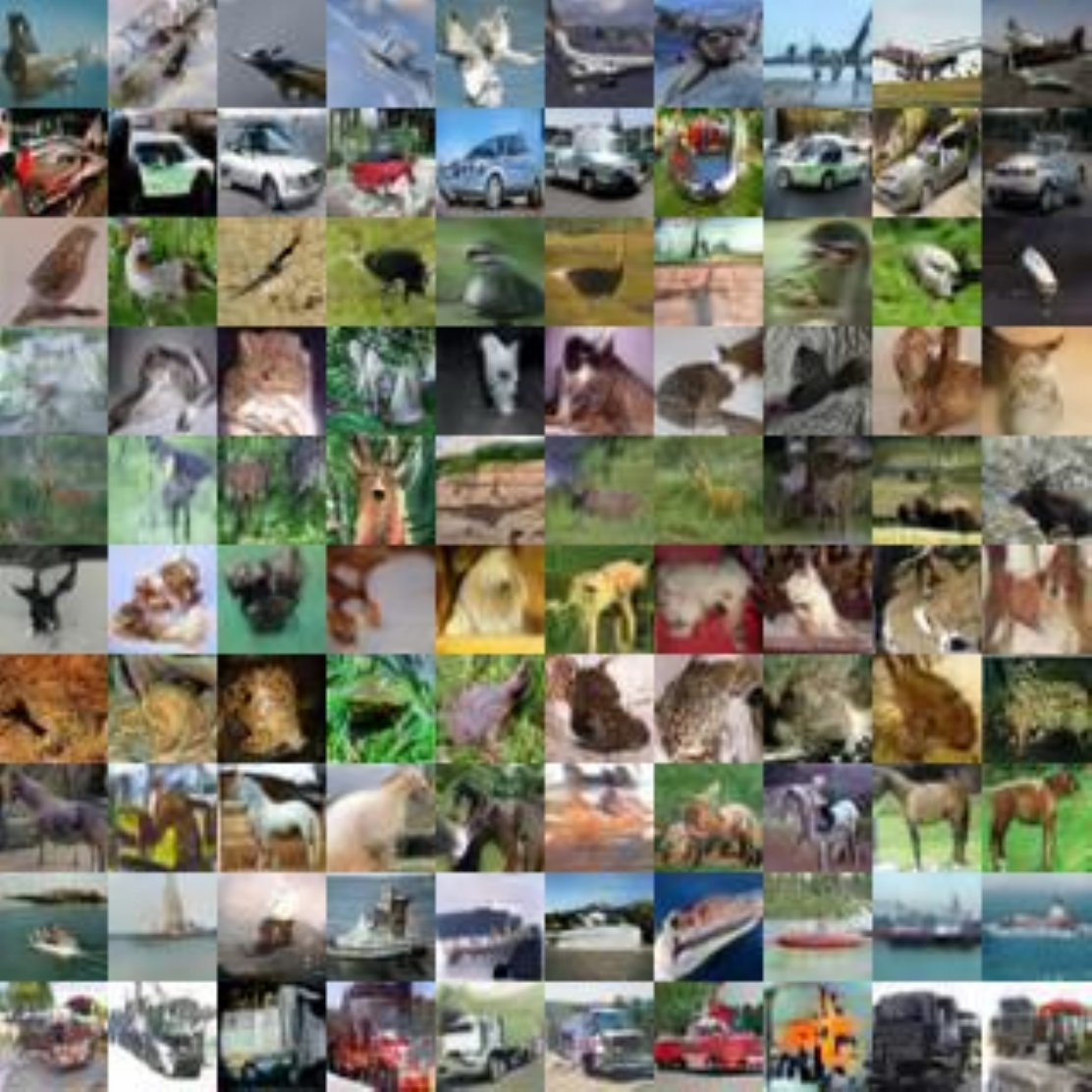}\,\,
\includegraphics[width=.315\textwidth]{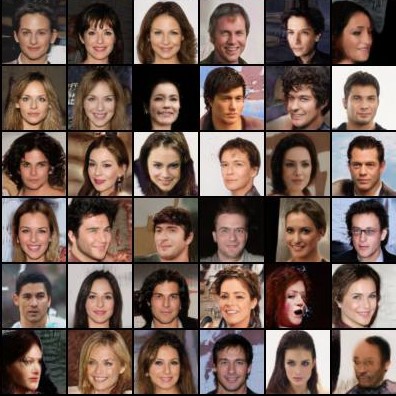}\,\,
\includegraphics[width=.315\textwidth]{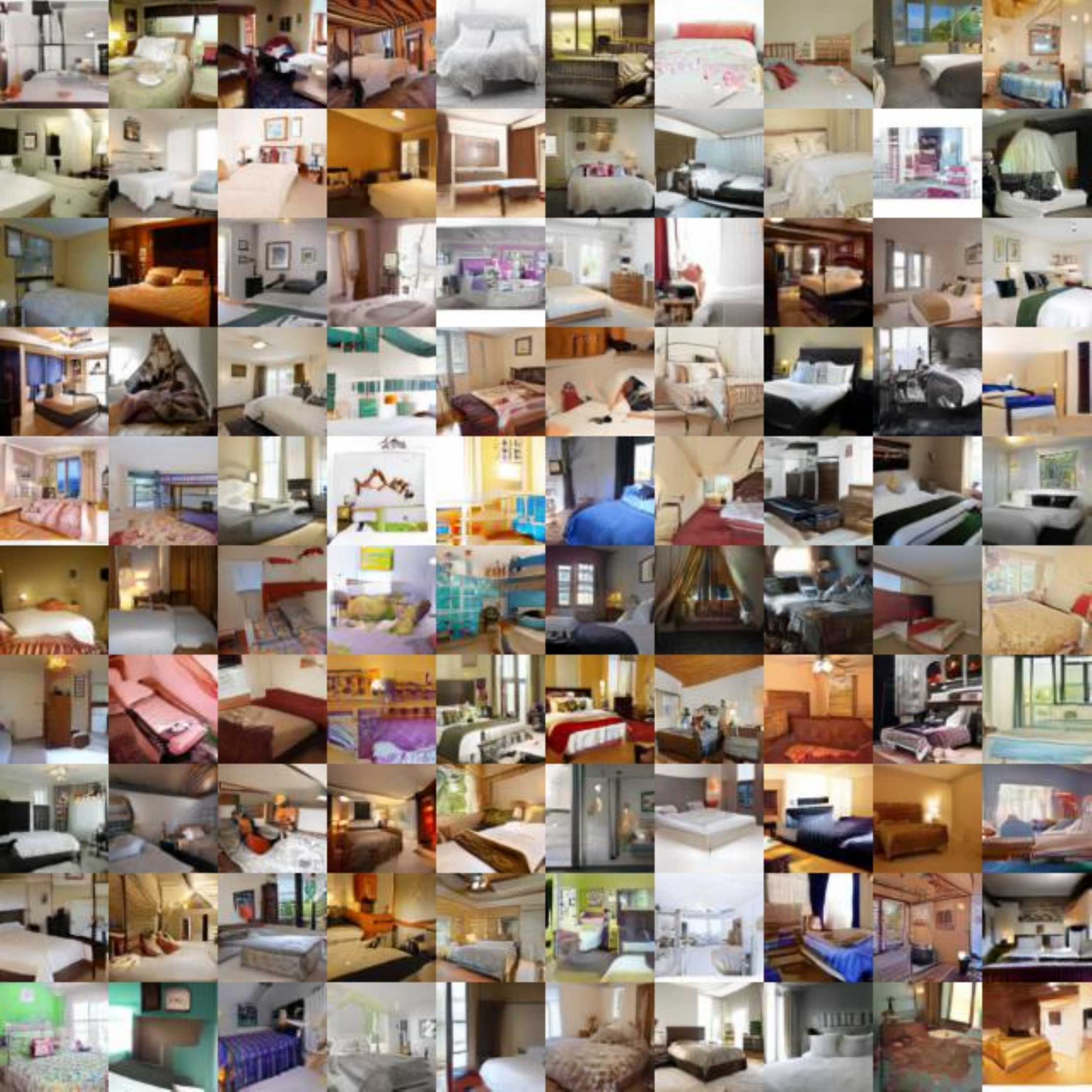}
\caption{\small Generated samples of the deep generative model that adopts the backbone of SNGAN but is optimized with the CT cost %
on CIFAR-10, CelebA, and LSUN-Bedroom. See Appendix \ref{app:results} for more results.}\label{fig:generation}
\vspace{-6mm}
\end{figure}

\begin{figure}[!t]
\centering
\begin{subfigure}[t]{0.48\textwidth}
\includegraphics[width=\textwidth]{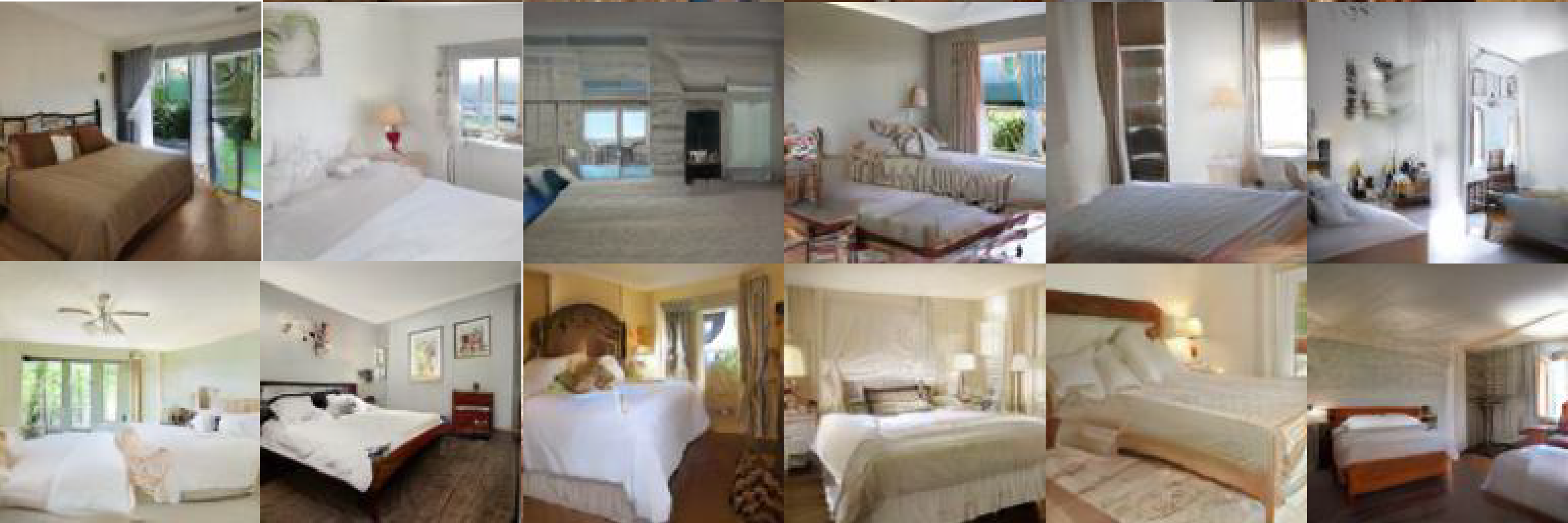}\vspace{-2mm}
\caption{\small LSUN-Bedroom(128x128)
}\label{fig:lsun_128}
\end{subfigure}
\begin{subfigure}[t]{0.48\textwidth}
\includegraphics[width=\textwidth]{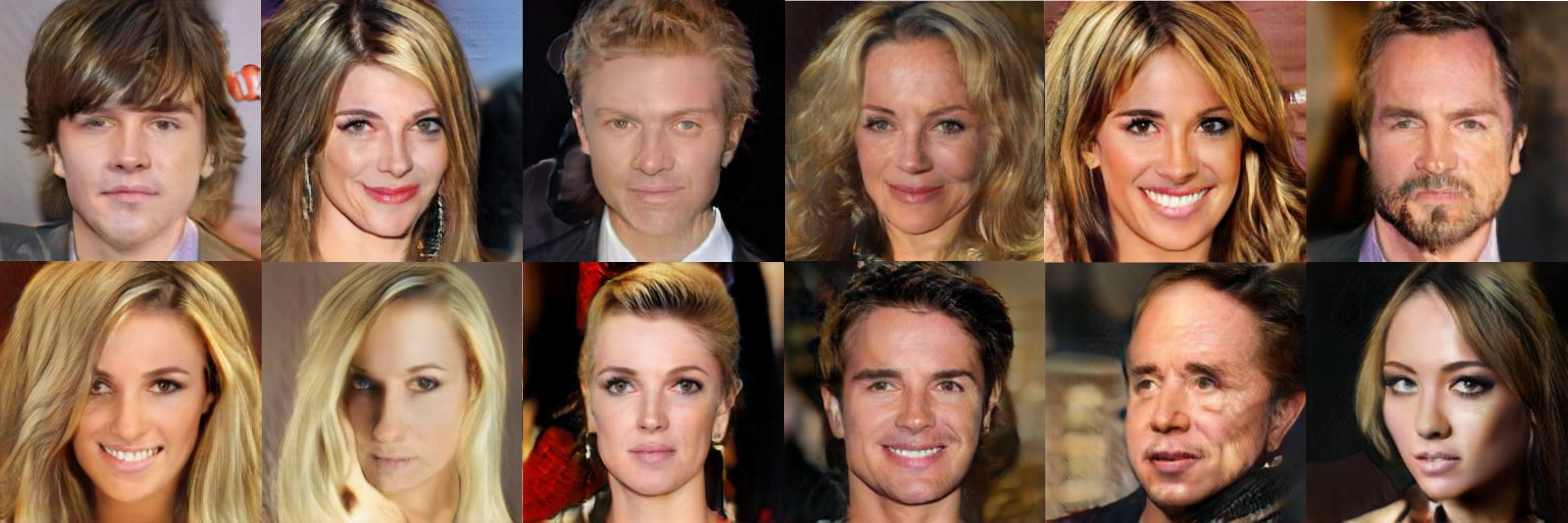}\vspace{-2mm}
\caption{\small CelebA-HQ(256x256)
}\label{fig:celebahq}
\end{subfigure}
\begin{subfigure}[t]{0.48\textwidth}
\includegraphics[width=\textwidth]{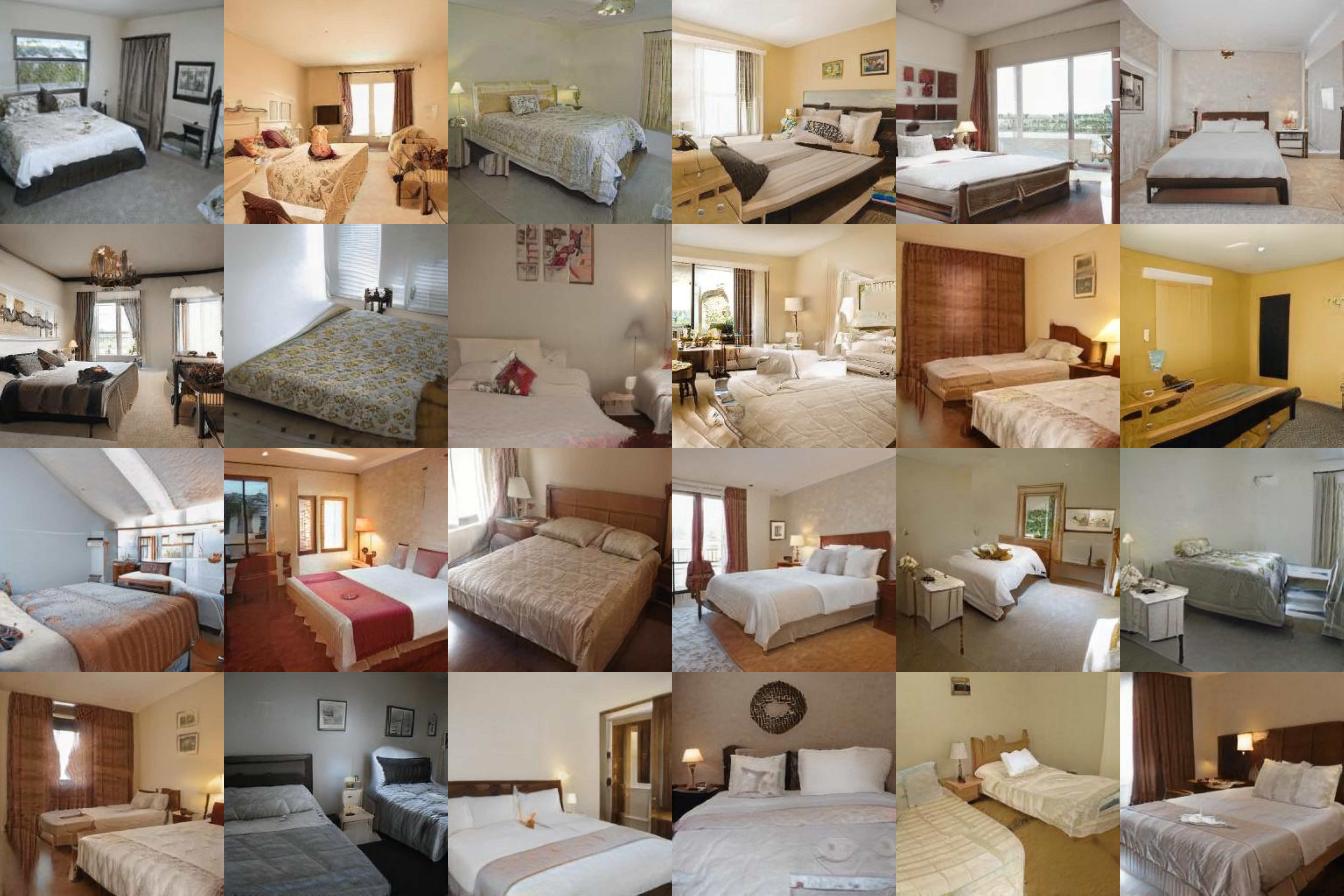}\vspace{-2mm}
\caption{\small LSUN-Bedroom(256x256)
}\label{fig:lsun_256}
\end{subfigure}
\begin{subfigure}[t]{0.48\textwidth}
\includegraphics[width=\textwidth]{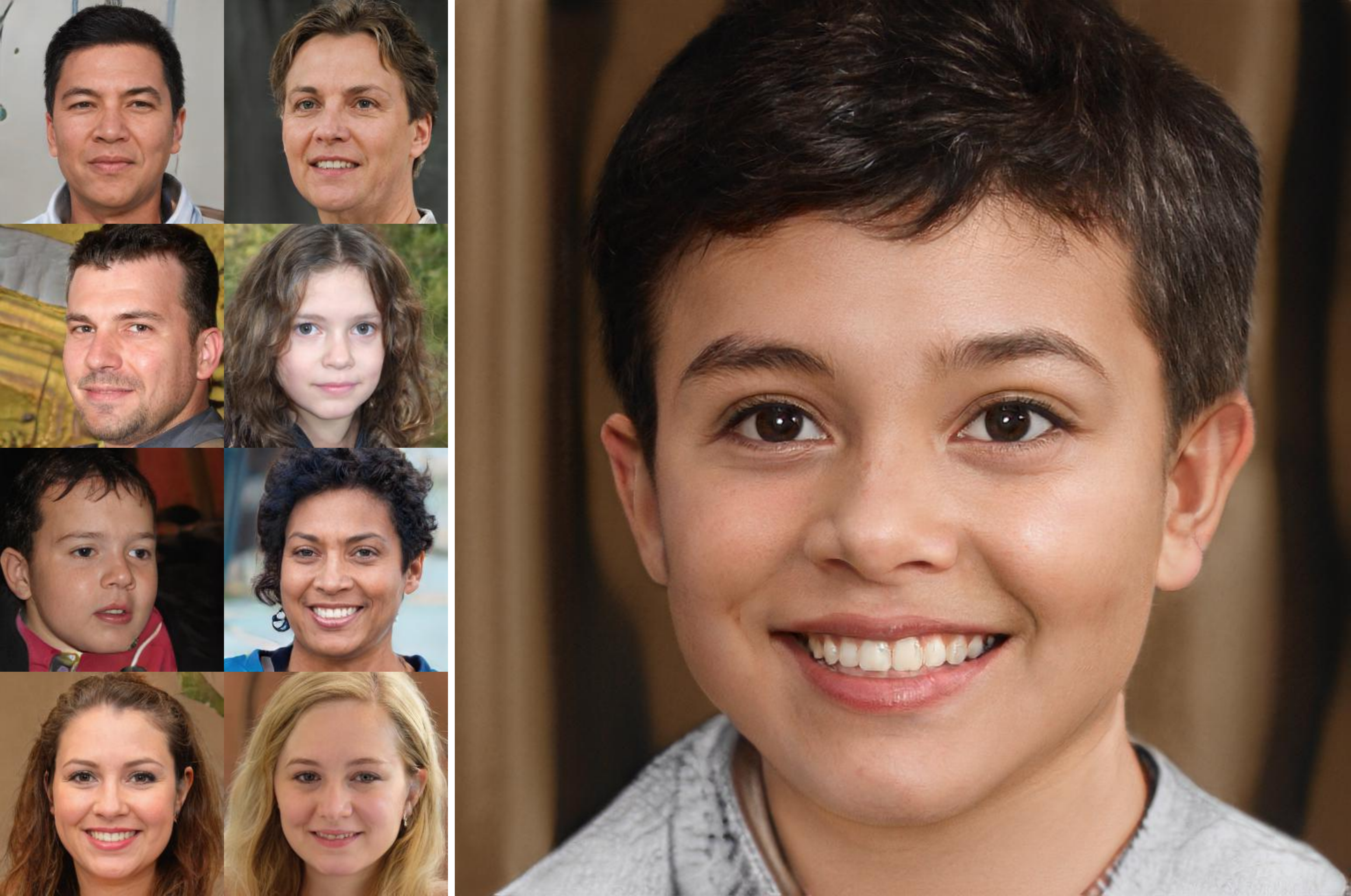}\vspace{-2mm}
\caption{\small FFHQ (256x256/1024x1024)
}\label{fig:ffhq}
\end{subfigure}\vspace{-2mm}
\caption{\small Generation results in higher-resolution cases, with SNGAN and StyleGAN2 architecture. \textit{Top}:%
LSUN-Bedroom (128x128) and CelebA-HQ (256x256), done with CT-SNGAN. \textit{Bottom}: LSUN-Bedroom (256x256) and FFHQ (256x256/1024x1024), done with CT-StyleGAN2.
}\label{fig:celebahq_sngan_vis}
\vspace{-3mm}
\end{figure}

The qualitative results shown in Fig.\,\ref{fig:generation} are consistent with quantitative results in Table \ref{tab:fid}. To additionally show how CT works for more complex generation tasks, we show in %
Fig.\,\ref{fig:celebahq_sngan_vis} %
example higher-resolution images generated by CT-SNGAN on LSUN bedroom (128x128) and CelebA-HQ (256x256), as well as images generated by CT-StyleGAN2 on LSUN bedroom (256x256), FFHQ (256x256), and FFHQ (1024x1024).

\begin{wraptable}{r}{.44\columnwidth}
\vspace{-13pt}
\caption{FID of generation results on CIFAR-10, trained with different $\rho$.}\label{tab:cifar10-rho}
\renewcommand{\arraystretch}{1.}
    \setlength{\tabcolsep}{1.0mm}{ 
    \begin{tabular}{c|ccccc}
    \toprule
        $\rho$ & 1 & 0.75 & 0.5 & 0.25 & 0 \\ 
        \midrule
        CT-DCGAN & 25.1 & 22.1 & 22.1 & \textbf{21.4} & 72.1 \\ 
        CT-SNGAN & 23.2& 17.5 & \textbf{17.2} & \textbf{17.2} & 33.2  \\ 
        \bottomrule
    \end{tabular}
    }\vspace{-14pt}
\end{wraptable}  
\textcolor{black}{\textbf{On the choice of $\rho$ 
for natural images:} In previous experiments, we fix $\rho=0.5$ by default when we prefer neither  mode-covering nor mode-seeking. We further tune $\rho$ as an additional ablation study on CIFAR-10 dataset with both the CT + DCGAN backbone and CT + SNGAN backbone to see its affects in terms of certain metrics, such as the FID score. The results shown in Table \ref{tab:cifar10-rho} suggest that CT is not sensitive to the choice of $\rho$ as long as $0<\rho<1$, and the FID score could be further improved if we choose a smaller $\rho$ to bias towards mode-seeking.
}

%
%

%
%
%
%
%
%

%
\section{Conclusion}
We propose conditional transport (CT) as a new criterion to quantify the difference between two probability distributions, via the use of both forward and backward conditional distributions. The forward and backward expected cost are respectively with respect to a source-dependent and target-dependent conditional distribution defined via Bayes' theorem. The CT cost can be approximated with discrete samples and optimized with existing stochastic gradient descent-based methods. Moreover, the forward and backward CT possess mode-covering and mode-seeking properties, respectively. By  combining them, CT nicely incorporates and balances these two properties, showing robustness in resisting mode collapse. On complex and high-dimensional data, CT is able to be calculated and stably guide the generative models in a valid feature space, which can be learned by adversarially maximizing CT or cooperatively deploying existing methods. On various benchmark datasets for deep generative modeling, we successfully train advanced models with CT. Our results consistently show improvement over the original ones, justifying  the effectiveness of the proposed CT loss.

\textbf{Discussion:} Note CT brings consistent improvement to these DGMs
by neither improving their network architectures nor %
gradient regularization. Thus it has great potential to work in conjunction with other state-of-the-art architectures or methods, such as BigGAN \citep{brock2018large}, self-attention GANs \citep{zhang2019self}, partition-guided GANs \citep{armandpour2021partition}, multimodal-DGMs \citep{Zhang2020Variational}, BigBiGAN \citep{%
donahue2019large}, self-supervised learning \citep{chen2019self}, and data augmentation \citep{karras2020training,zhao2020differentiable,zhao2020image}, which we leave for future study. 
As the paper is primarily focused on constructing and validating a new approach to quantify the difference between two probability distributions, we have focused on demonstrating the efficacy and interesting properties of the proposed CT on toy data and benchmark image data. We have focused on these previously mentioned models as the representatives in GAN, MMD-GAN, WGAN under CT, and we leave to future work using the CT to optimize more choices of DGMs, such as VAE-based models \cite{kingma2013auto} and neural-SDE \cite{song2021scorebased}.
\section*{Acknowledgments}
The authors acknowledge the support of
NSF IIS-1812699, the APX 2019 project sponsored
by the Office of the Vice President for Research at The University of Texas at Austin, the support of a gift fund from
ByteDance Inc., and the Texas Advanced Computing Center
(TACC) for providing HPC resources that have contributed
to the research results reported within this paper.

\bibliographystyle{unsrtnat}
\bibliography{reference.bib,References052016.bib}

\clearpage
\appendix
\onecolumn
\begin{center}
  \textbf{\Large Exploiting Chain Rule and Bayes' Theorem to Compare\\
  \vspace{2mm}
   Probability Distributions: Appendix}  
\end{center}

\section{Broader impact}\label{appendix: broader impact}
This paper proposes to quantify the difference between two probability distributions with conditional transport, a bidirectional cost that we exploit to balance the mode seeking and covering behaviors of a generative model.
The generative models trained with the proposed CT and datasets used in the experiments are classic in the area. Thus the capacities of these models are similar to existing ones, where we can see both positive and negative perspectives, depending on how the models are used. For example, good generative models can generate images for datasets that are expensive to collect, and be used to denoise and recover images. Meanwhile, they can also be misused to generate fake images for malicious purposes.

\section{Proof of Lemma \ref{lemma:limit}}\label{sec:proof}

\begin{proof} 
According to the strong law of large numbers, when $M\rightarrow \infty$, 
$
\frac{1}M\sum_{j=1}^M e^{-d_{\phiv}(\xv,\yv_{j})}
$,
where $\yv_j\stackrel{iid}\sim p_Y(\yv)$,
converges almost surely to $\int e^{-d_{\phiv}(\xv,\yv)}p_Y(\yv)d{\yv}$ and $
\frac{1}M\sum_{j=1}^M c(\xv,\yv_j) e^{-d_{\phiv}(\xv,\yv_{j})}
$ converges almost surely to $\int c(\xv,\yv) e^{-d_{\phiv}(\xv,\yv)}p_Y(\yv)d{\yv}$. Thus when $M\rightarrow \infty$, the term $\textstyle \sum_{j=1}^M c(\xv,\yv_j)
\hat{\pi}_M(\yv_j\given \xv,\phiv)$ in 
\eqref{eq:x2nu}
%
converges almost surely to $\frac{\int c(\xv,\yv) e^{-d_{\phiv}(\xv,\yv)}p_Y(\yv)d\yv}{\int e^{-d_{\phiv}(\xv,\yv)}p_Y(\yv)d\yv}= \int c(\xv,\yv)\pi_Y(\yv\given \xv) d\yv
$. Therefore, $\mathcal C_{\phiv,\thetav}(X\rightarrow \hat Y_M)$ defined in \eqref{eq:x2nu} converges almost surely to the forward CT cost $\mathcal C_{\phiv,\thetav}(X\rightarrow Y)$ defined in \eqref{eq:OT_E_ygivenx} when $M\rightarrow \infty$. Similarly, we can show that $\mathcal C_{\phiv,\thetav}( 
\hat X_N \leftarrow Y
) $ defined in \eqref{eq:y2mu} converges almost surely to the backward CT $\mathcal C_{\phiv,\thetav}(
X\leftarrow Y)$ defined in \eqref{eq:OT_E_xgiveny} when $N\rightarrow \infty$.

%
\end{proof}

\section{Additional details for %
the univariate normal toy example shown in \eqref{eq:1Dnormal}}\label{appendix: 1d_gaussian}
For the toy example specified in \eqref{eq:1Dnormal},
exploiting the normal-normal conjugacy, we have an analytical conditional distribution for the forward navigator as
\bas{
\pi_{\phi}(y\given x) & \propto e^{-\frac{(x-y)^2}{2 e^{\phi}}}\mathcal{N}(y;0,e^{\theta})\\
&\propto \mathcal{N}(x;y,e^{\phi})\mathcal{N}(y;0,e^{\theta})\\
&= \mathcal{N}\left(\frac{e^{\theta}}{e^{\theta}+e^{\phi}} {x},\frac{e^{\phi}e^{\theta}}{e^{\theta}+e^{\phi}}\right),
}
and an analytical conditional distribution for the backward navigator as
\bas{
\pi_{\phi}(x\given y) 
& \propto e^{-\frac{(x-y)^2}{2 e^{\phi}}}\mathcal{N}(x;0,1)\\
& \propto \mathcal{N}(y;x,e^{\phi})\mathcal{N}(x;0,1)\\
&= \mathcal{N}\left( \frac{y}{1+e^{\phi}},\frac{e^{\phi}}{1+e^{\phi}}\right).
}
Plugging them into \eqref{eq:OT_E_ygivenx} and \eqref{eq:OT_E_xgiveny}, respectively, and solving the expectations, we have
\bas{
\mathcal C_{\phi,\theta}(\mu\rightarrow \nu) 
&= \E_{x\sim\mathcal{N}(0,1)}\left[ \frac{e^{\phi}}{e^{\theta}+e^{\phi}}\left(e^{\theta}+\frac{e^{\phi}}{e^{\theta}+e^{\phi}}x^2\right)\right]\\
&= \frac{e^{\phi}}{e^{\theta}+e^{\phi}}\left(e^{\theta}+\frac{e^{\phi}}{e^{\theta}+e^{\phi}}\right), 
}
\bas{
\mathcal C_{\phi,\theta}(\mu\leftarrow \nu) 
&=
\E_{y\sim \mathcal{N}(0,e^{\theta})}\left[\frac{e^{\phi}}{1+e^{\phi}}\left(1+\frac{e^{\phi}}{1+e^{\phi}}y^2\right)\right]\\
&=\frac{e^{\phi}}{1+e^{\phi}}\left(1+\frac{e^{\phi}}{1+e^{\phi}}e^{\theta}\right). 
}

\section{Experiment details}\label{app:experiment detail}

\paragraph{Preparation of datasets} 
We apply the commonly used training set of MNIST (50K gray-scale images, $28 \times 28$
pixels) \citep{mnist}, {Stacked-MNIST (50K images, $28 \times 28$ with 3 channels
pixels)} \citep{srivastava2017veegan}, CIFAR-10 (50K color images, $32 \times 32$ pixels) \citep{cifar10}, CelebA (about 203K color images, resized to $64\times64$ pixels) \citep{celeba}, and LSUN bedrooms 
(around 3 million color images, resized to $64\times64$ pixels) \citep{lsun}. For MNIST, when calculate the inception score, we repeat the channel to convert each gray-scale image into a RGB format. For high-resolution generation, we use CelebA-HQ (30K images, resized to $256 \times 256$ pixels) \cite{CelebAMask-HQ} and FFHQ (70K images, with both original size $1024\times 1024$ and resized size $256\times 256$) \cite{karras2019style}. All image pixels are normalized to range $[-1, 1]$.

\paragraph{Experiment setups}
{To avoid  a large increase in model 
complexity, 
the navigator is parameterized as $d_{\phiv}(\xv, \yv) := d_{\phiv}((\xv - \yv) \circ (\xv - \yv))$, where $\circ$ denotes the Hadamard product, \textit{i.e.}, the element-wise product}. To be clear, we provide a Pytorch-like pseudo-code in Algorithm~\ref{alg:code_act_joint}. 
For the toy datasets, we apply the network architectures presented in Table~\ref{tab:architecture-toy}, where we set $H=100$ for generator, navigator and feature encoder. For navigator, we set input dimension $V=2$ and output dimension $d=1$. If apply a feature encoder, we have $V=2$, $d=10$ for feature encoder and $V=10$, $d=1$ for navigator. The input dimension of generator is set as 50. The slopes of all leaky ReLU functions in the networks are set to 0.1 by default. We use the the Adam optimizer \citep{adam} with learning rate $\alpha=2\times 10^{-4}$ and $\beta_1 = 0.5$, $\beta_2 = 0.99$ for the parameters of the generator, and discriminator/critic. The learning rate of navigator is divided by 5. Typically, $5,000$ training epochs are sufficient. However, our experiments show that the DGM optimized with the CT cost can be stably trained at least over $10,000$ epochs (or possibly even more if allowed to running non-stop) regardless of whether the navigators are frozen or not after a certain number of iterations, where the GAN's discriminator usually diverges long before reaching that many training epochs even if we do not freeze it after a certain number of iterations. 

For image experiments, to make the comparison fair, we strictly adopt the architecture of DCGAN \citep{radford2015unsupervised}\footnote{DCGAN architecture follows: \text{https://github.com/pytorch/examples/tree/master/dcgan}}, Sliced Wasserstein Generative model (SWG) \cite{deshpande2018generative}\footnote{SWG architecture follows: \text{https://github.com/ishansd/swg}}, MMD-GAN \citep{binkowski2018demystifying}\footnote{MMD-GAN architecture follows: \text{https://github.com/mbinkowski/MMD-GAN}}, SNGAN \citep{miyato2018spectral}\footnote{SN-GAN architecture follows: \text{https://github.com/pfnet-research/sngan\_projection}}, and StyleGAN2 \cite{Karras2019stylegan2}\footnote{StyleGAN2 architecture follows: \text{https://github.com/NVlabs/stylegan2}. We use their config-f.}, and follow their experiment setting: DCGAN and SWG apply CNN architecture on all datasets; MMD-GAN applies CNN on CIFAR-10 and ResNet architecture on other datasets; SN-GAN and StyleGAN2 apply their modified ResNet architecture. A summary of CNN and ResNet architecture is presented from Tables~\ref{tab:architecture-dcgan-cifar10}-\ref{tab:architecture-sngan-celeba-HQ}.
To adapt the navigator, we apply the backbone of the discriminator in these GAN models as feature encoder and suppose the output dimension as $m$. The navigator is an MLP with architecture shown in Table~\ref{tab:architecture-toy} by setting $V=m$, $H=512$, and $d=1$. All models are able to be trained on a single GPU, Nvidia GTX 1080-TI/Nvidia RTX 3090 in our CIFAR-10, CelebA, LSUN-bedroom experiments. For high-resolution experiments, all experiments are done on 4 Tesla-V100-16G GPUs.

\makeatletter
\AfterEndEnvironment{algorithm}{\let\@algcomment\relax}
\AtEndEnvironment{algorithm}{\kern2pt\hrule\relax\vskip3pt\@algcomment}
\let\@algcomment\relax
\newcommand\algcomment[1]{\def\@algcomment{\footnotesize#1}}
\renewcommand\fs@ruled{\def\@fs@cfont{\bfseries}\let\@fs@capt\floatc@ruled
  \def\@fs@pre{\hrule height.8pt depth0pt \kern2pt}%
  \def\@fs@post{}%
  \def\@fs@mid{\kern2pt\hrule\kern2pt}%
  \let\@fs@iftopcapt\iftrue}
\makeatother

\begin{algorithm}[ht]
\caption{PyTorch-like style pseudo-code of CT loss.}
\label{alg:code_act_joint}
\definecolor{codeblue}{rgb}{0.25,0.5,0.5}
\definecolor{codekw}{rgb}{0.85, 0.18, 0.50}
\lstset{
  backgroundcolor=\color{white},
  basicstyle=\fontsize{7.3pt}{7.3pt}\ttfamily\selectfont,
  columns=fullflexible,
  breaklines=true,
  captionpos=b,
  commentstyle=\fontsize{7.3pt}{7.3pt}\color{codeblue},
  keywordstyle=\fontsize{7.3pt}{7.3pt}\color{codekw},
}
\begin{lstlisting}[language=python]
######################## Inputs ######################
# x: data B x C x W x H;  
# y: generated samples B x C x W x H;
# netN: navigator network  d -> 1
# netD: critic network C x W x H -> d
# rho: balance coefficient of forward-backward, default = 0.5

def ct_loss(x, y, netN, netD, rho):
    ######################## compute cost ######################
    f_x = netD(x) # feature of x: B x d
    f_y = netD(y) # feature of y: B x d
    cost = torch.norm(f_x[:,None] - f_y, dim=-1).pow(2) # pairwise cost: B x B
    
    ######################## compute transport map ######################
    mse_n = (f_x[:,None] - f_y).pow(2) # pairwise mse for navigator network: B x B x d
    d = netN(mse_n).squeeze().mul(-1) # navigator distance: B x B
    forward_map = torch.softmax(d, dim=1) # forward map is in y wise
    backward_map = torch.softmax(d, dim=0) # backward map is in x wise
    
    ######################## compute CT loss ######################
    # element-wise product of cost and transport map
    ct = rho * (cost * forward_map).sum(1).mean() + (1-rho) * (cost * backward_map).sum(0).mean() 
    return ct
\end{lstlisting}
\end{algorithm}

\begin{table}[ht]
\centering
\caption{Network architecture for toy datasets ($V$, $H$ and $d$ indicate the dimensionality).} \vspace{-2mm}
\label{tab:architecture-toy}
\begin{subtable}[t]{.4\textwidth}
\caption{Generator $G_{\thetav}$}\vspace{-2mm}
\centering
\begin{tabular}{c}
\toprule 
$\epsilonv \in \mathbb{R}^{50} \sim \mathcal{N}(0,1)$ \\ \cmidrule(lr){1-1}
$50 \rightarrow H$, dense, BN, lReLU \\ \cmidrule(lr){1-1}
$H \rightarrow \lfloor \frac{H}{2} \rfloor$, dense, BN, lReLU \\ \cmidrule(lr){1-1}
$\lfloor \frac{H}{2} \rfloor \rightarrow V$, dense, linear \\ 
\bottomrule
\end{tabular}
\end{subtable}\hfill
\begin{subtable}[t]{.4\textwidth}
\caption{Navigator $d_{\phiv}$ / Feature encoder $\mathcal{T}_{\etav}$ }\vspace{-2mm}
\centering
\begin{tabular}{c}
\toprule 
$\xv \in \mathbb{R}^{V}$ \\ \cmidrule(lr){1-1}
$V \rightarrow H$, dense, BN, lReLU \\ \cmidrule(lr){1-1}
$H \rightarrow \lfloor \frac{H}{2} \rfloor$, dense, BN, lReLU \\ \cmidrule(lr){1-1}
$\lfloor \frac{H}{2} \rfloor \rightarrow d$, dense, linear \\ 
\bottomrule
\end{tabular}
\end{subtable}\vspace{-2mm}
\end{table}

\newpage
\section{Supplementary experiment results}\label{app:results}

\subsection{Results of 2D toy datasets and robustness in adversarial feature extraction}\label{appendix:2d_toy}
\begin{figure}[!ht]
\centering
\includegraphics[width=.9\textwidth]{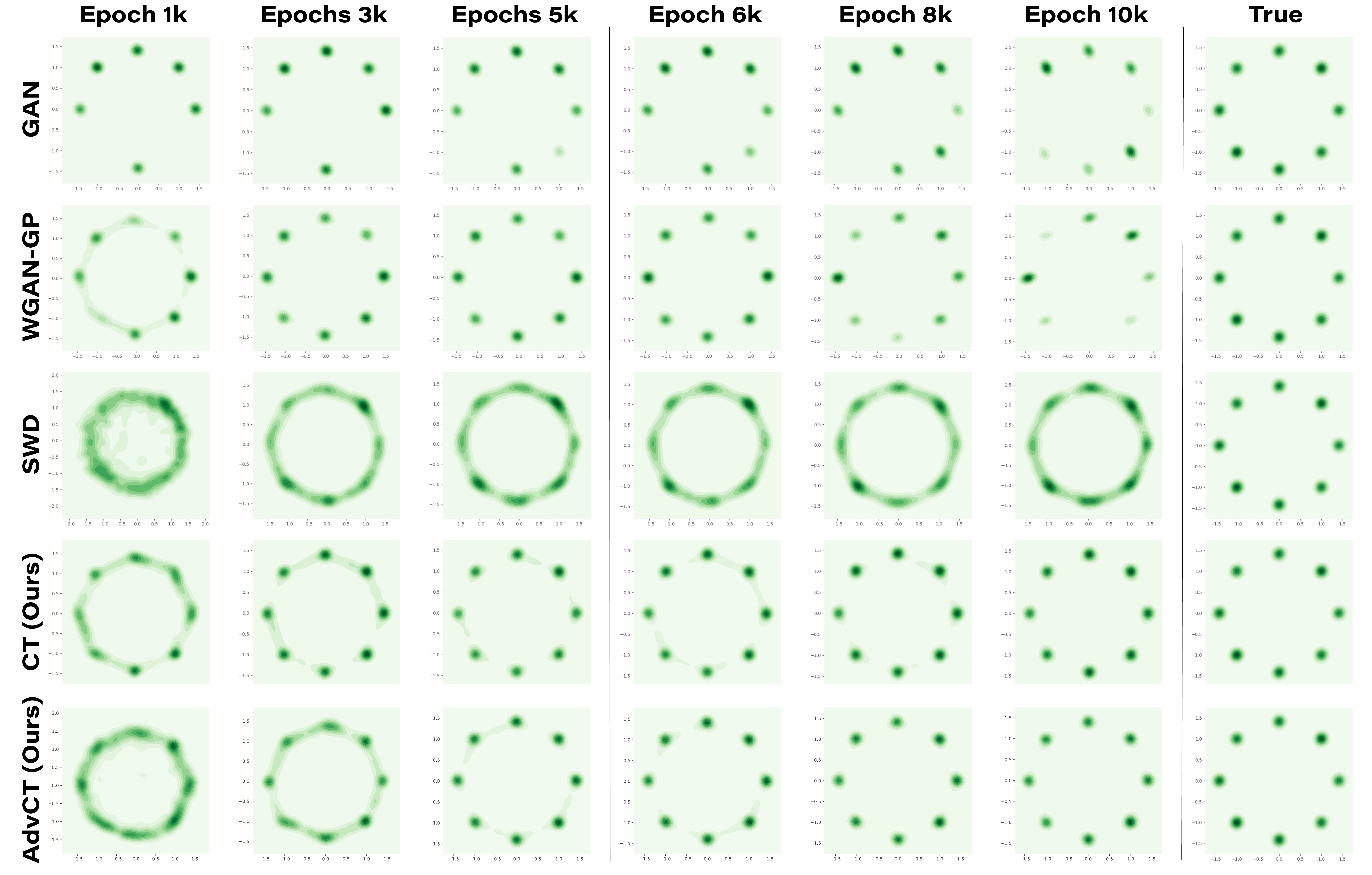}%
\caption{%
On a 8-Gaussian mixture data, comparison of generation quality and training stability between two min-max deep generative models (DGMs), including vallina GAN and Wasserstein GAN with gradient penalty (WGAN-GP), and two min-max-free DGMs, whose generators are trained under the sliced Wasserstein distance (SWD) and the proposed CT cost, respectively. The critics of GAN, WGAN-GP, the navigators of CT and the adversarially trained feature encoders of AdvCT are fixed after $5k$ training epochs. The last column shows the true data density.}\label{fig:2d_8gaussian} %
\end{figure}

\begin{figure}[!ht]
\centering
\includegraphics[width=.9\textwidth]{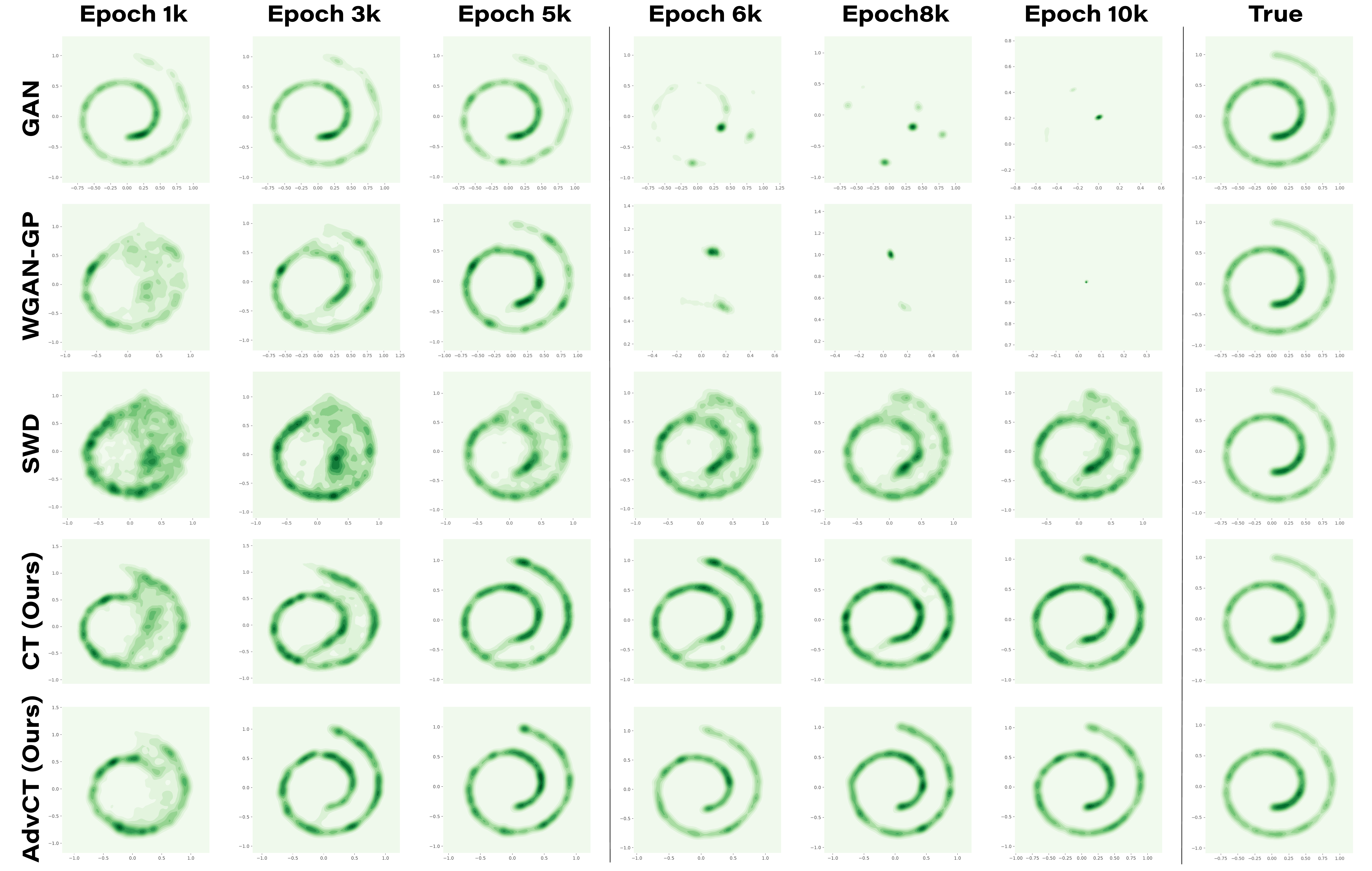}
\caption{
Analogous plot to Fig. \ref{fig:2d_8gaussian} for the Swiss-Roll dataset.
}\label{fig:2d_swiss_roll}
\end{figure}

\begin{figure}[!ht]
\centering
\includegraphics[width=.9\textwidth]{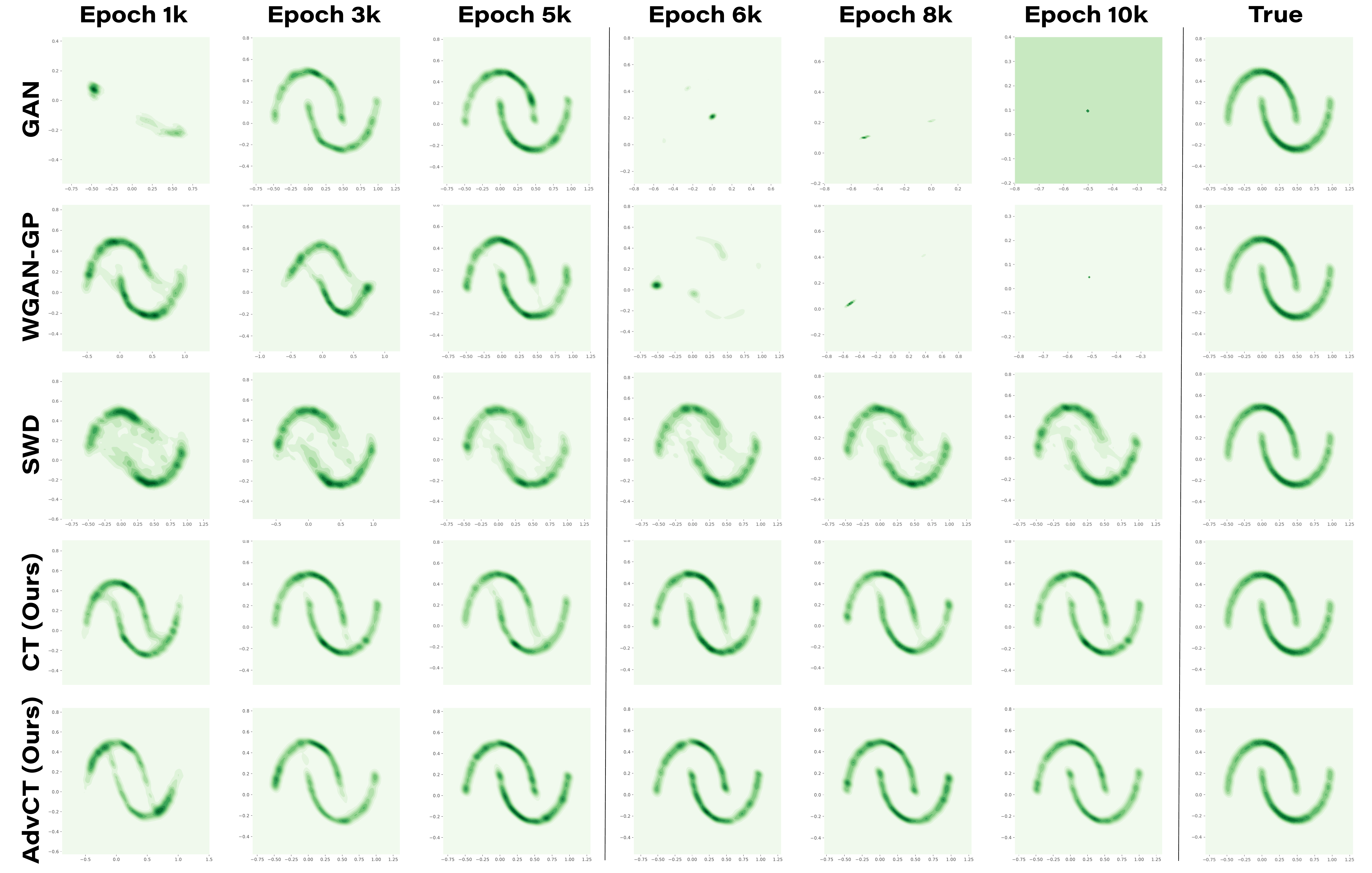}
\caption{
Analogous plot to Fig. \ref{fig:2d_8gaussian} for the Half-Moon dataset.
}\label{fig:2d_half_moon}
\end{figure}

\begin{figure}[!ht]
\centering
\includegraphics[width=.9\textwidth]{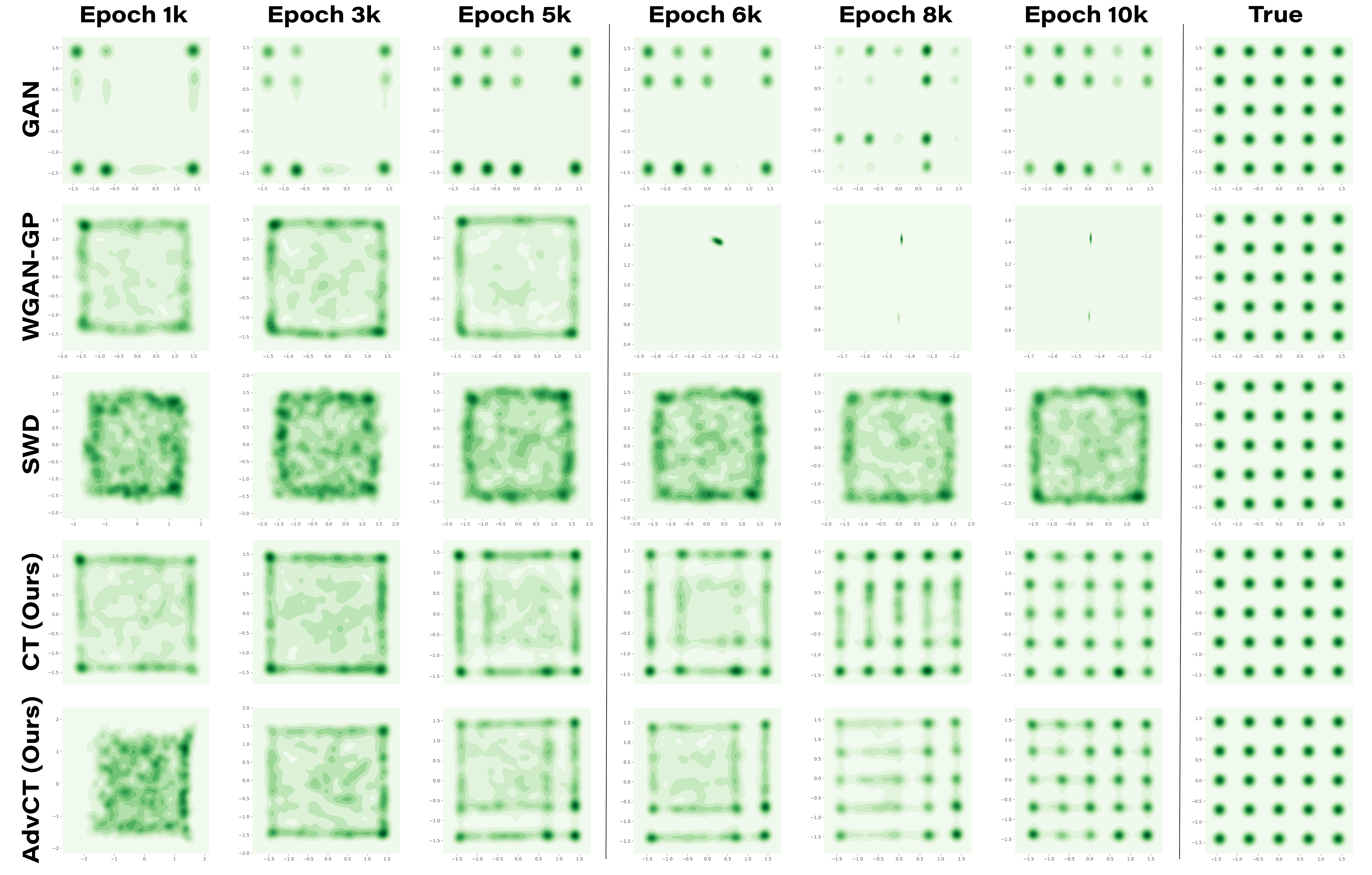}
\caption{
Analogous plot to Fig. \ref{fig:2d_8gaussian} for the 25-Gaussian mixture dataset.
}\label{fig:2d_25gaussian}
\end{figure}

We visualize the results on the 8-Gaussian mixture toy dataset and other three commonly-used 2D toy datasets: Swiss-Roll, Half-Moon and 25-Gaussian mixture. As shown in Figs.~\ref{fig:2d_8gaussian}-\ref{fig:2d_25gaussian}, in the first 5k epochs, all DGMs are normally trained and the generative distributions are getting close to the true data distribution, while on 8-Gaussian and 25-Gaussian data, Vanilla GANs show mode missing behaviors. After 5k epochs, as the discriminator/navigator/feature encoder components in all DGMs are fixed, we can observe GAN and WGAN that solve min-max loss appear to collapse. This mode collapse issue of both GAN and WGAN-GP becomes more severe on the Swiss-Roll, Half-Moon, and 25-Gaussian datasets, since they rely on an optimized discriminator/critic to guide the generator. SWG relies on the slicing projection and is not affected, while its generated samples only cover the modes and ignore the correct density, indicating the effectiveness of slicing methods rely on the slicing \cite{kolouri2019generalized}. The proposed CT cost show consistent good performance on the fitting of all these toy datasets, even after the navigator and the feature encoder are fixed after 5k epochs. This justifies our analysis about the robustness of CT cost.

\clearpage
\subsection{Additional results of cooperative \textit{vs.} adversarial encoder training}\label{appendix:co-adv training}

\begin{figure}[ht!]
\centering
\begin{subfigure}[t]{0.20\textwidth}
    \includegraphics[width=\textwidth]
    {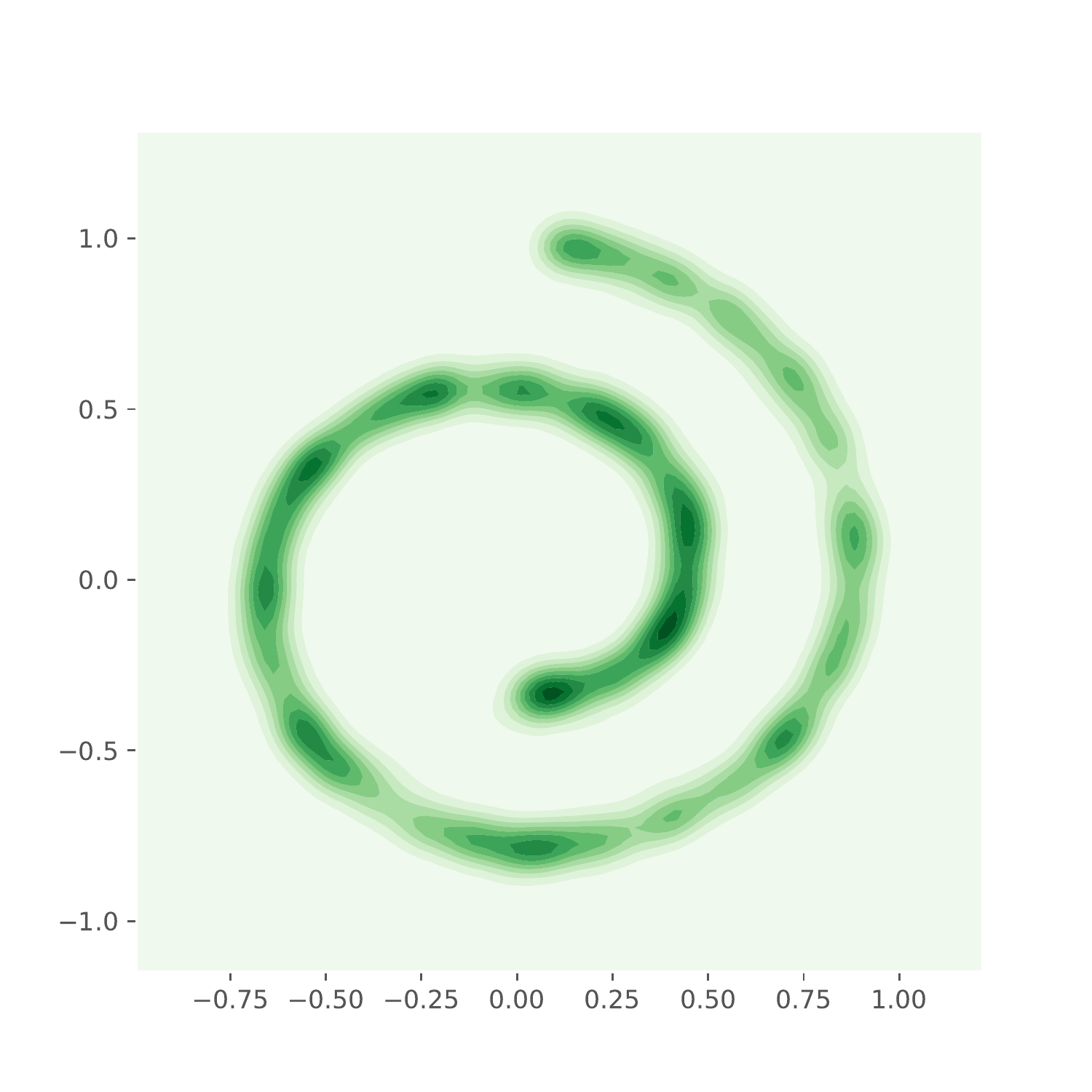}\vspace{-2mm}
\end{subfigure}\vrule\hfill
\begin{subfigure}[t]{0.19\textwidth}
    \includegraphics[width=\textwidth]
    {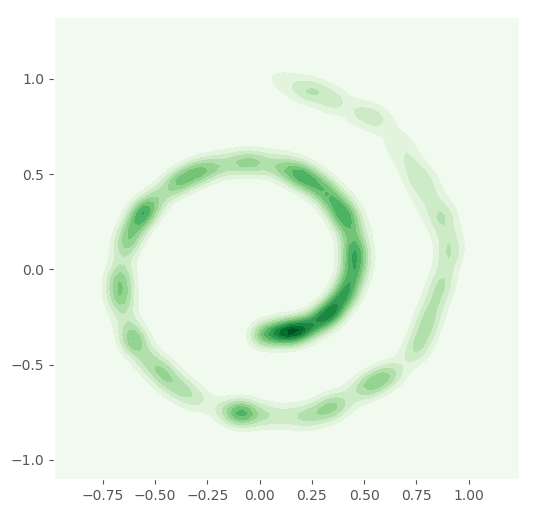}\vspace{-2mm}
\end{subfigure}\hfill
\begin{subfigure}[t]{0.20\textwidth}
    \includegraphics[width=\textwidth]
    {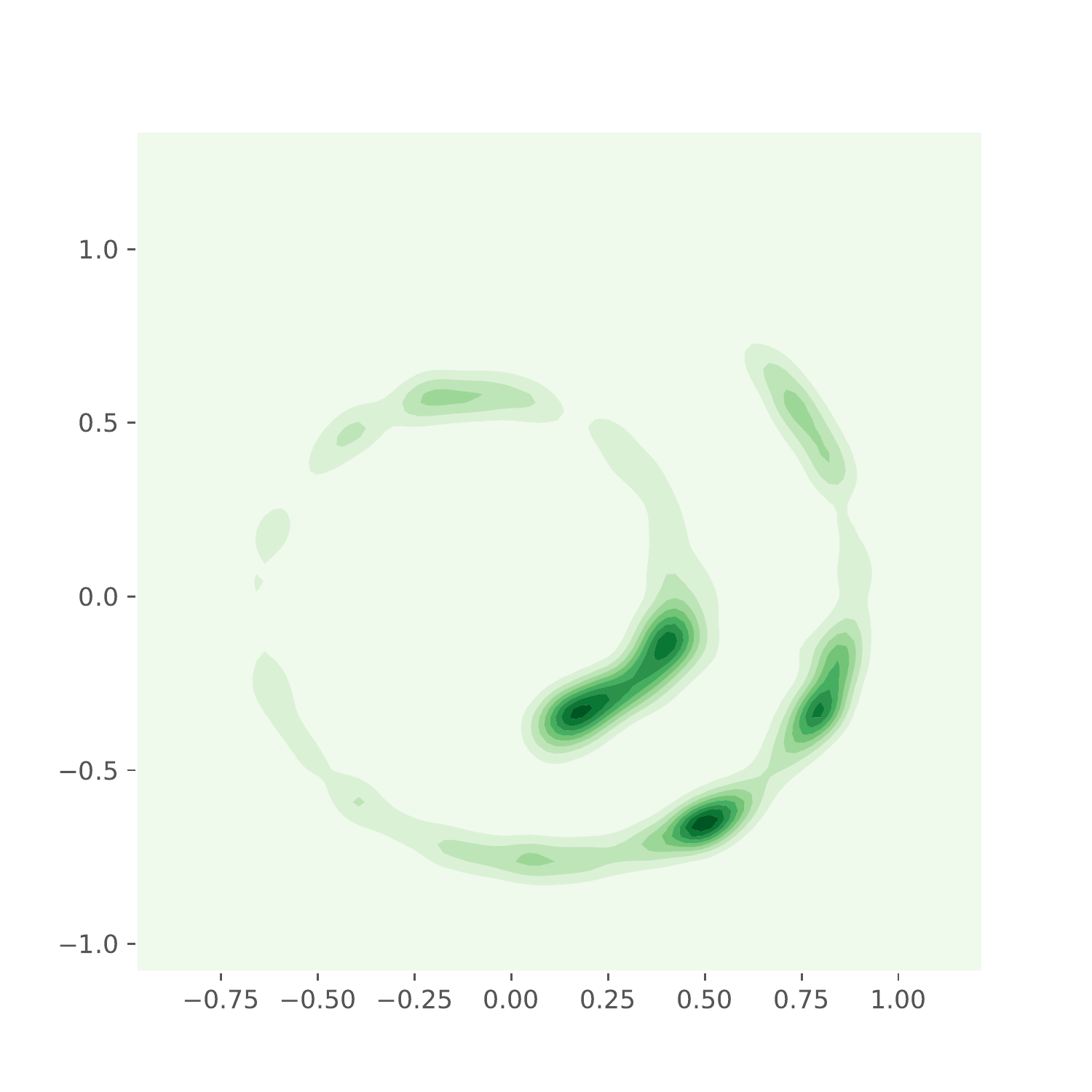}\vspace{-2mm}
\end{subfigure}\vrule\hfill
\begin{subfigure}[t]{0.19\textwidth}
\includegraphics[width=\textwidth]
{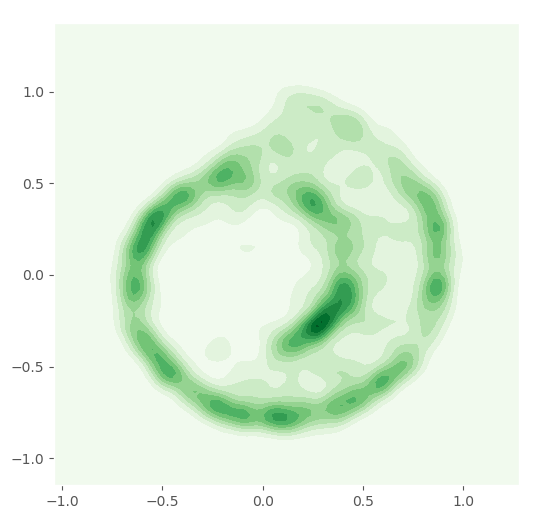}\vspace{-2mm}
\end{subfigure}\hfill
\begin{subfigure}[t]{0.20\textwidth}
\includegraphics[width=\textwidth]
{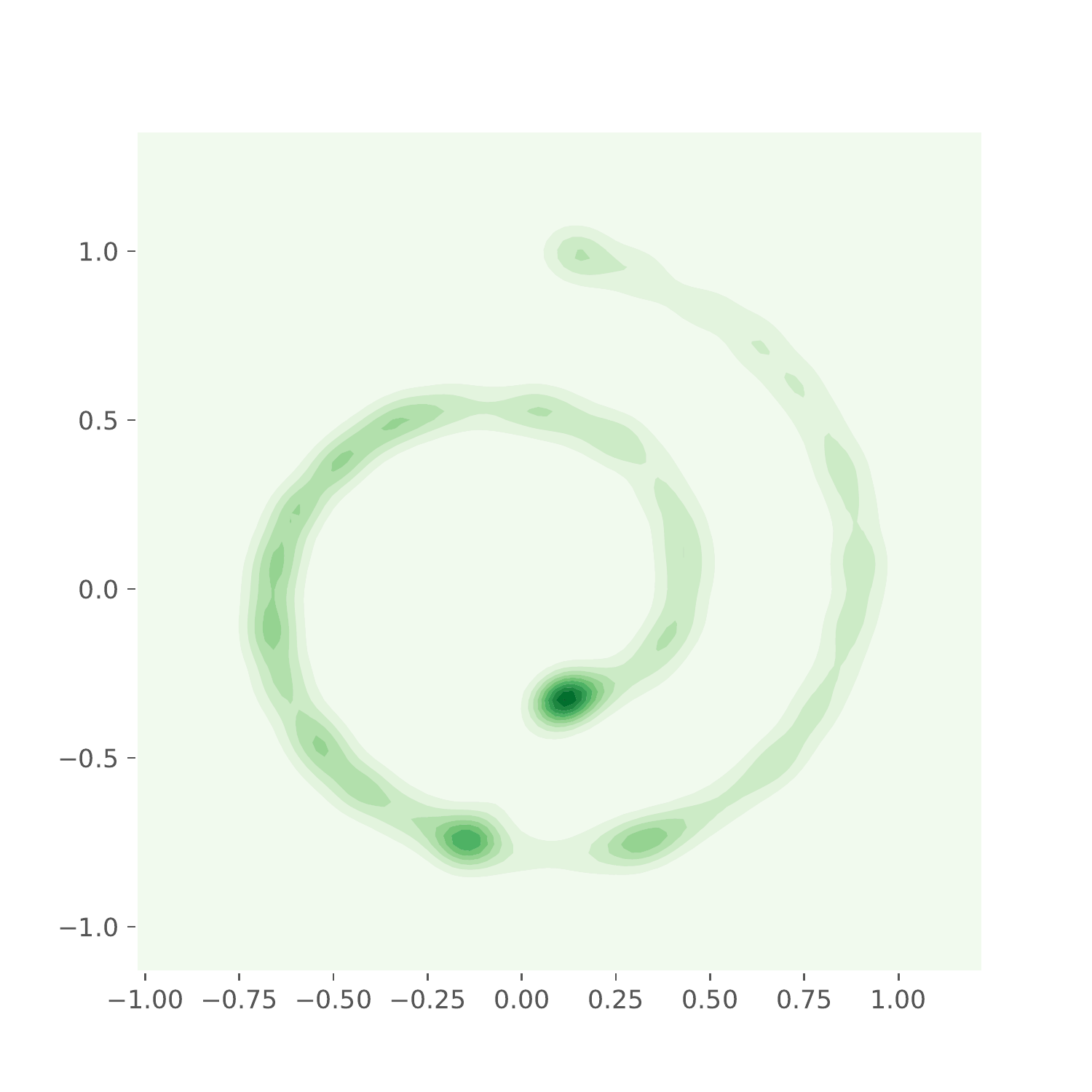}\vspace{-2mm}
\end{subfigure}
\begin{subfigure}[t]{0.20\textwidth}
    \includegraphics[width=\textwidth]
    {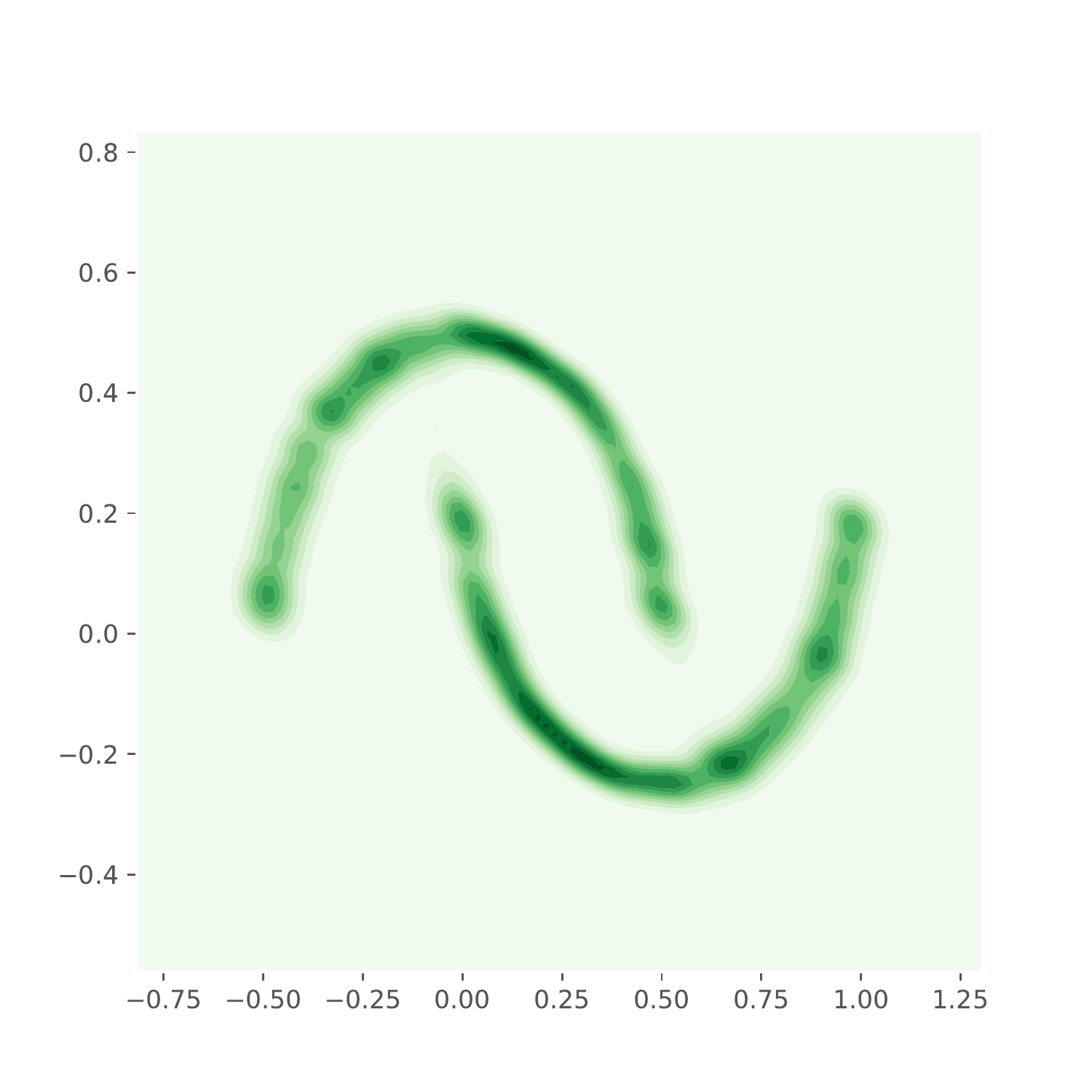}\vspace{-2mm}
\end{subfigure}\vrule\hfill
\begin{subfigure}[t]{0.19\textwidth}
    \includegraphics[width=\textwidth]
    {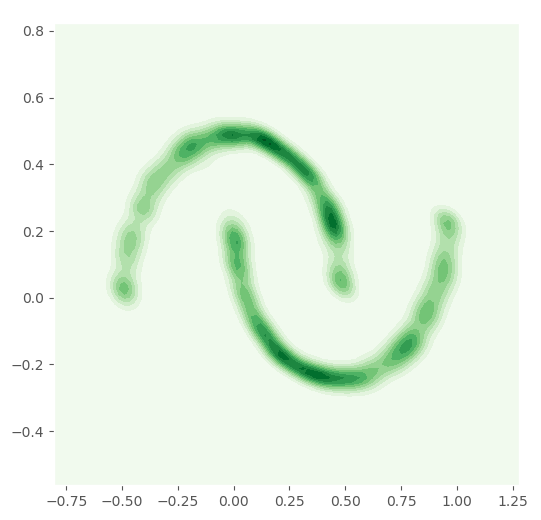}\vspace{-2mm}
\end{subfigure}\hfill
\begin{subfigure}[t]{0.20\textwidth}
    \includegraphics[width=\textwidth]
    {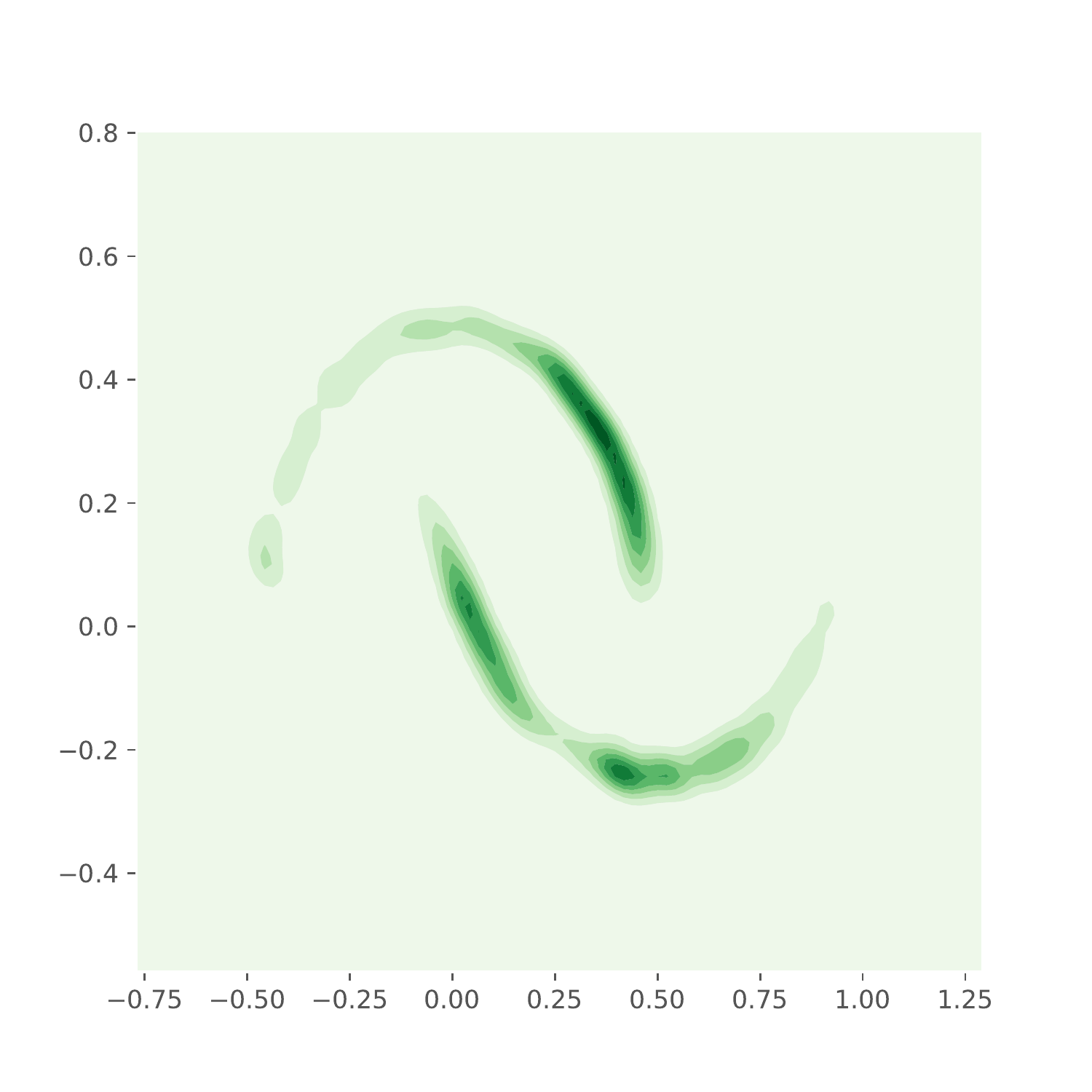}\vspace{-2mm}
\end{subfigure}\vrule\hfill
\begin{subfigure}[t]{0.19\textwidth}
\includegraphics[width=\textwidth]
{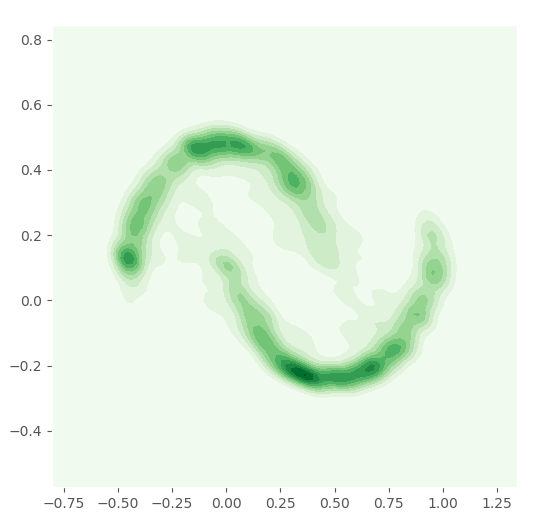}\vspace{-2mm}
\end{subfigure}\hfill
\begin{subfigure}[t]{0.20\textwidth}
\includegraphics[width=\textwidth]
{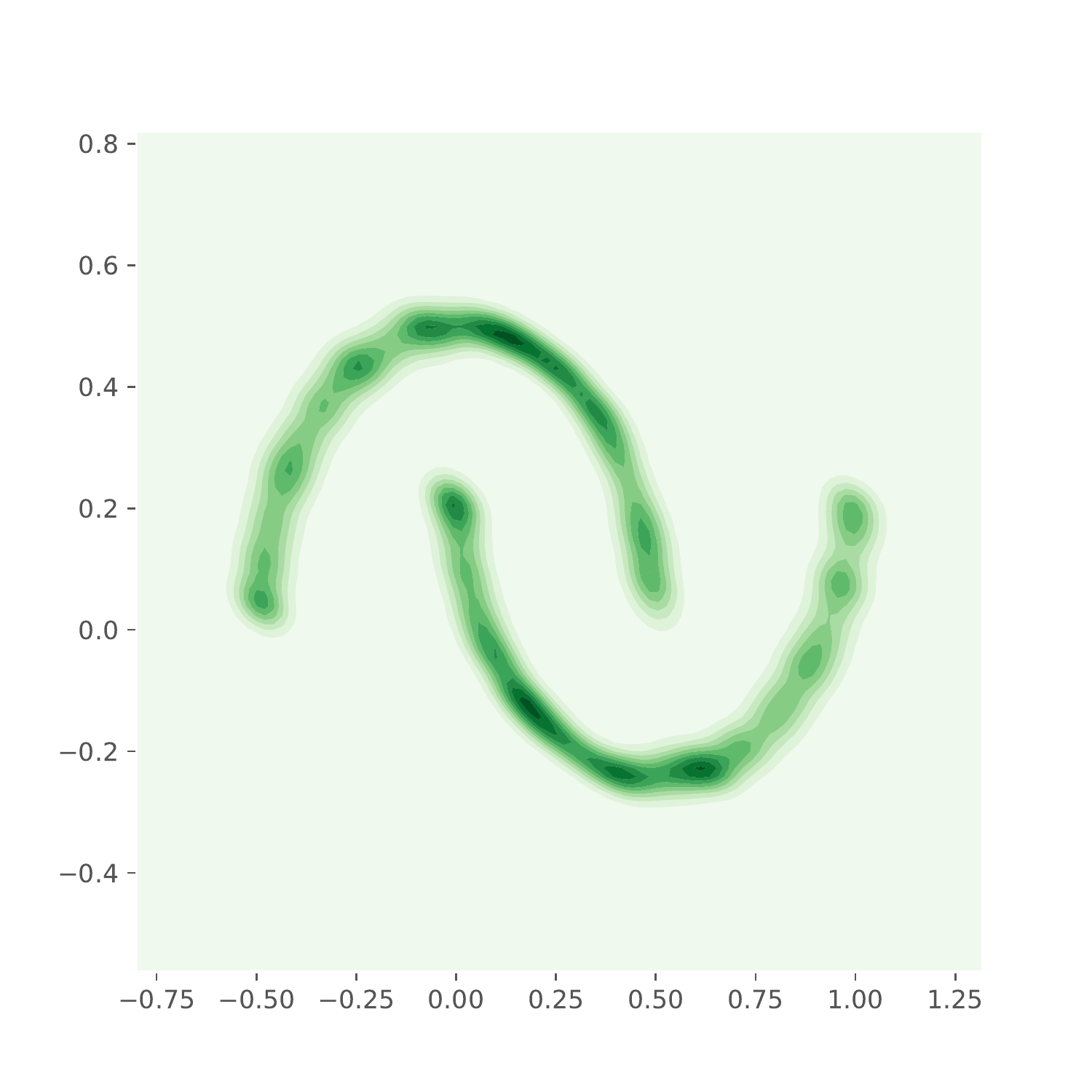}\vspace{-2mm}
\end{subfigure}
\begin{subfigure}[t]{0.20\textwidth}
    \includegraphics[width=\textwidth]
    {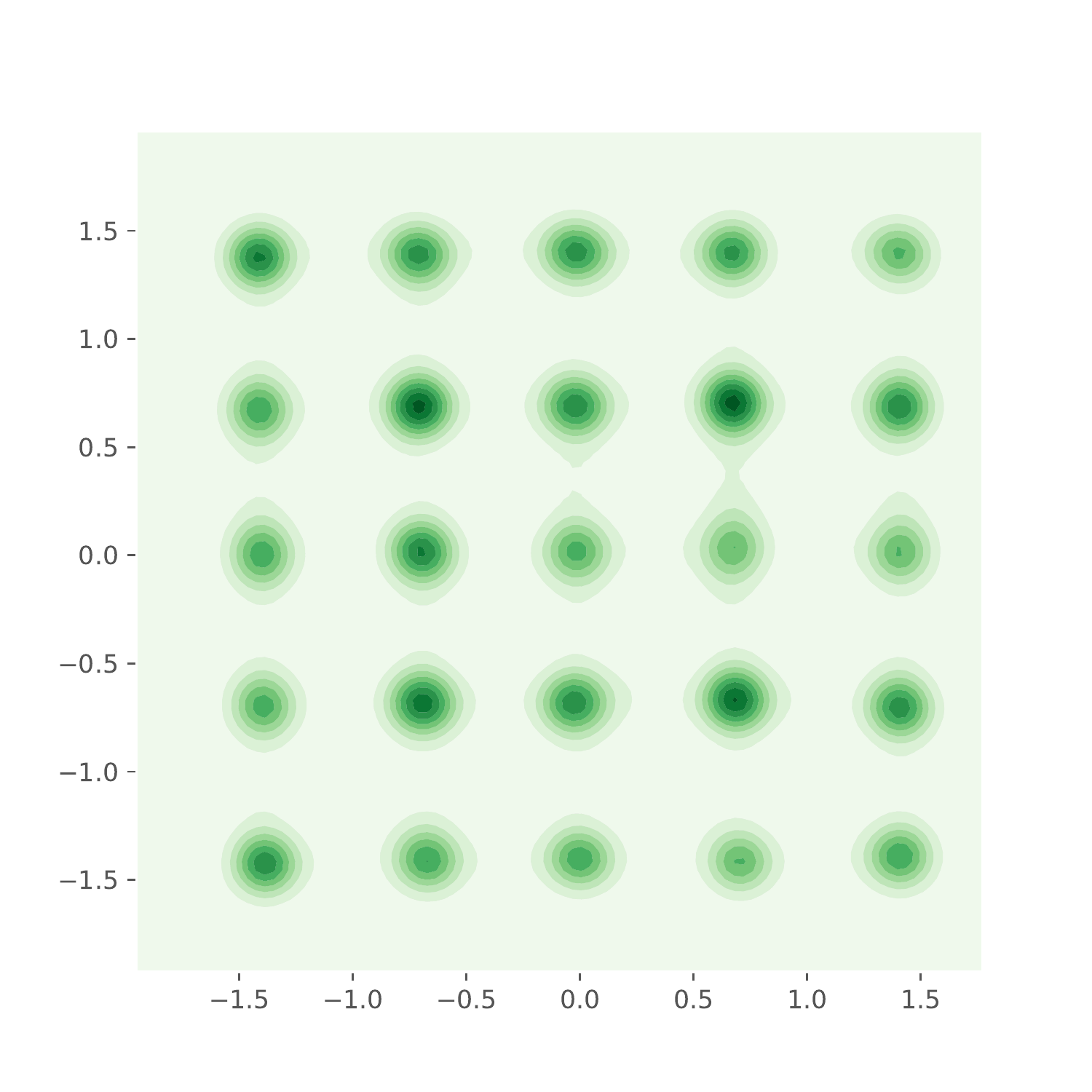}\vspace{-2mm}
\caption{\small Adv CT.
}\label{fig:minmax_CT_appendix}
\end{subfigure}\vrule\hfill
\begin{subfigure}[t]{0.19\textwidth}
    \includegraphics[width=\textwidth]
    {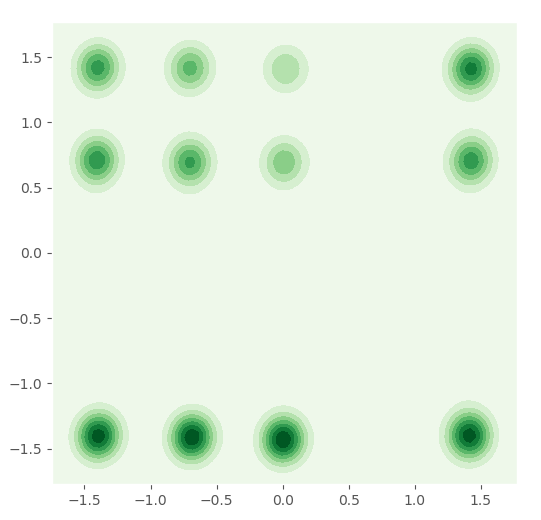}\vspace{-2mm}
\caption{\small GAN.
}\label{fig:min_gan_appendix}
\end{subfigure}\hfill
\begin{subfigure}[t]{0.20\textwidth}
    \includegraphics[width=\textwidth]
    {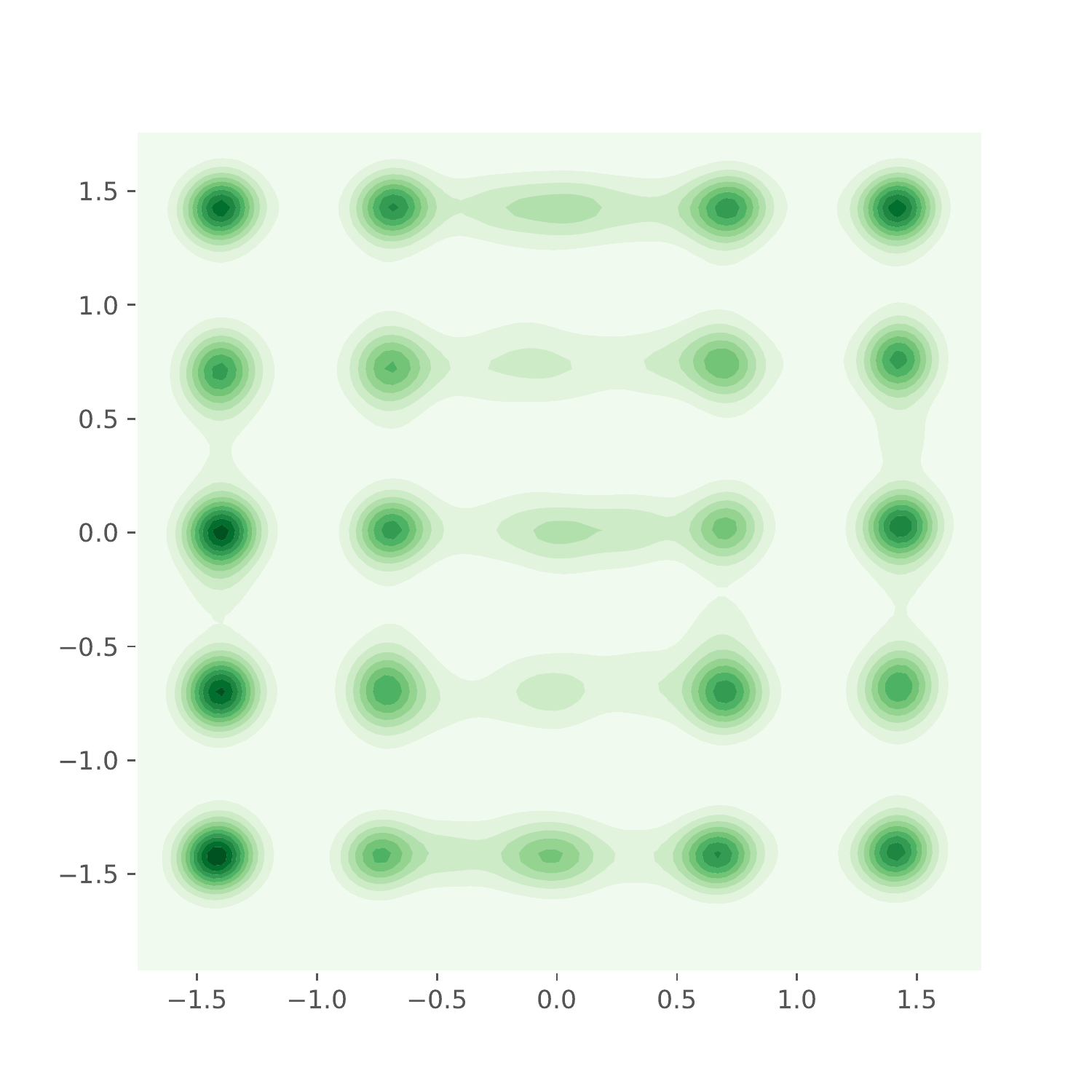}\vspace{-2mm}
\caption{\small $\mathcal{L}_D$ + CT.
}\label{fig:min_CT_max_d_appendix}
\end{subfigure}\vrule\hfill
\begin{subfigure}[t]{0.19\textwidth}
\includegraphics[width=\textwidth]
{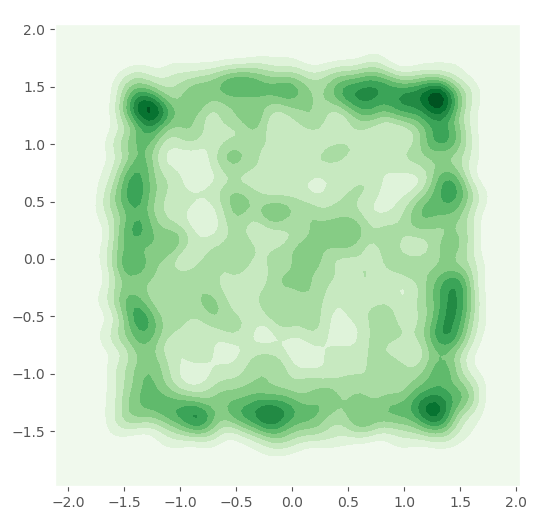}\vspace{-2mm}
\caption{\small SWD.
}\label{fig:min_slice_appendix}
\end{subfigure}\hfill
\begin{subfigure}[t]{0.20\textwidth}
\includegraphics[width=\textwidth]
{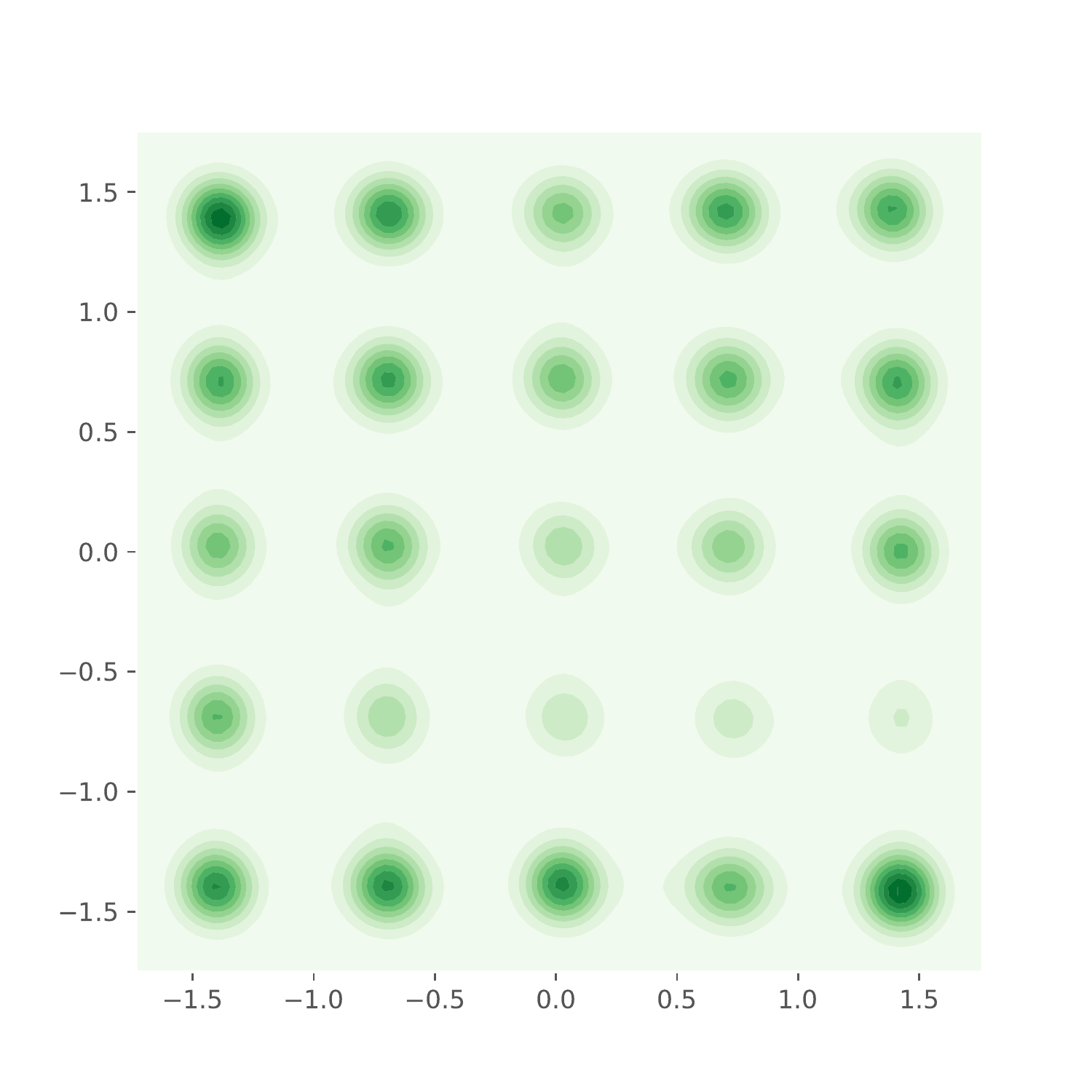}\vspace{-2mm}
\caption{\small Slicing + CT.
}\label{fig:min_slicedCT_appendix}
\end{subfigure}\vspace{-3mm}
\caption{\small Analogous plot to Fig.~\ref{fig:ablation_coop} on Swiss roll, half-moon and 25 Gaussians datasets. Ablation of fitting results by minimizing CT in different spaces
}\label{fig:ablation_coop_appendix}\vspace{-1mm}
\end{figure}

\begin{figure}[ht]
\centering
\begin{subfigure}[t]{0.23\textwidth}
    \includegraphics[width=\textwidth]{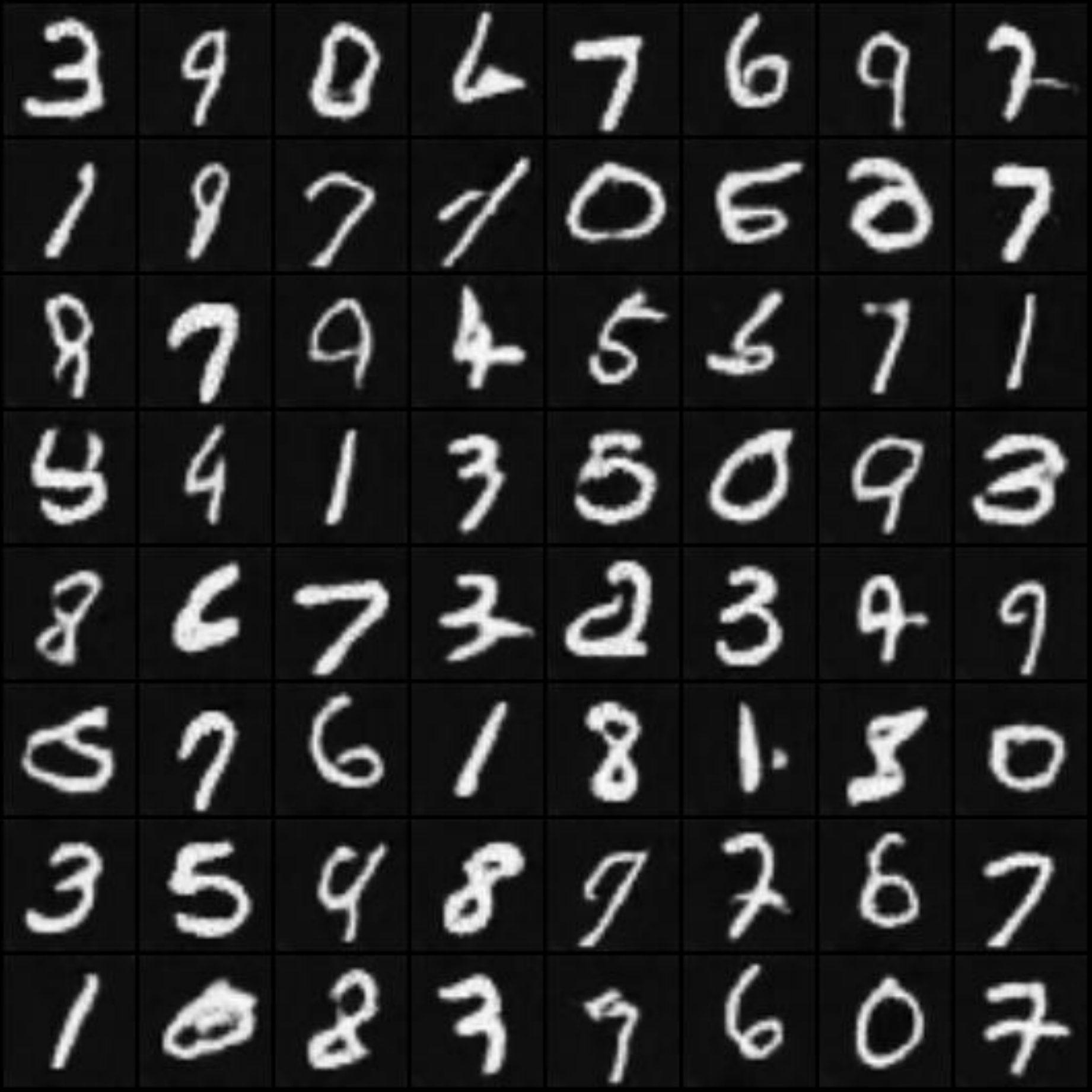}\vspace{-2mm}
\end{subfigure}\hfill
\begin{subfigure}[t]{0.23\textwidth}
    \includegraphics[width=\textwidth]{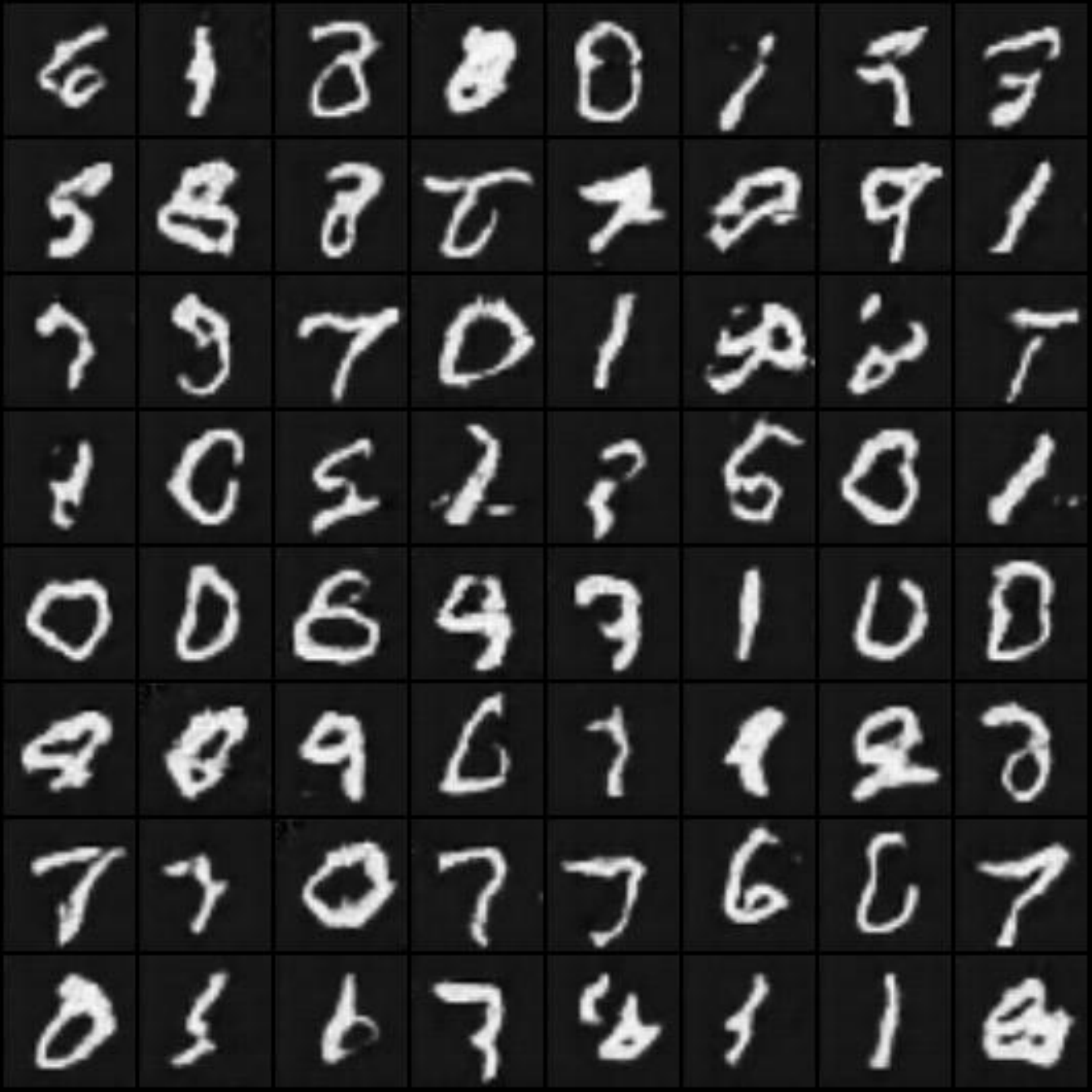}\vspace{-2mm}
\end{subfigure}\hfill
\begin{subfigure}[t]{0.23\textwidth}
    \includegraphics[width=\textwidth]{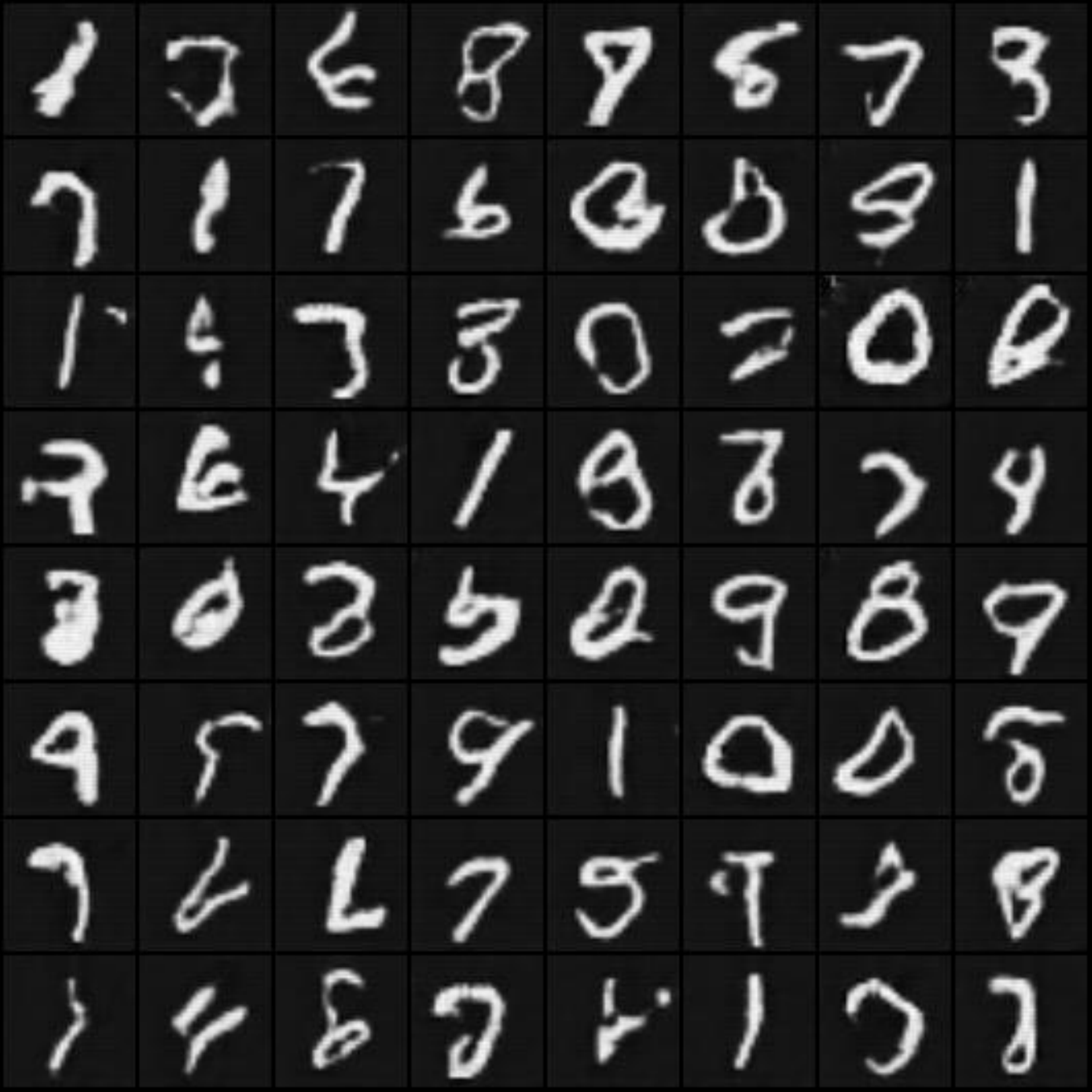}\vspace{-2mm}
\end{subfigure}\hfill
\begin{subfigure}[t]{0.23\textwidth}
    \includegraphics[width=\textwidth]{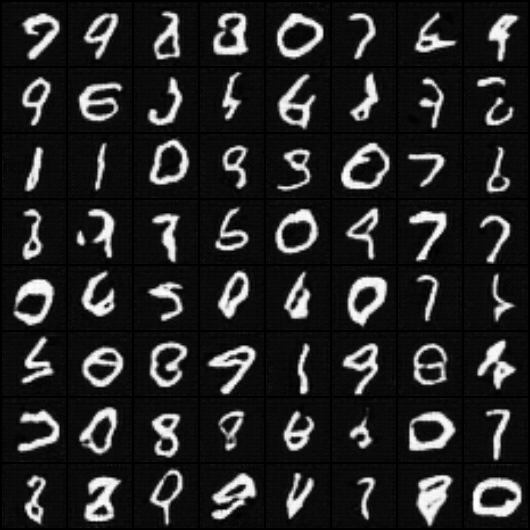}\vspace{-2mm}
\end{subfigure}
\begin{subfigure}[t]{0.23\textwidth}
    \includegraphics[width=\textwidth]{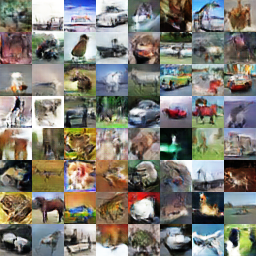}\vspace{-2mm}
\end{subfigure}\hfill
\begin{subfigure}[t]{0.23\textwidth}
    \includegraphics[width=\textwidth]{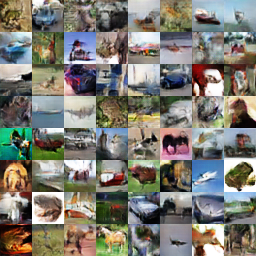}\vspace{-2mm}
\end{subfigure}\hfill
\begin{subfigure}[t]{0.23\textwidth}
    \includegraphics[width=\textwidth]{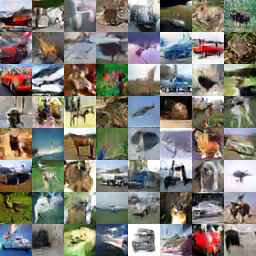}\vspace{-2mm}
\end{subfigure}\hfill
\begin{subfigure}[t]{0.23\textwidth}
    \includegraphics[width=\textwidth]{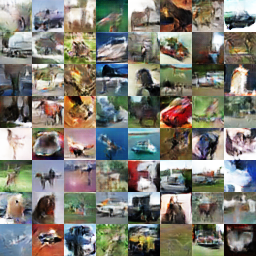}\vspace{-2mm}
\end{subfigure}
\begin{subfigure}[t]{0.23\textwidth}
    \includegraphics[width=\textwidth]{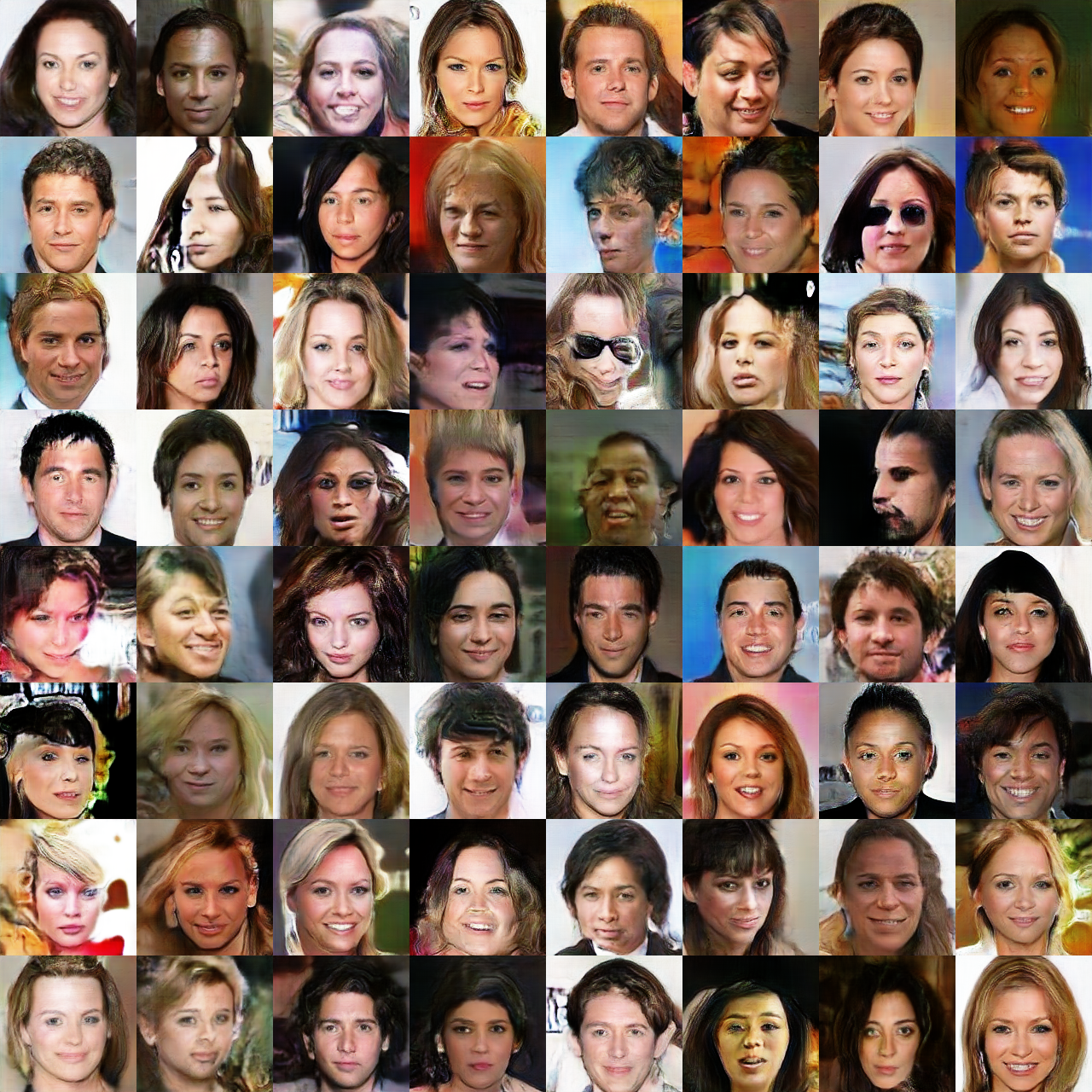}\vspace{-2mm}
\end{subfigure}\hfill
\begin{subfigure}[t]{0.23\textwidth}
    \includegraphics[width=\textwidth]{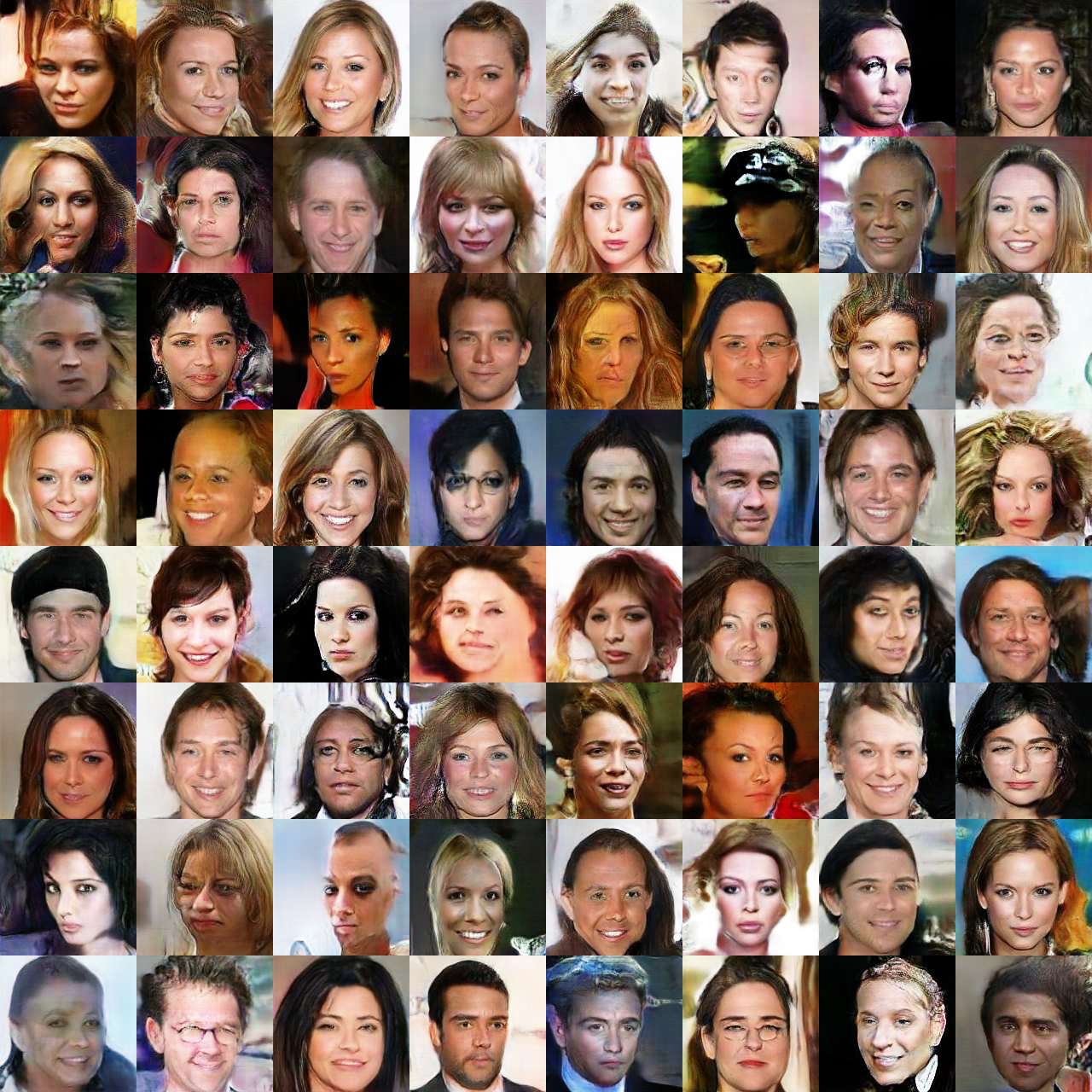}\vspace{-2mm}
\end{subfigure}\hfill
\begin{subfigure}[t]{0.23\textwidth}
    \includegraphics[width=\textwidth]{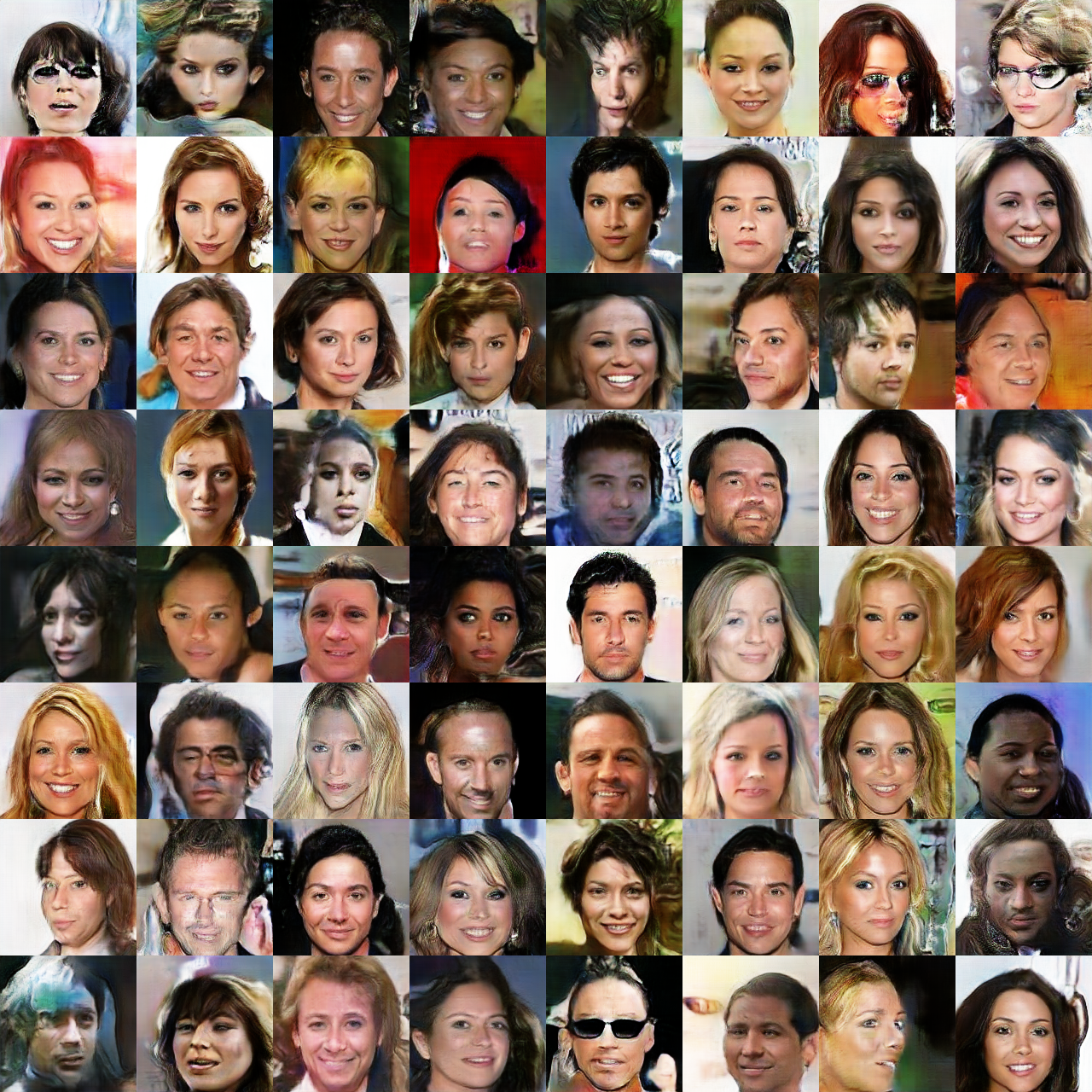}\vspace{-2mm}
\end{subfigure}\hfill
\begin{subfigure}[t]{0.23\textwidth}
    \includegraphics[width=\textwidth]{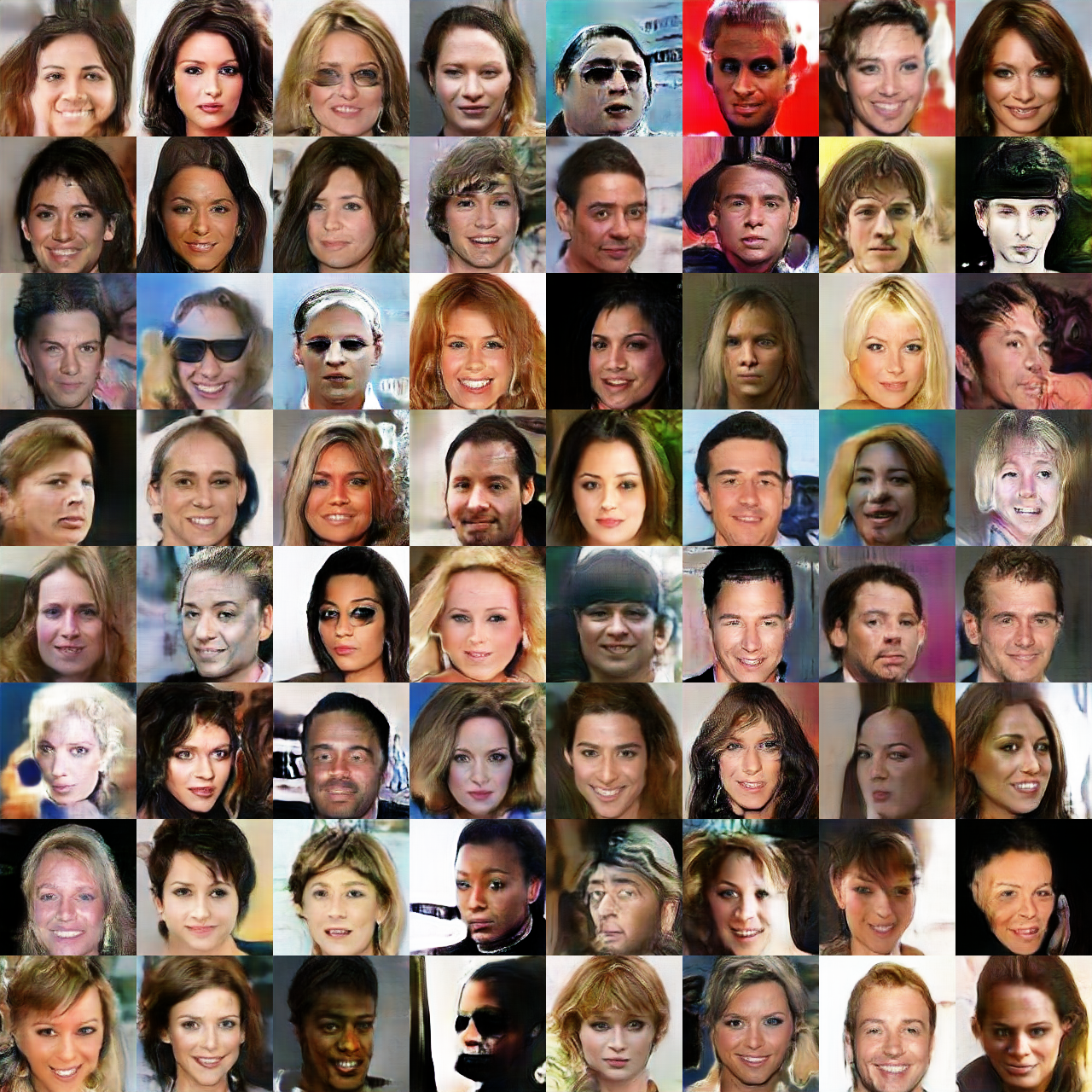}\vspace{-2mm}
\end{subfigure}
\begin{subfigure}[t]{0.23\textwidth}
    \includegraphics[width=\textwidth]{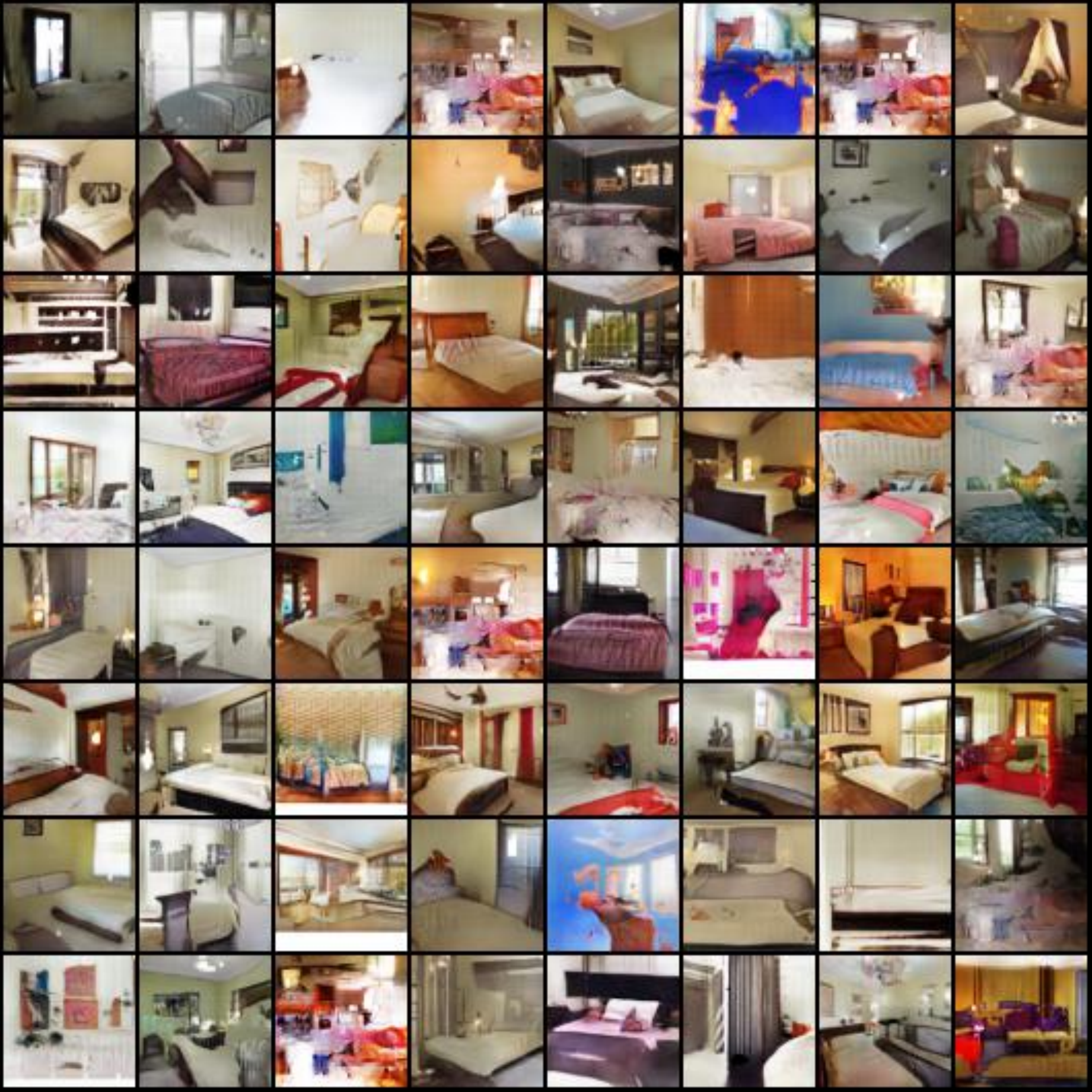}
\caption{\small $\mathcal{L}_D$ + CT.
}
\end{subfigure}\hfill
\begin{subfigure}[t]{0.23\textwidth}
    \includegraphics[width=\textwidth]{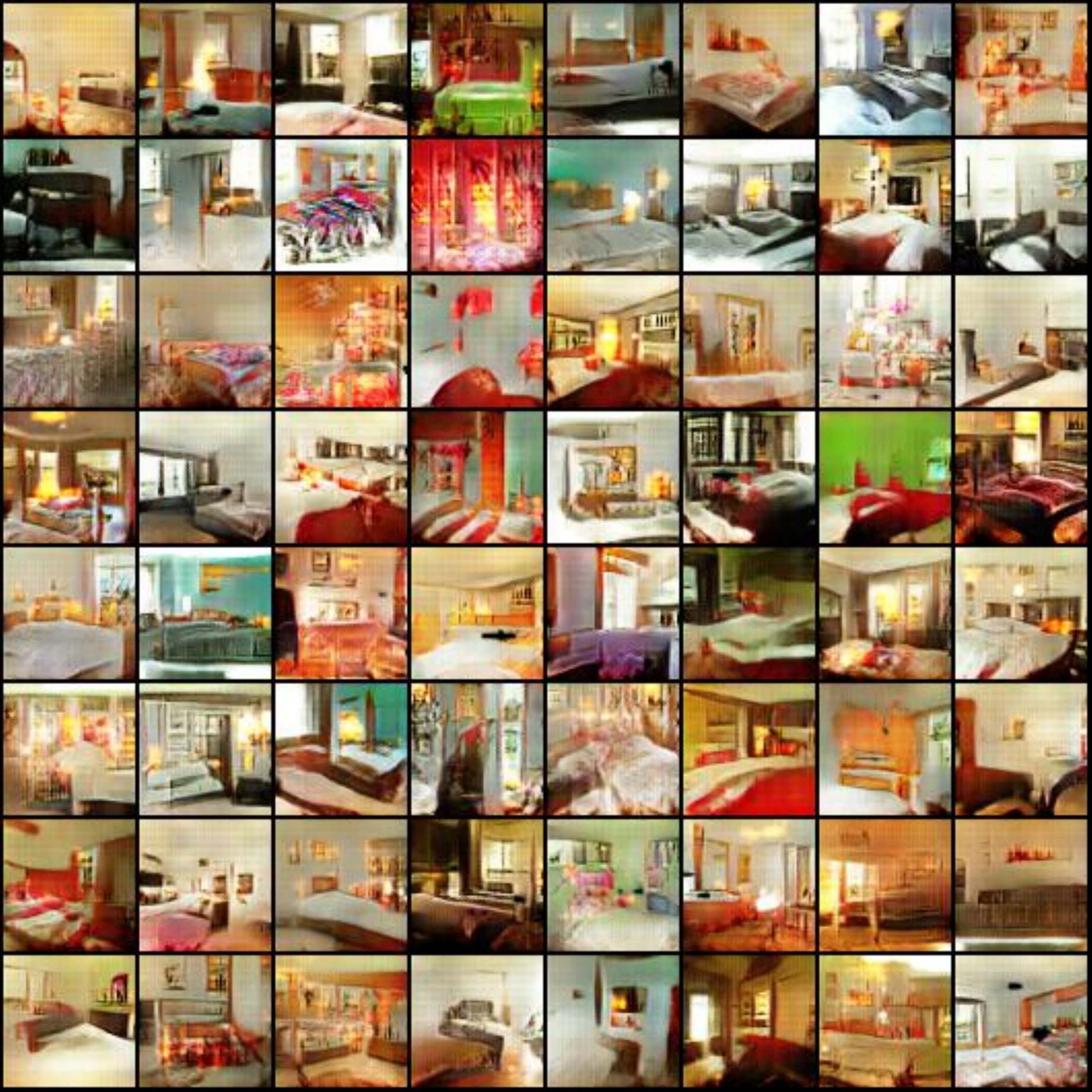}
\caption{\small Slicing + CT.
}
\end{subfigure}\hfill
\begin{subfigure}[t]{0.23\textwidth}
    \includegraphics[width=\textwidth]{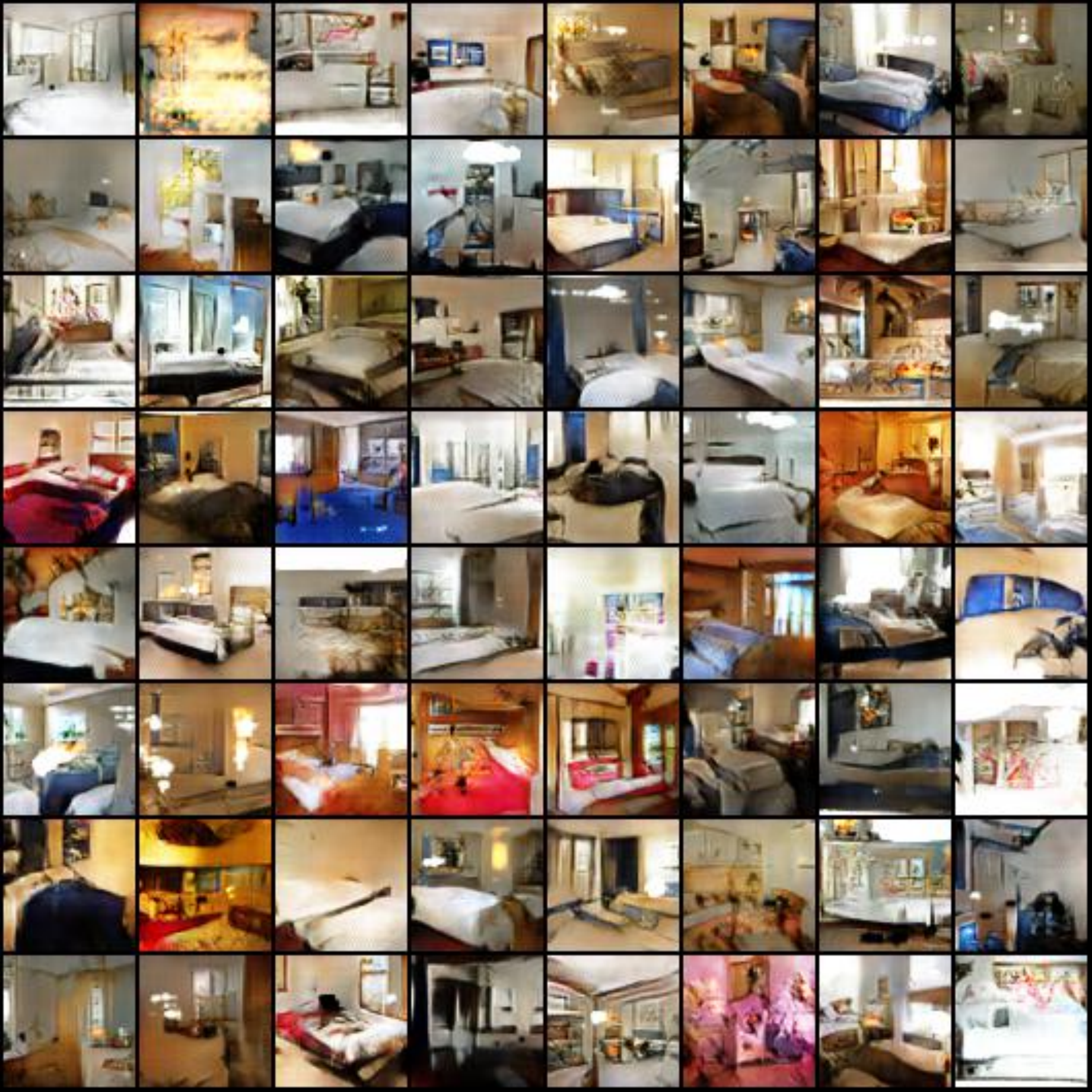}
\caption{\small MMD + CT.
}
\end{subfigure}\hfill
\begin{subfigure}[t]{0.23\textwidth}
    \includegraphics[width=\textwidth]{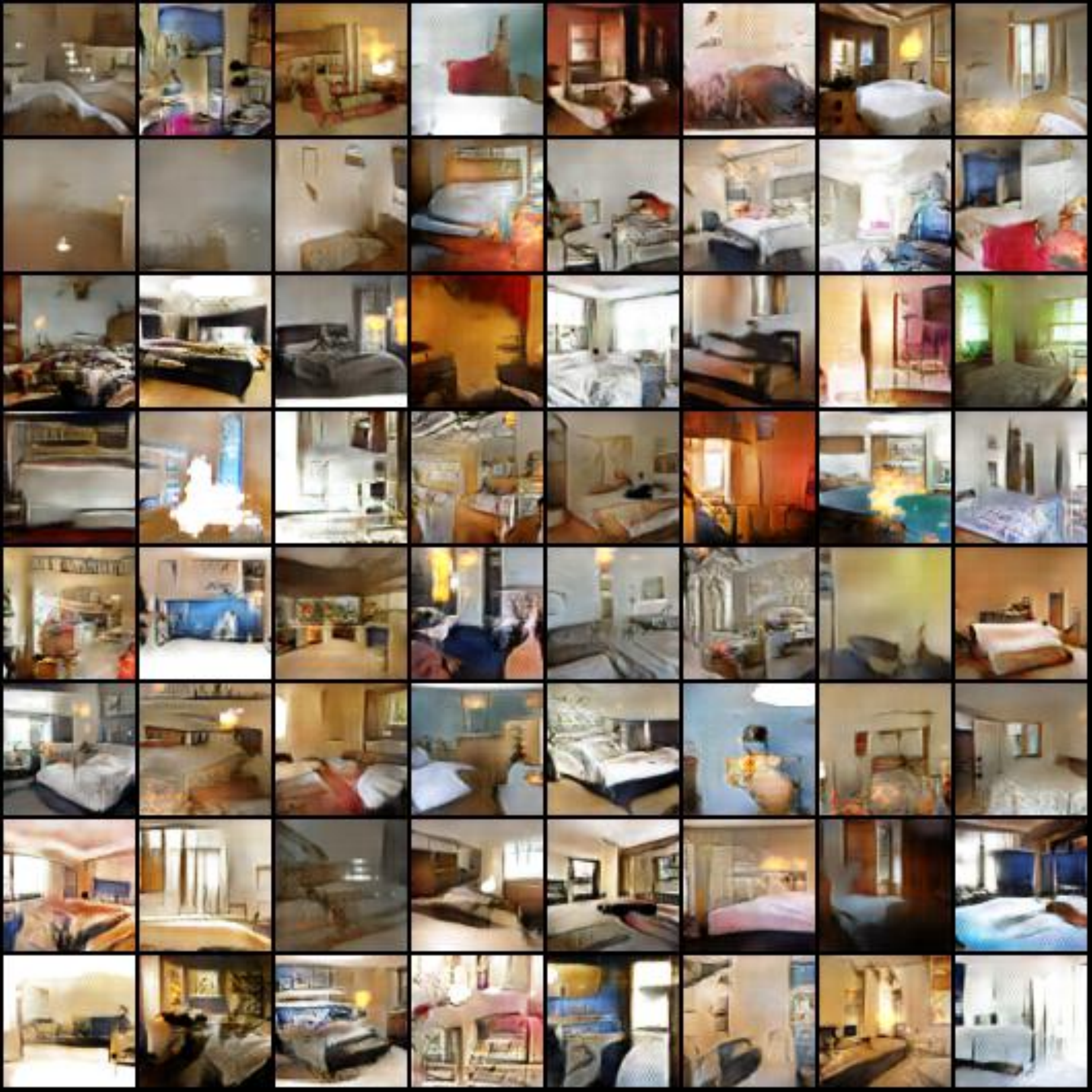}
\caption{\small Adv CT.
}
\end{subfigure}
\caption{\small Analogous plot to Fig.~\ref{fig:ablation_coop} and Fig.~\ref{fig:ablation_coop_appendix} on image datasets. Ablation of fitting results by minimizing CT in different spaces. }\label{fig:image_coop_appendix}\vspace{-2mm}
\end{figure}

Here we provide additional results to the cooperative experiments, where we minimize CT in the feature encoder spaces trained by: 1) maximizing discriminator loss in GANs, 2) using random slicing projections, 3) maximizing MMD and 4) maximizing CT cost. Fig.~\ref{fig:ablation_coop_appendix} shows the results analogous to Fig.~\ref{fig:ablation_coop} on other three synthetic datasets: Swiss-Roll, Half-Moon and 25-Gaussian mixture. Fig.~\ref{fig:image_coop_appendix} provide qualitative results of Table~\ref{tab:comparison_co-train} and Table~\ref{tab:comparison_MMD}.

\subsection{Empirical Wasserstein loss vs empirical CT}\label{appendix:WvsCT}
From Table~\ref{tab:comparison_co-train}, Fig.~\ref{fig:ablation_coop}, and Fig.~\ref{fig:ablation_coop_appendix} we notice the proposed CT can improve the fitting with SWG~\cite{kolouri2018sliced} in the sliced 1D space. Considering SWG applies random slicing projections to project high-dimensional data to several 1D spaces, since the empirical Wasserstein distance has a close form in 1D case and can be calculated with ordered statistics, here we compare the empirical Wasserstein loss and empirical CT cost with a 1D toy experiments. 

\begin{figure}[!t]
\centering

 \centering
 \includegraphics[width=\columnwidth]{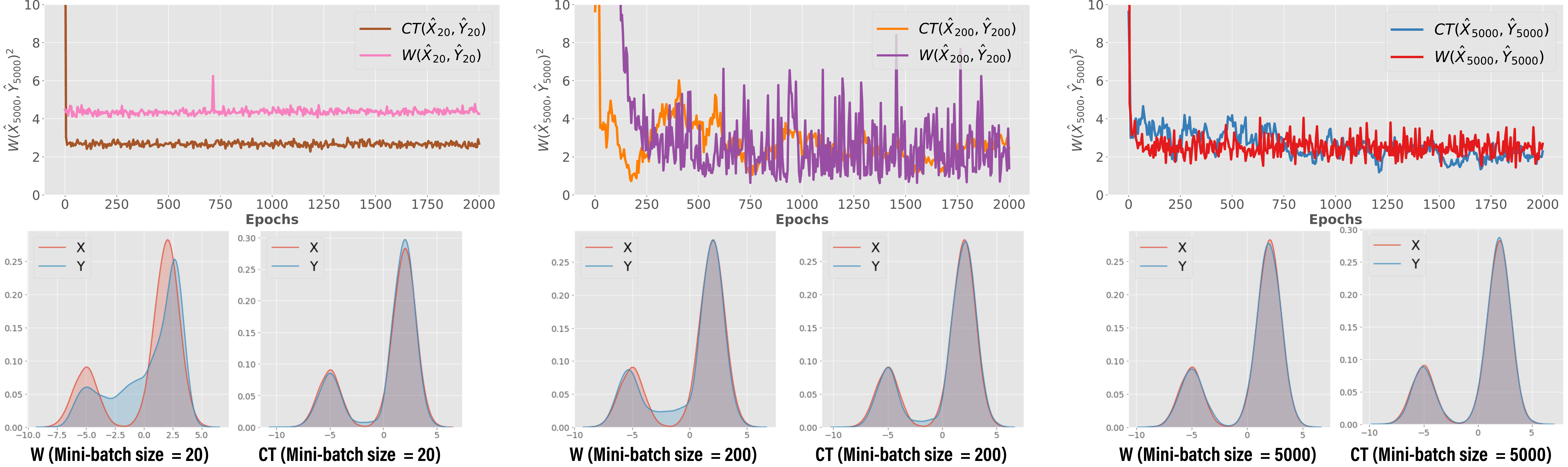}
 \vspace{-4mm}
 \caption{\small \textit{Top}: Plot of the sample Wasserstein distance $W_2(\hat X_{5000}, \hat Y_{5000})^2$ against the number of training epochs, where %
 the generator is trained with either $W_2(\hat X_N, \hat Y_N)^2$ or the CT cost between $\hat X_N$ and $\hat Y_N$, with the mini-batch size set as $N=20$ (left), $N=200$ (middle), or $N=5000$ (right); one epoch consists of $5000/N$ SGD iterations. 
 \textit{Bottom}: The fitting results of different configurations, where the KDE curves of the data distribution and the generative one are marked in red and blue, respectively.}\vspace{-4mm}
 \label{fig:1d_act_wasserstein}
\end{figure}

Let's consider the same 1D Gaussian mixture data used in Fig.~\ref{fig:ACT_fb}, where the bimodal Gaussian mixture has a density form $p_X(x) = \frac{1}{4}\mathcal{N}(x;-5,1) + \frac{3}{4}\mathcal{N}(x;2,1)$. We use an empirical sample set $\mathcal X$, consisting of $|\mathcal X|=5,000$ samples, and train a generative model with the Wasserstein loss and CT cost estimated with these empirical data and generated samples. We vary the training mini-batch size from small to large. Fig.~\ref{fig:1d_act_wasserstein} shows the training curve \textit{w.r.t.} each training epoch and the fitting results with mini-batch size 20, 200 and 5000. We can observe when the mini-batch size $N$ is as large as 5000, both Wasserstein and CT lead to a well-trained generator. However, as shown in the left and middle columns, when $N$ is getting much smaller, the generator trained with Wasserstein under-performs that trained with ACT, especially when the mini-batch size becomes as small as $N=20$. While the Wasserstein distance $\mathcal{W}(X,Y)$ in theory can well guide the training of a generative model, the sample Wasserstein distance $\mathcal{W}(\hat X_N,\hat Y_N)$, whose optimal transport plan is locally re-computed for each mini-batch, could be sensitive to the mini-batch size $N$, which also explains why in practice the SWG are difficult to fit desired distribution. By contrast, CT shows better robustness across mini-batches, leading to a well-trained generator whose performance has low sensitivity to the mini-batch size.

\clearpage

\subsection{Additional results on mode-covering/mode-seeking study}\label{sec:B3}
The mode covering and mode seeking behaviors discussed in Figs.~\ref{fig:ACT_fb} also exist in the real image case. For illustration, we use the Stacked-MNIST dataset \cite{srivastava2017veegan} and fit CT in three configurations: normal, forward only, and backward only. DCGAN \cite{radford2015unsupervised}, VEEGAN \cite{srivastava2017veegan}, PacGAN \cite{lin2018pacgan}, and PresGAN \cite{dieng2019prescribed} are applied here as the baseline models to evaluate the mode-capturing capability. 

\begin{table}[h]
\centering
\caption{Assessing mode collapse on Stacked-MNIST. The true total number of modes is 1,000. DCGAN, VEEGAN, and CT (Backward only) all suffer from collapse. 
The other models capture
nearly all the modes of the data distribution. Furthermore, the distribution of the labels predicted from the images produced by 
these models
is closer to the data distribution, which shows lower KL scores.}
\label{tab:mode-capture}
\resizebox{.65\textwidth}{!}{%
\begin{tabular}{lll}
\toprule
Method  & Mode Captured & KL            \\
\cmidrule(lr){1-1}  \cmidrule(lr){2-2}   \cmidrule(lr){3-3}
DCGAN \cite{radford2015unsupervised}   & 392.0  $\pm$ 7.376 & 8.012  $\pm$ 0.056 \\
VEEGAN \cite{srivastava2017veegan} & 761.8  $\pm$ 5.741 & 2.173  $\pm$ 0.045 \\
PacGAN \cite{lin2018pacgan} & 992.0  $\pm$ 1.673 & 0.277  $\pm$ 0.005 \\
PresGAN \cite{dieng2019prescribed} & 999.4  $\pm$ 0.80  & 0.102  $\pm$ 0.003 \\ \cmidrule(lr){1-1}  \cmidrule(lr){2-2}   \cmidrule(lr){3-3}
CT     & 999.07 $\pm$ 0.162      & 0.181  $\pm$ 0.003 \\
CT (Foward only)   & 999.18 $\pm$ 0.9       & 0.124  $\pm$ 0.003 \\
CT (Backward only)   & 192  $\pm$ 1.912        & 9.166 $\pm$ 0.06 \\
\bottomrule
\end{tabular}%
}
\end{table}

\begin{figure}[!ht]
\centering
\includegraphics[width=\textwidth]{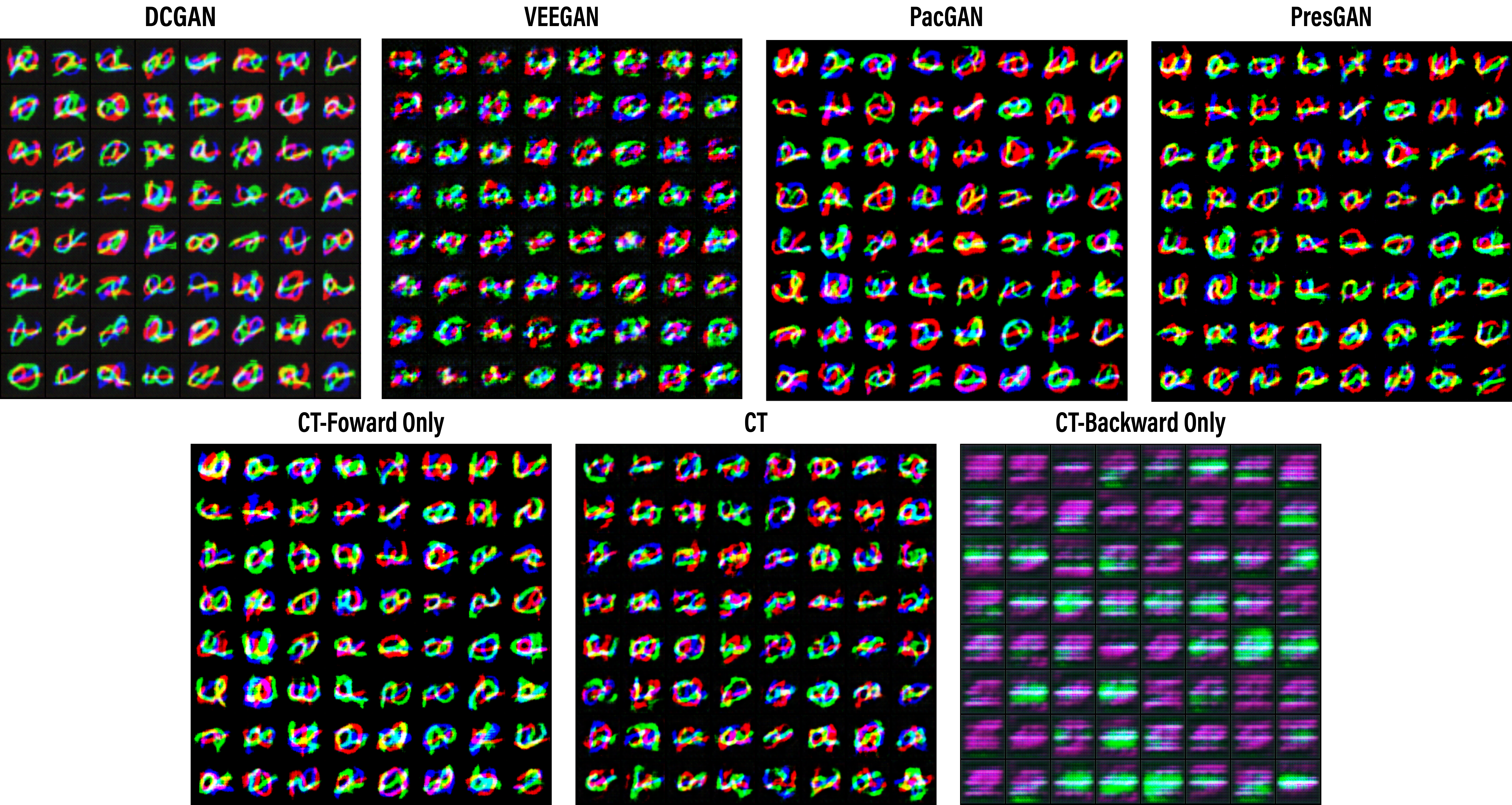}
\caption{Visual results of the generated samples produced by DCGAN, VEEGAN, PacGAN, PresGAN, and ACT-DCGAN on the Stacked-MNIST dataset.}\label{fig:stackedmnist}
\end{figure}

We calculate the captured mode number of each model, as well as the Kullback--Leibler (KL) divergence of the predicted label distributions between the generated samples and true data samples. For Stacked-MNIST data, there are 1000 modes in total. The results in Table~\ref{tab:mode-capture} justify CT using only forward or using both forward and backward can almost capture all the modes, thus we do not suffer from the mode collapse problem. Using backward only can only encourages the mode seeking/dropping behavior. Fig.~\ref{fig:stackedmnist} provides the visual justification of this experiment, where the observations is consistent with those on toy datasets: if we only apply forward CT, the generator is encouraged to cover all the modes; if we only apply the backward CT for optimization, we can observe the mode seeking behavior of the generator. 

\clearpage

\clearpage

\subsection{Additional results on image datasets}\label{appendix:results}
%

\begin{figure}[ht]
\centering
\begin{subfigure}[t]{0.32\textwidth}
    \includegraphics[width=\textwidth]{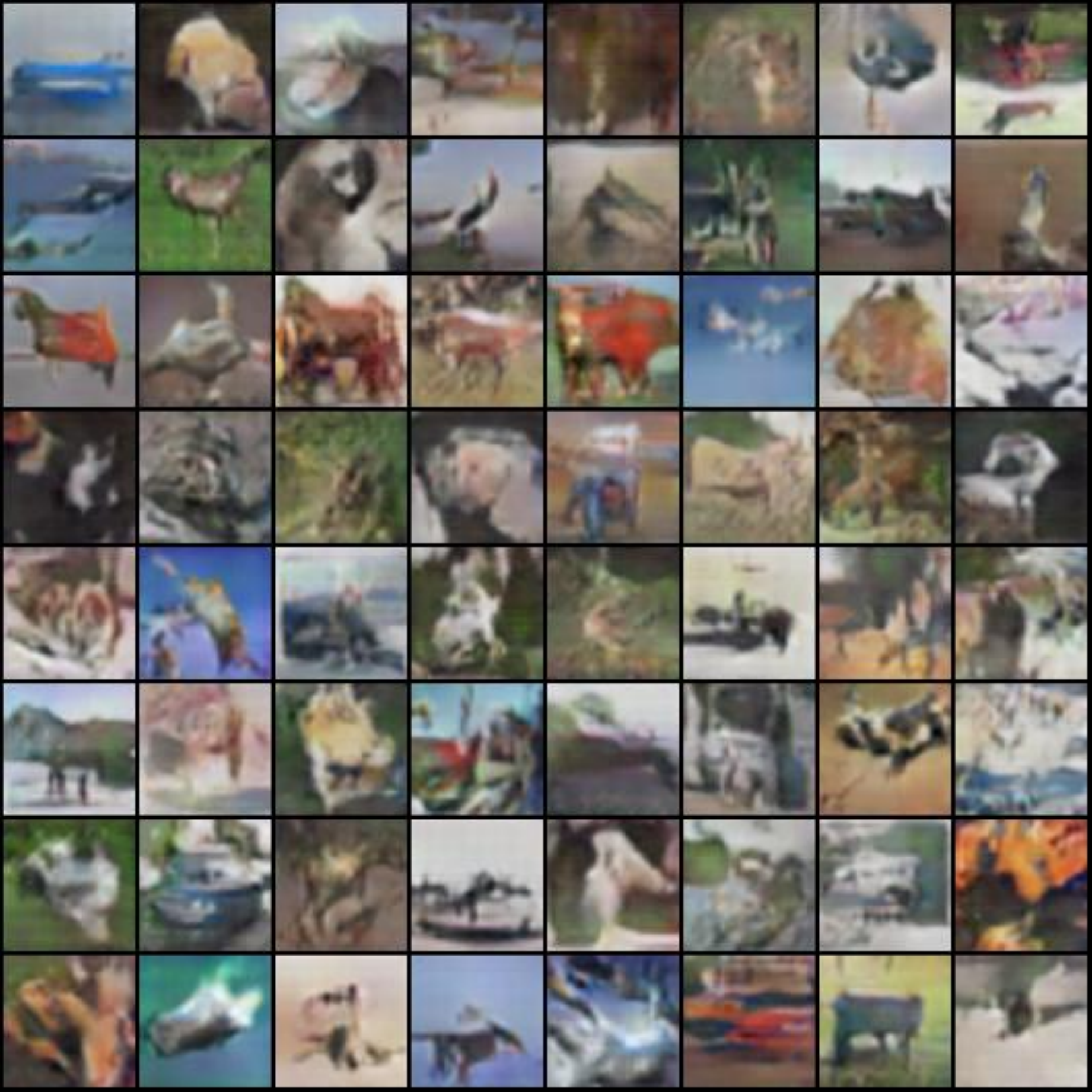}\vspace{-2mm}
\end{subfigure}\hfill
\begin{subfigure}[t]{0.32\textwidth}
    \includegraphics[width=\textwidth]{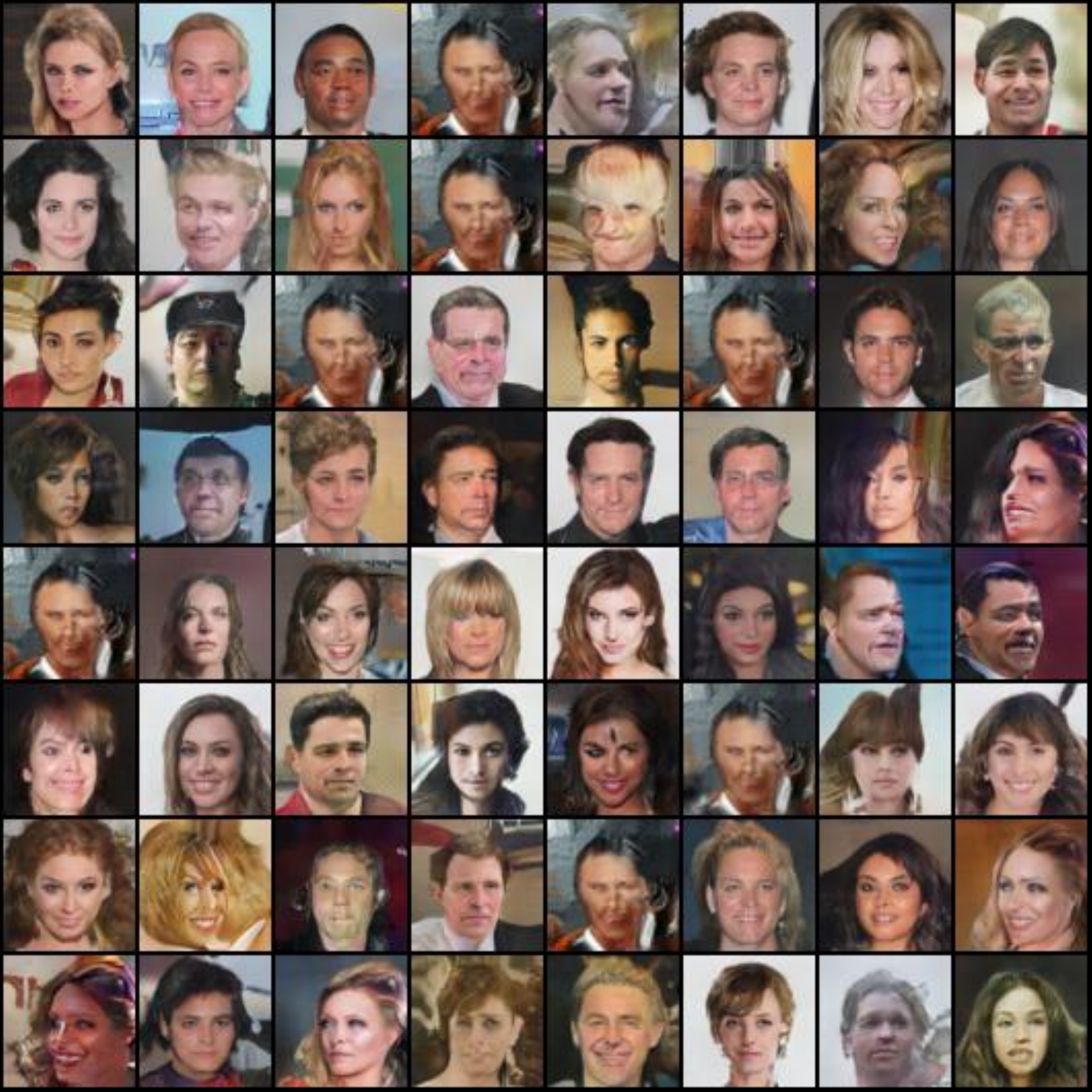}\vspace{-2mm}
\end{subfigure}\hfill
\begin{subfigure}[t]{0.32\textwidth}
    \includegraphics[width=\textwidth]{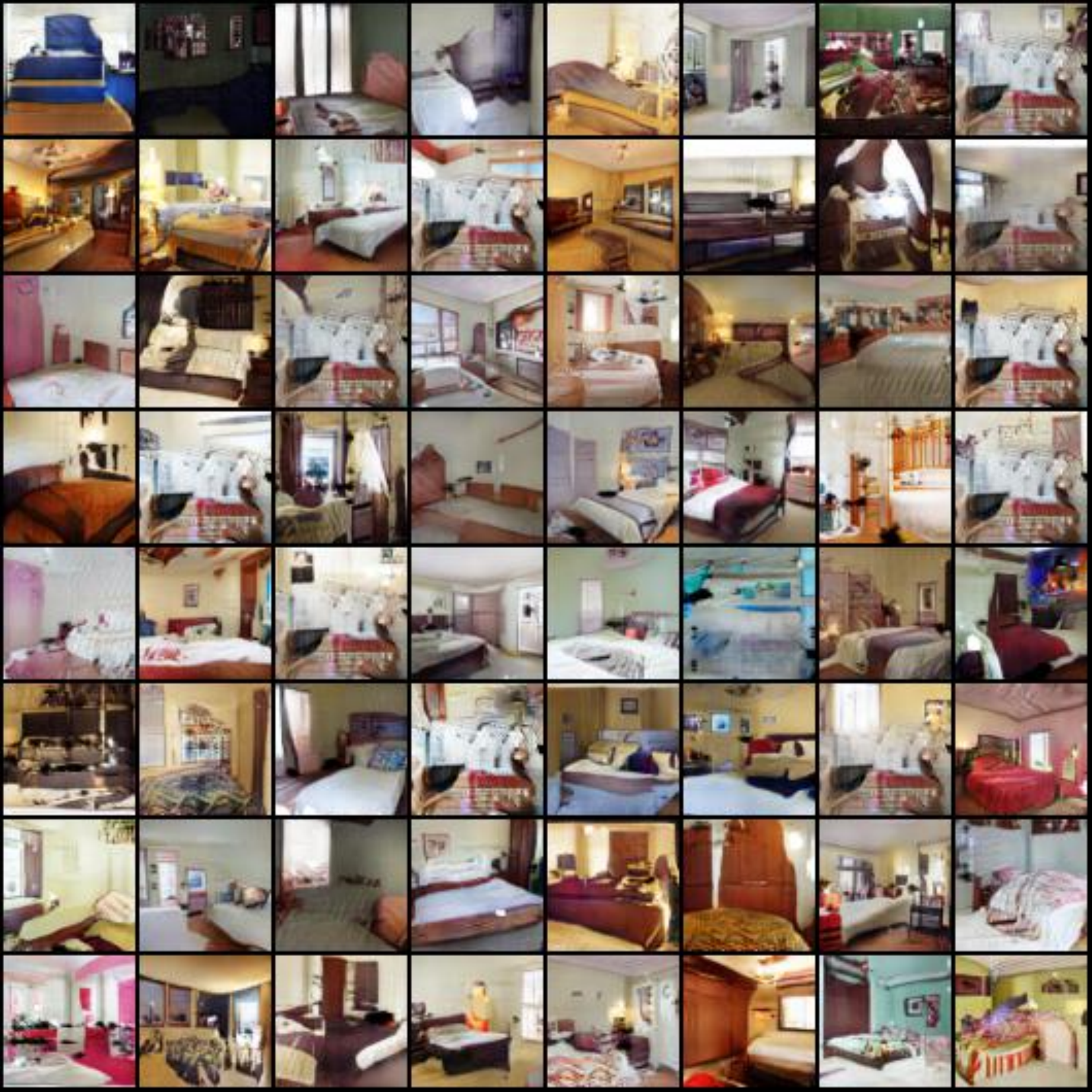}\vspace{-2mm}
\end{subfigure}
\begin{subfigure}[t]{0.32\textwidth}
    \includegraphics[width=\textwidth]{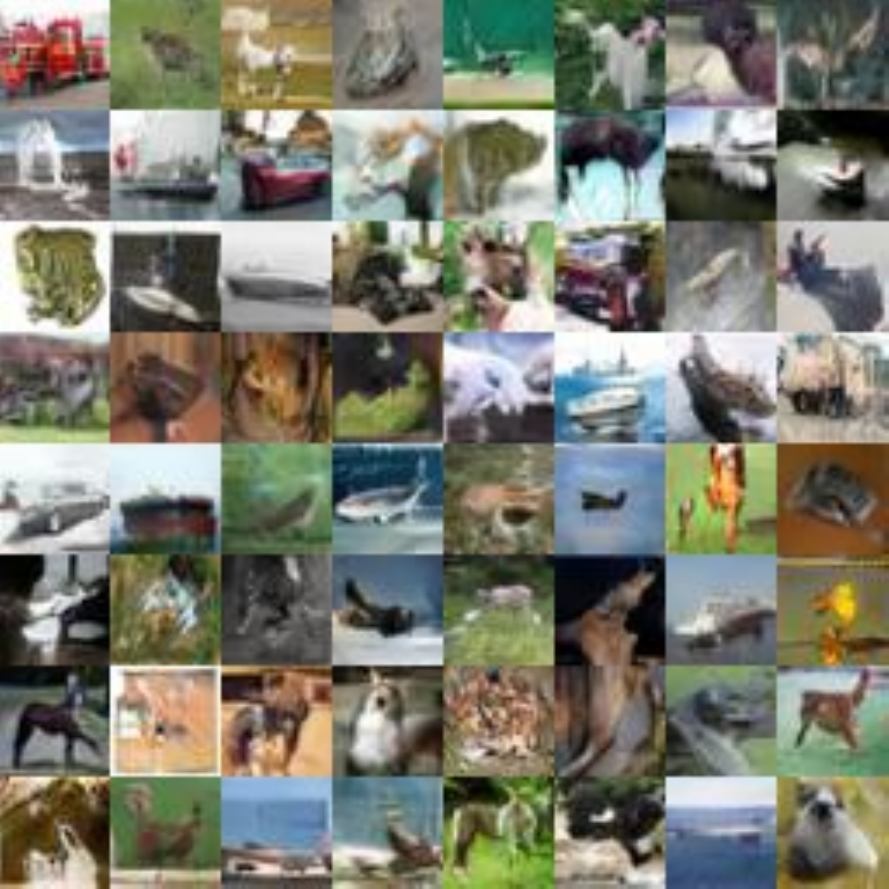}
\caption{\small CIFAR-10.
}
\end{subfigure}\hfill
\begin{subfigure}[t]{0.32\textwidth}
    \includegraphics[width=\textwidth]{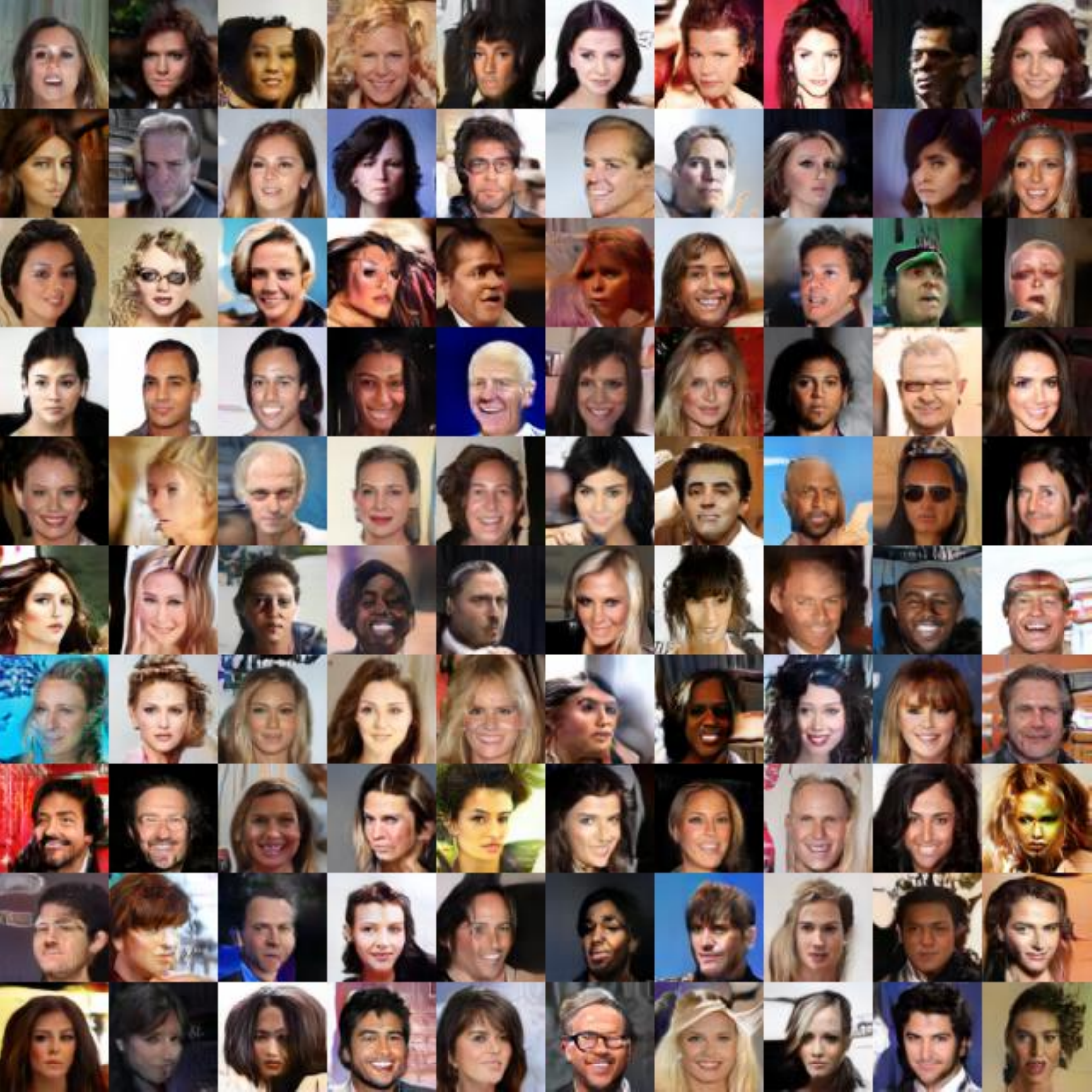}
\caption{\small CelebA.
}
\end{subfigure}\hfill
\begin{subfigure}[t]{0.32\textwidth}
    \includegraphics[width=\textwidth]{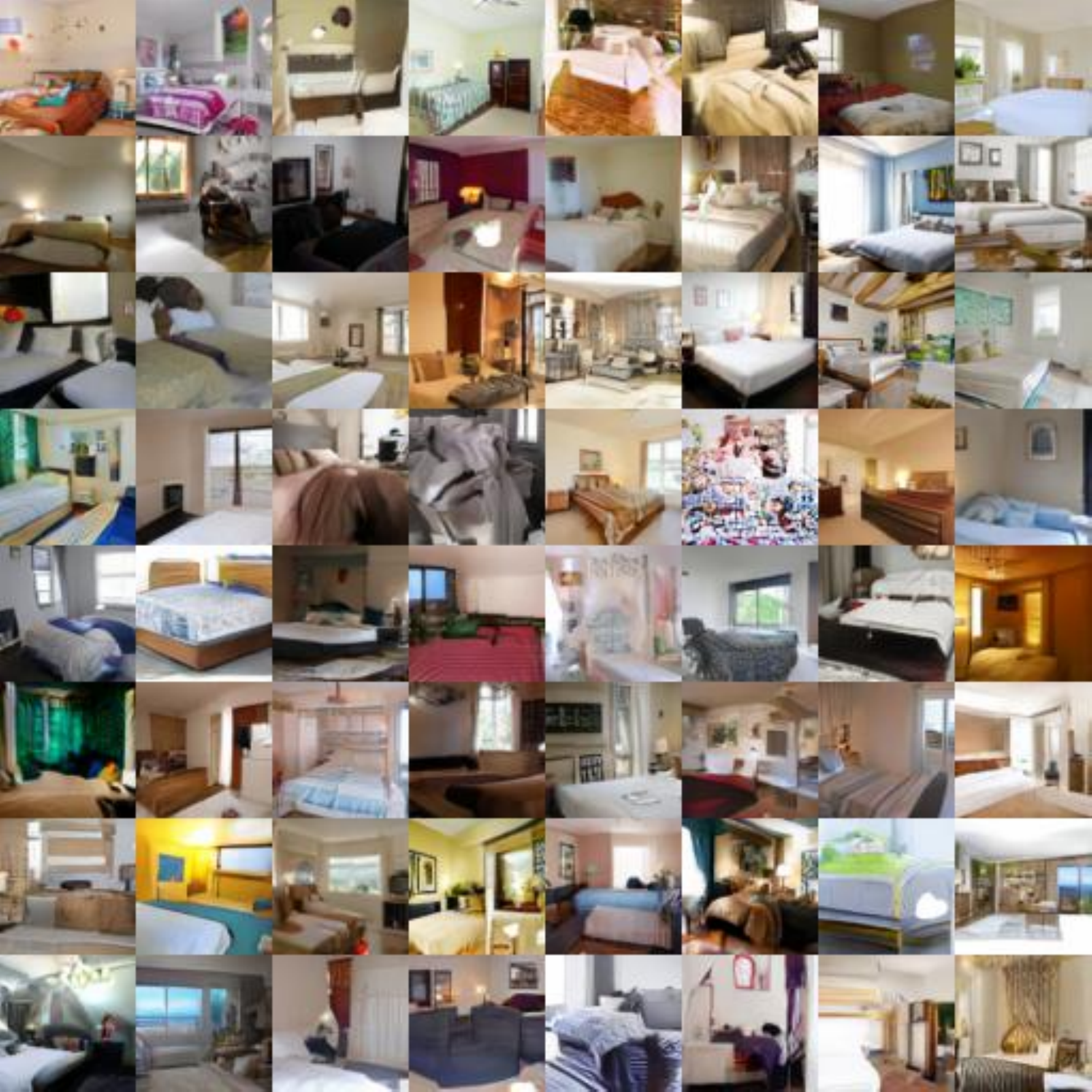}
\caption{\small LSUN-Bedroom.
}
\end{subfigure}
\vspace{-3mm}
\caption{\small Analogous plot to Fig.~\ref{fig:generation}, with additional generated samples. \textit{Top}: samples generated with CNN backbone; \textit{Bottom}: samples generated with ResNet backbone. }\label{fig:image_appendix}\vspace{-4mm}
\end{figure}

\begin{figure}[ht]
\centering
\begin{subfigure}[t]{0.32\textwidth}
    \includegraphics[width=\textwidth]{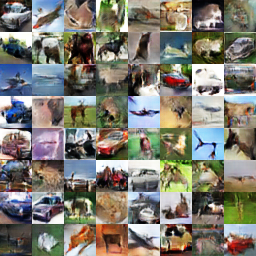}\vspace{-2mm}
\end{subfigure}\hfill
\begin{subfigure}[t]{0.32\textwidth}
    \includegraphics[width=\textwidth]{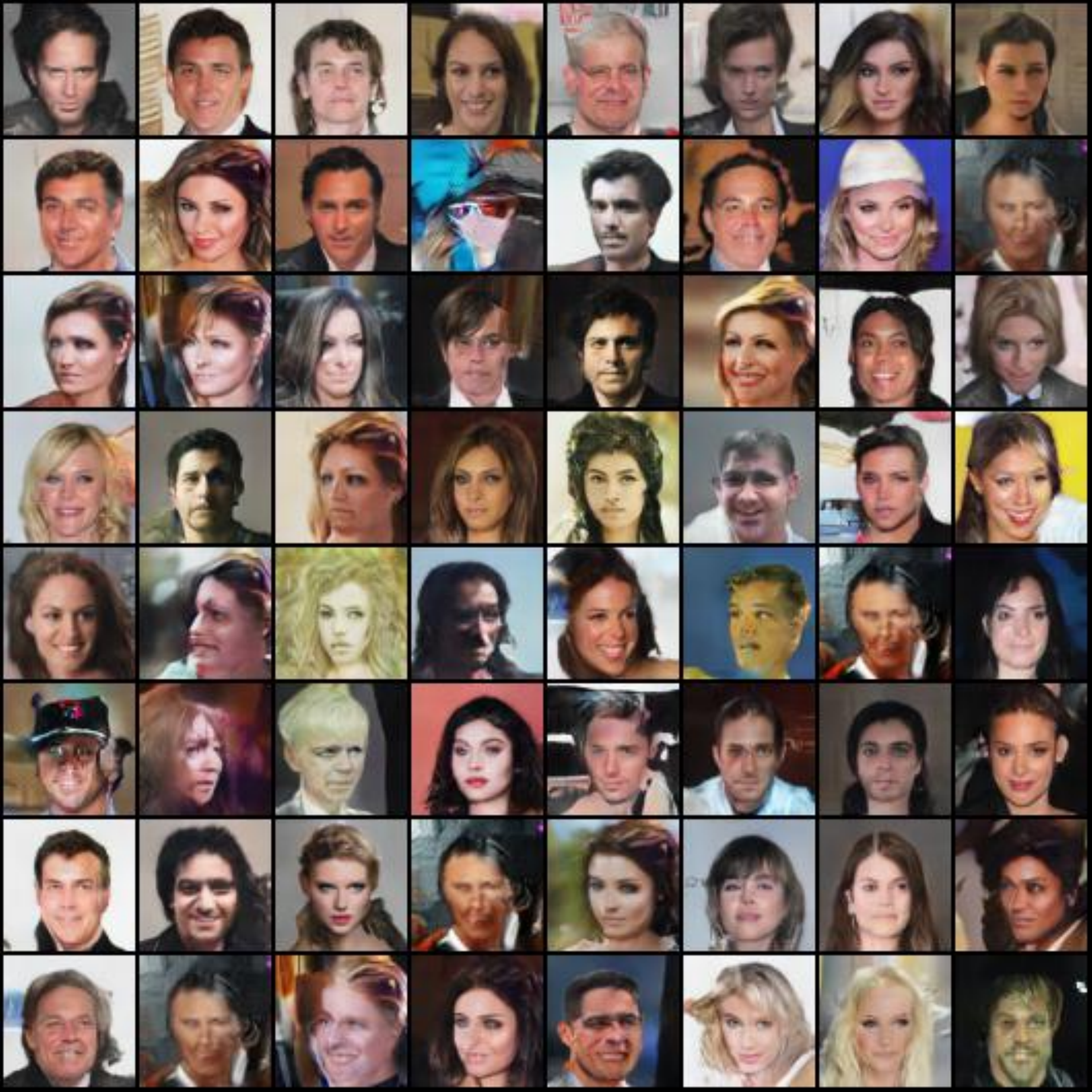}\vspace{-2mm}
\end{subfigure}\hfill
\begin{subfigure}[t]{0.32\textwidth}
    \includegraphics[width=\textwidth]{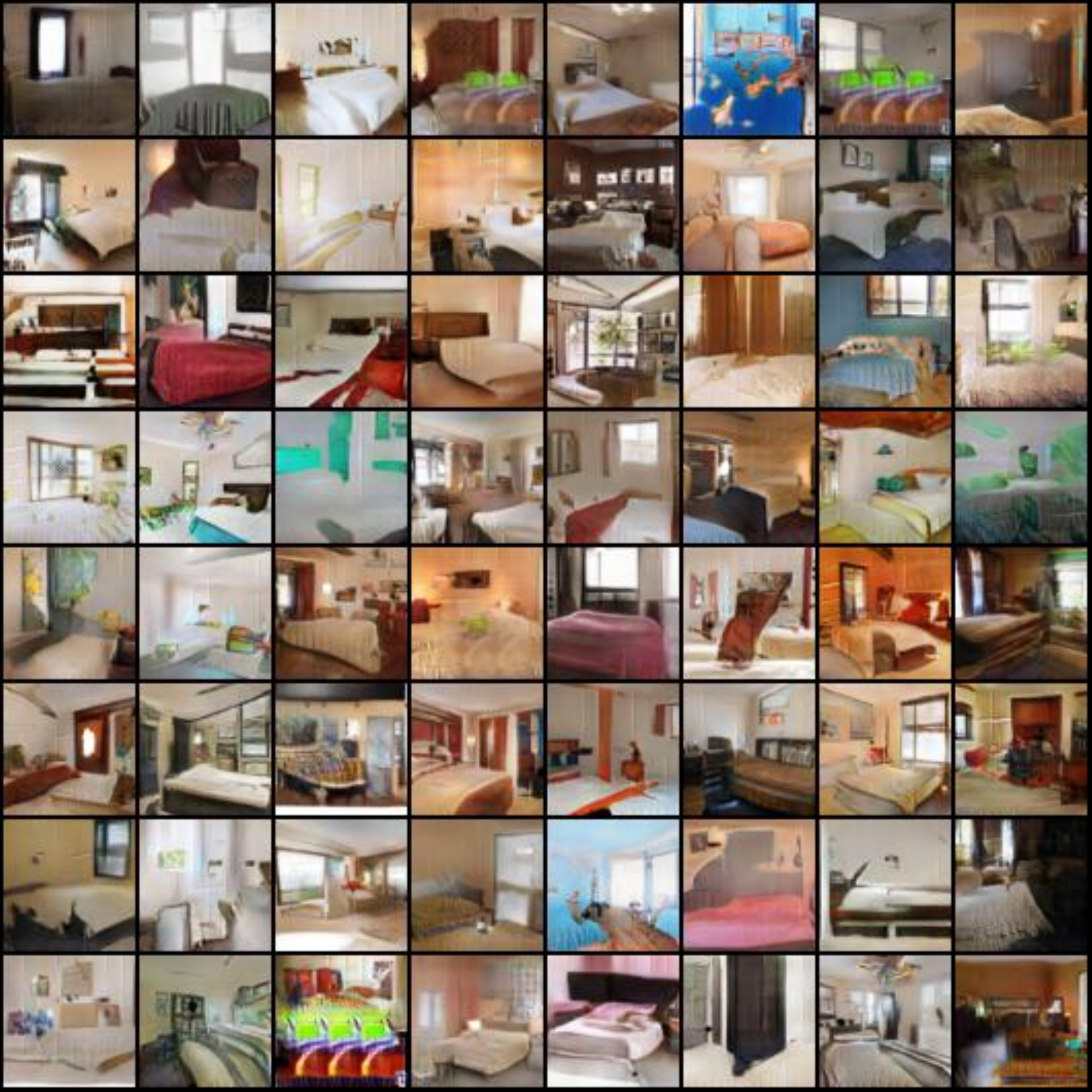}\vspace{-2mm}
\end{subfigure}
\begin{subfigure}[t]{0.32\textwidth}
    \includegraphics[width=\textwidth]{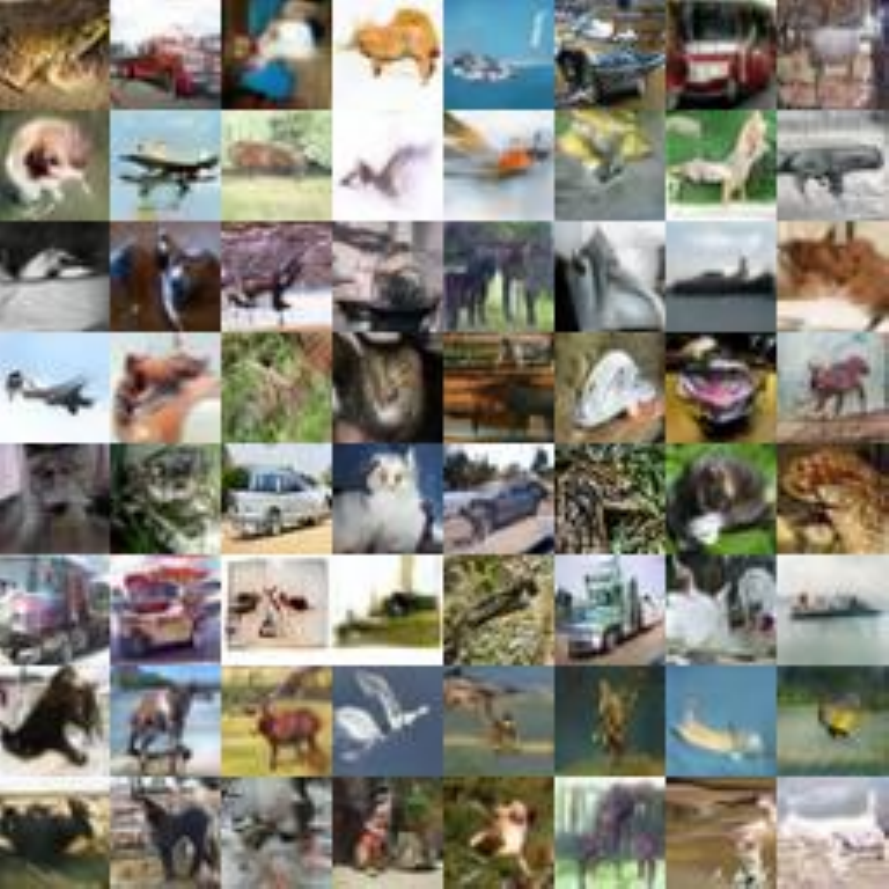}
\caption{\small CIFAR-10.
}
\end{subfigure}\hfill
\begin{subfigure}[t]{0.32\textwidth}
    \includegraphics[width=\textwidth]{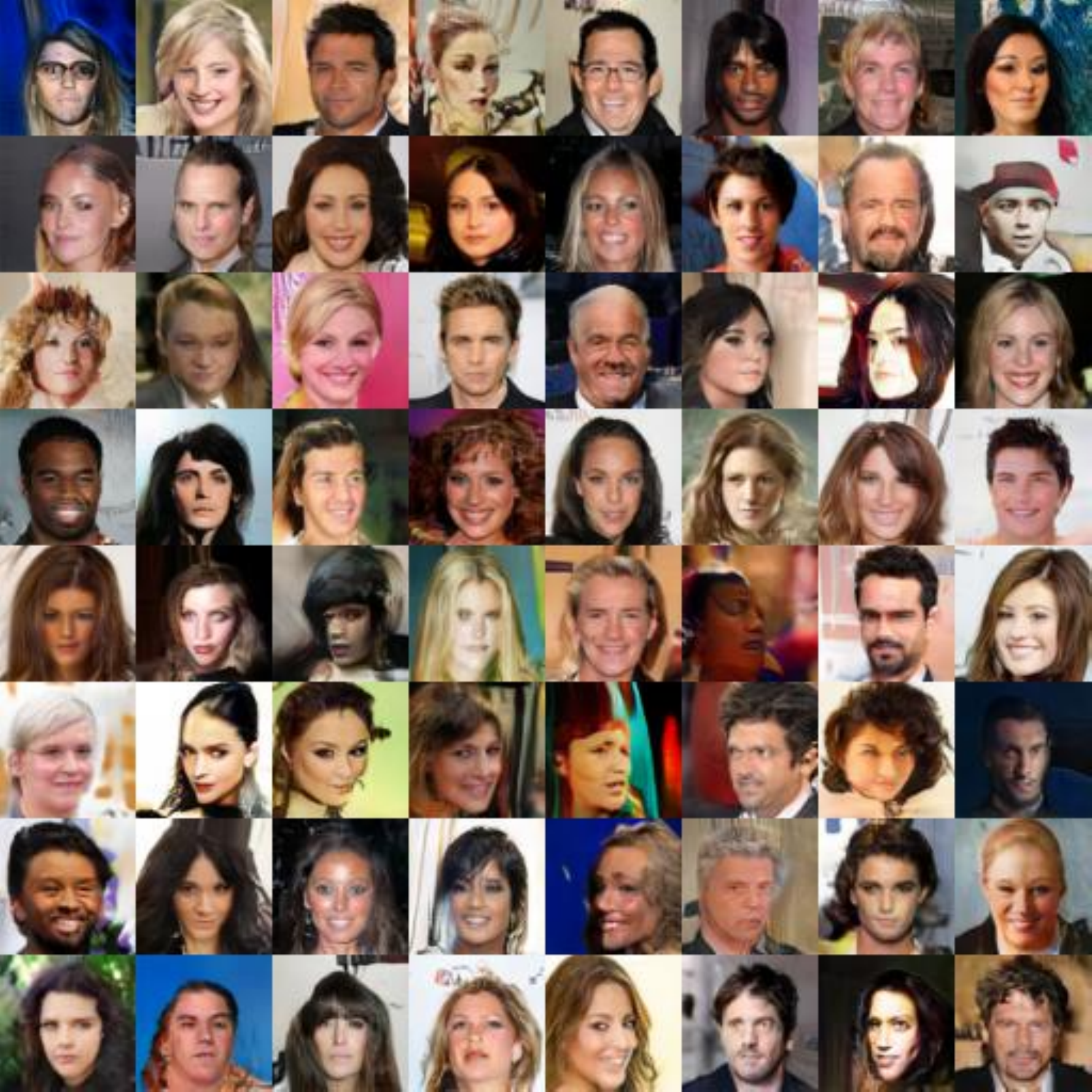}
\caption{\small CelebA.
}
\end{subfigure}\hfill
\begin{subfigure}[t]{0.32\textwidth}
    \includegraphics[width=\textwidth]{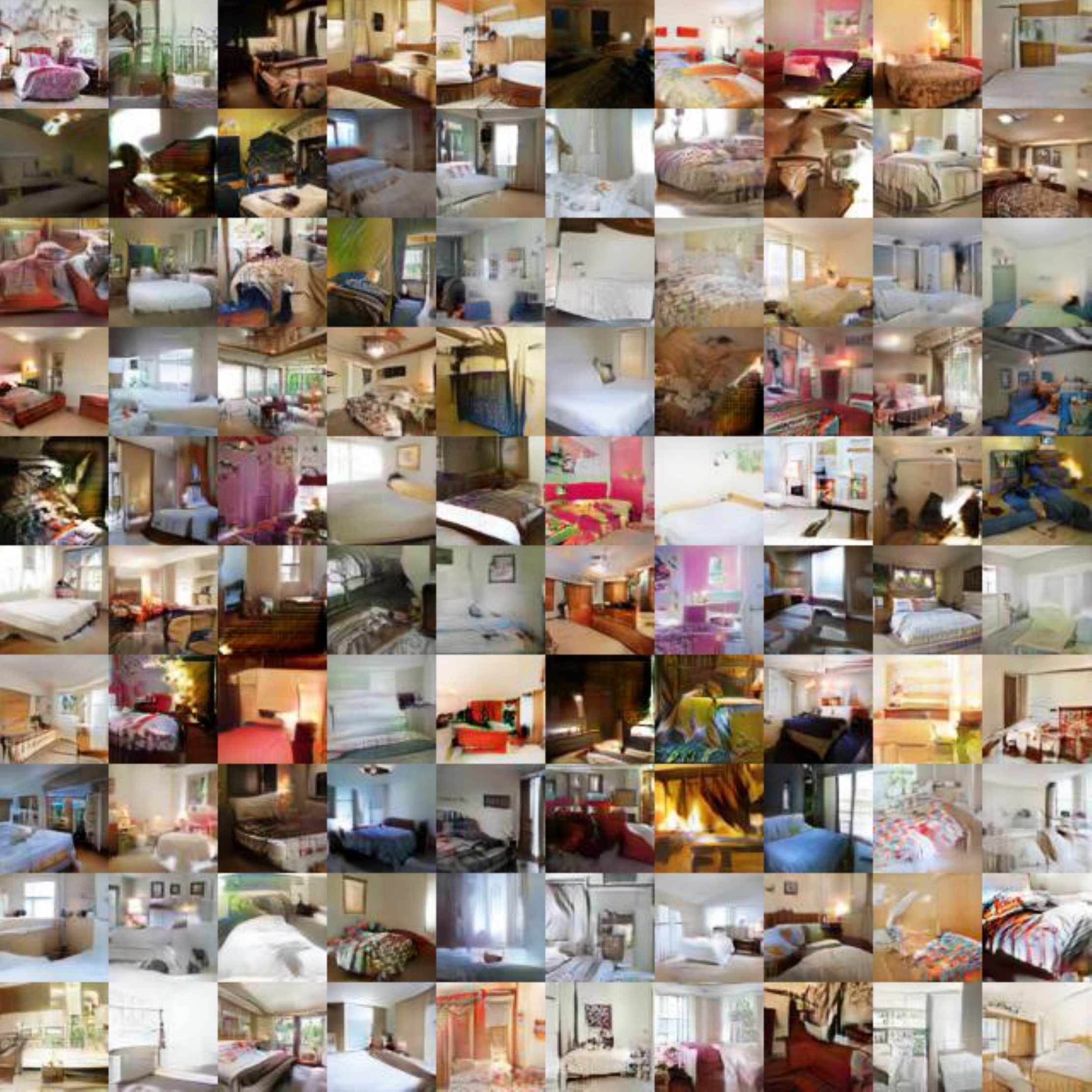}
\caption{\small LSUN-Bedroom.
}
\end{subfigure}
\vspace{-3mm}
\caption{\small Analogous plot to Fig.~\ref{fig:image_appendix}. }\label{fig:image2_appendix}\vspace{-4mm}
\end{figure}

\begin{figure}[ht]
\centering
\begin{subfigure}[t]{\textwidth}
    \includegraphics[width=\textwidth]{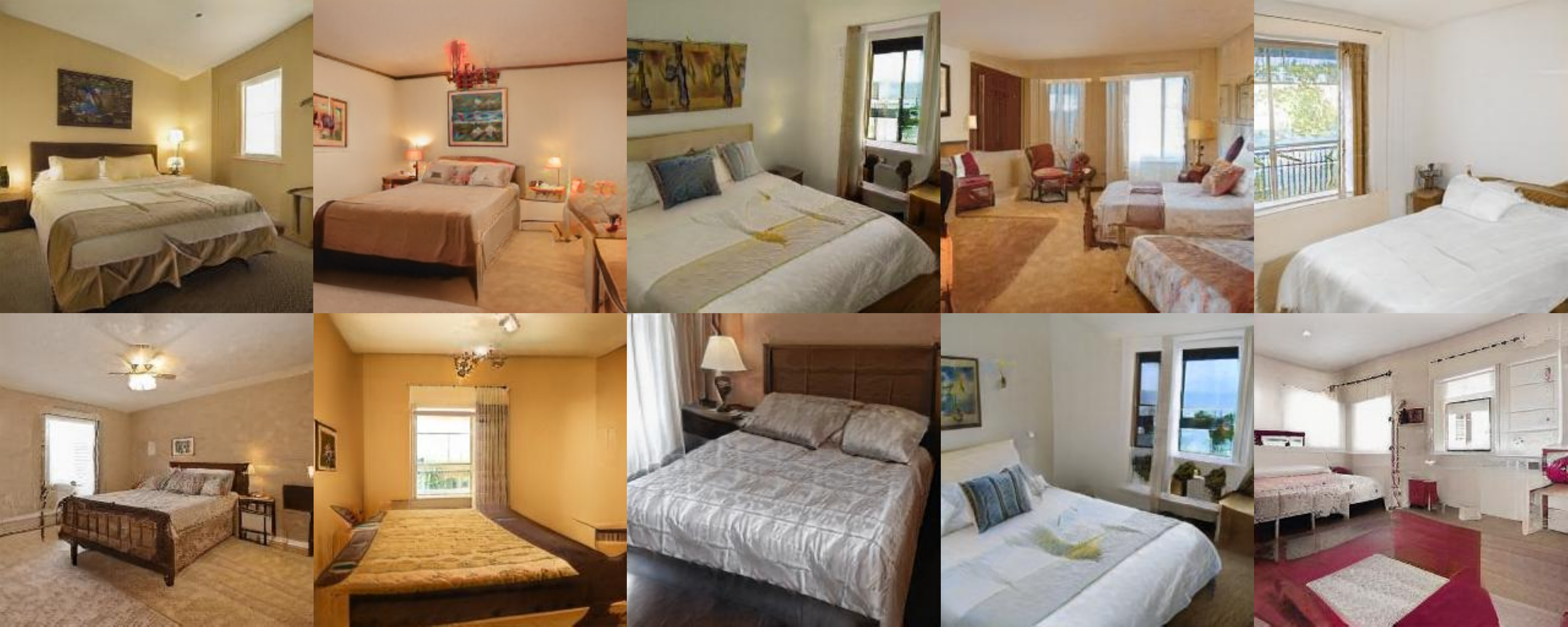}
\caption{\small LSUN-Bedroom (256x256).
}\vspace{2mm}
\end{subfigure}

\begin{subfigure}[t]{\textwidth}
    \includegraphics[width=\textwidth]{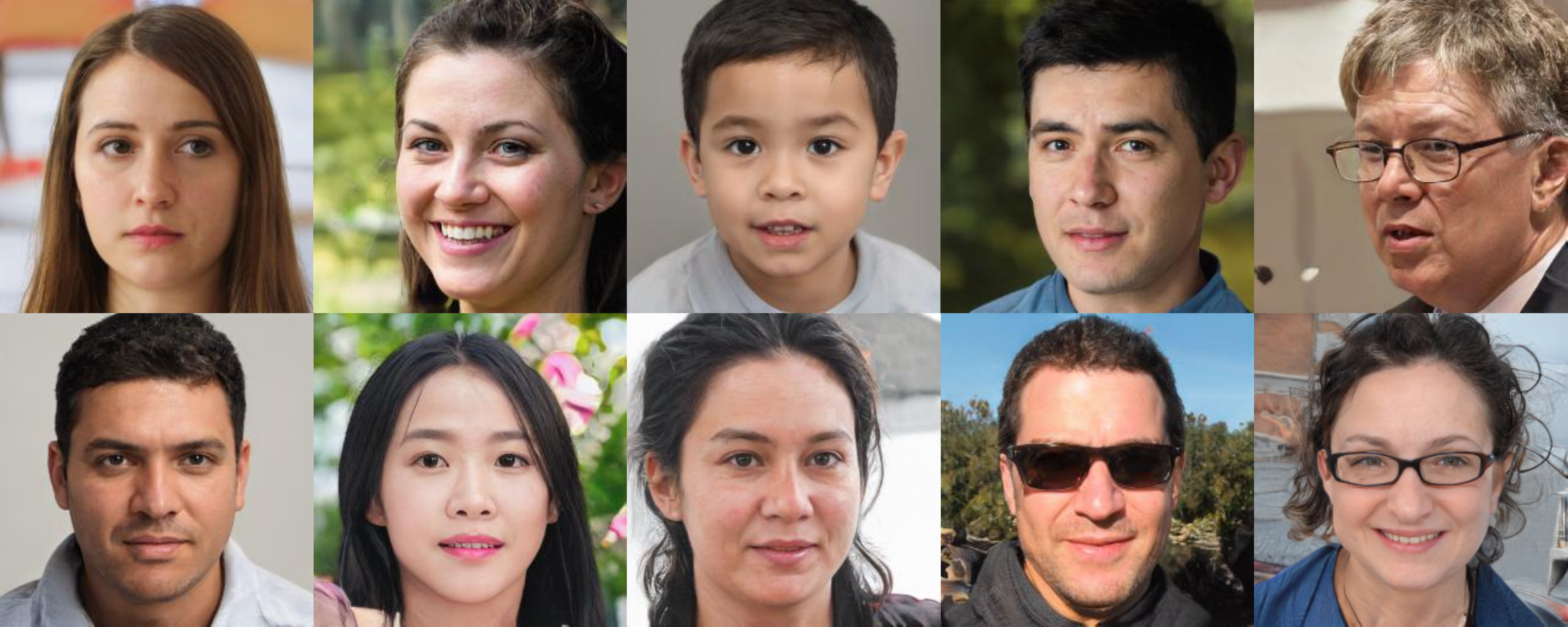}
\caption{\small FFHQ (256x256).
}\vspace{2mm}
\end{subfigure}
\begin{subfigure}[t]{\textwidth}
    \includegraphics[width=0.24\textwidth]{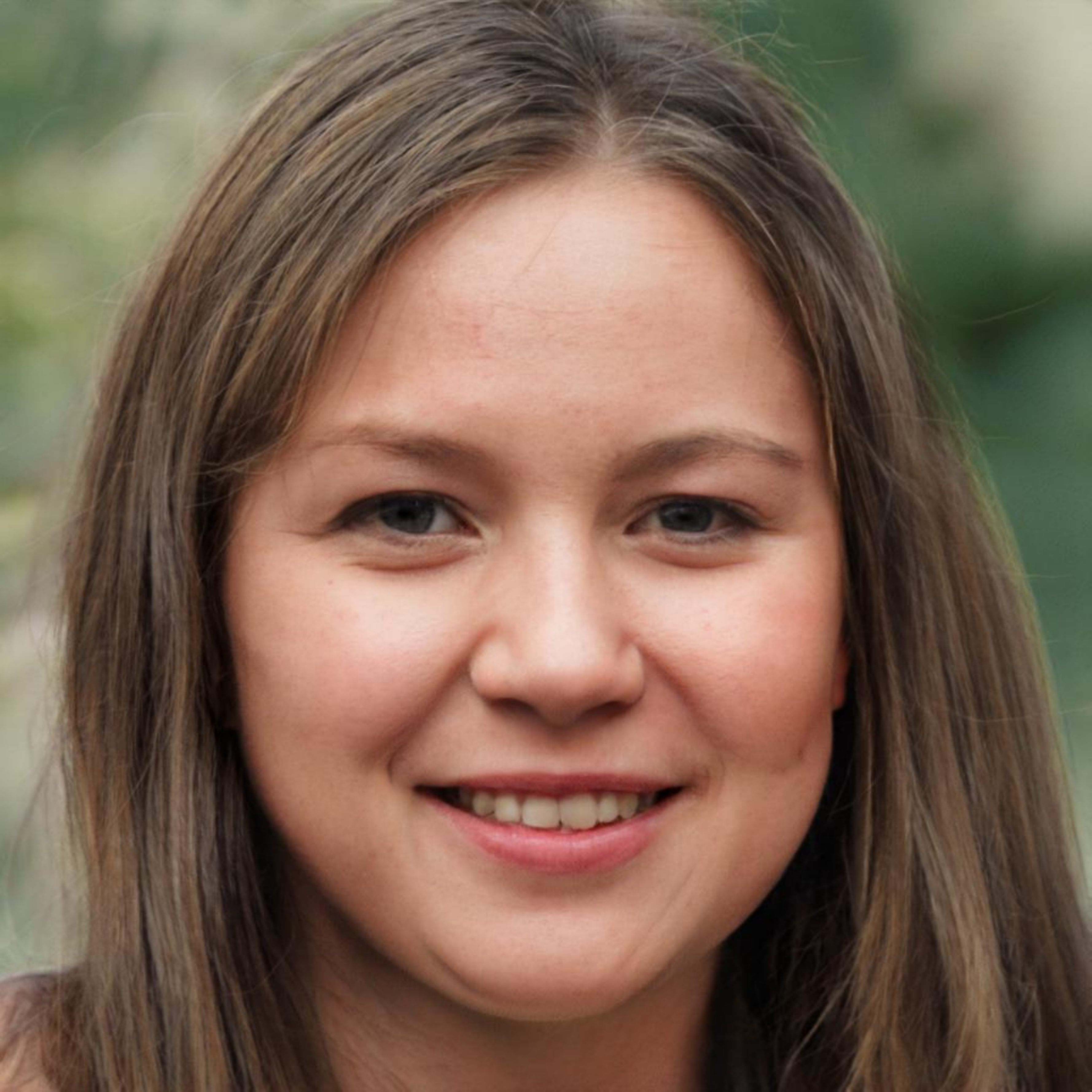}
    \includegraphics[width=0.24\textwidth]{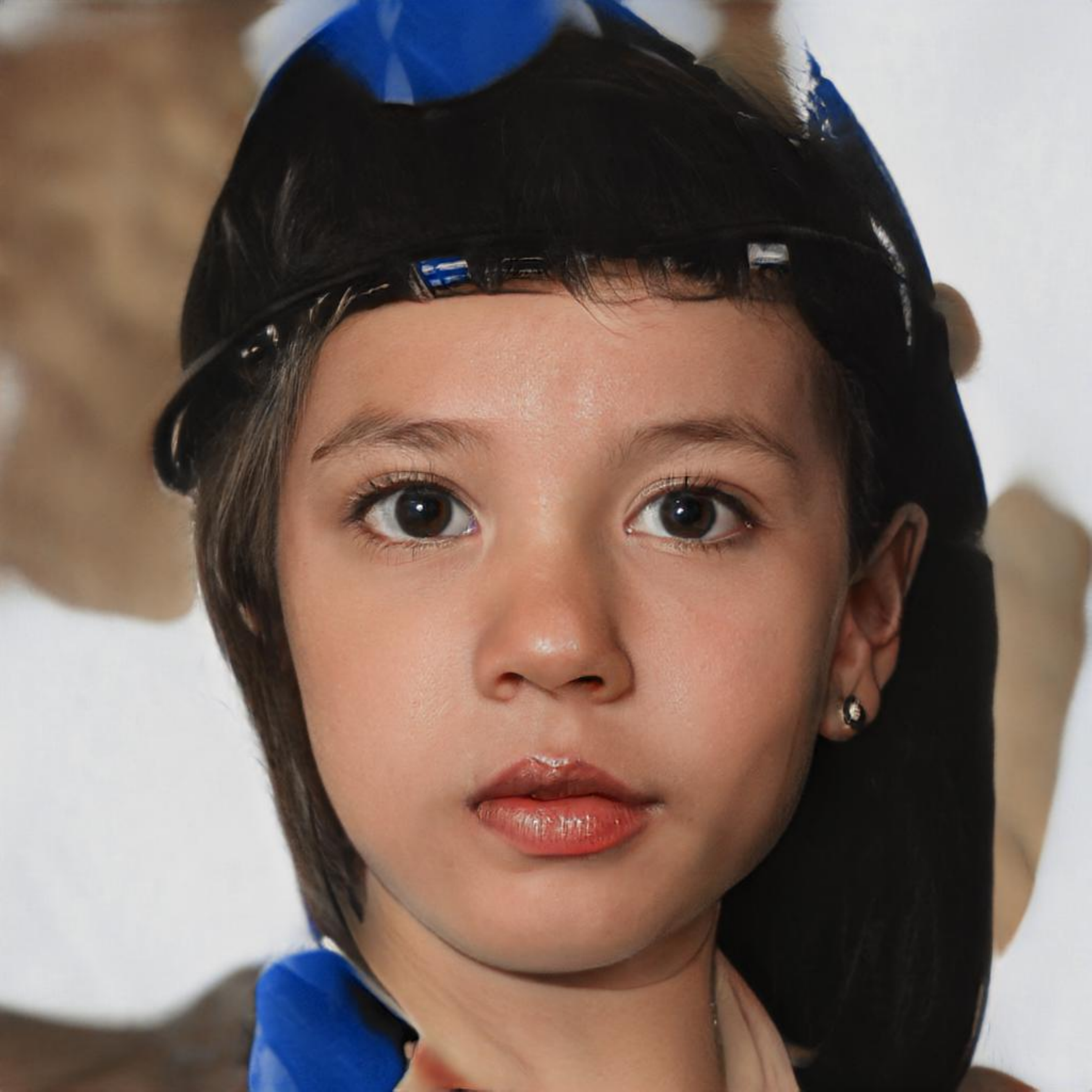}
    \includegraphics[width=0.24\textwidth]{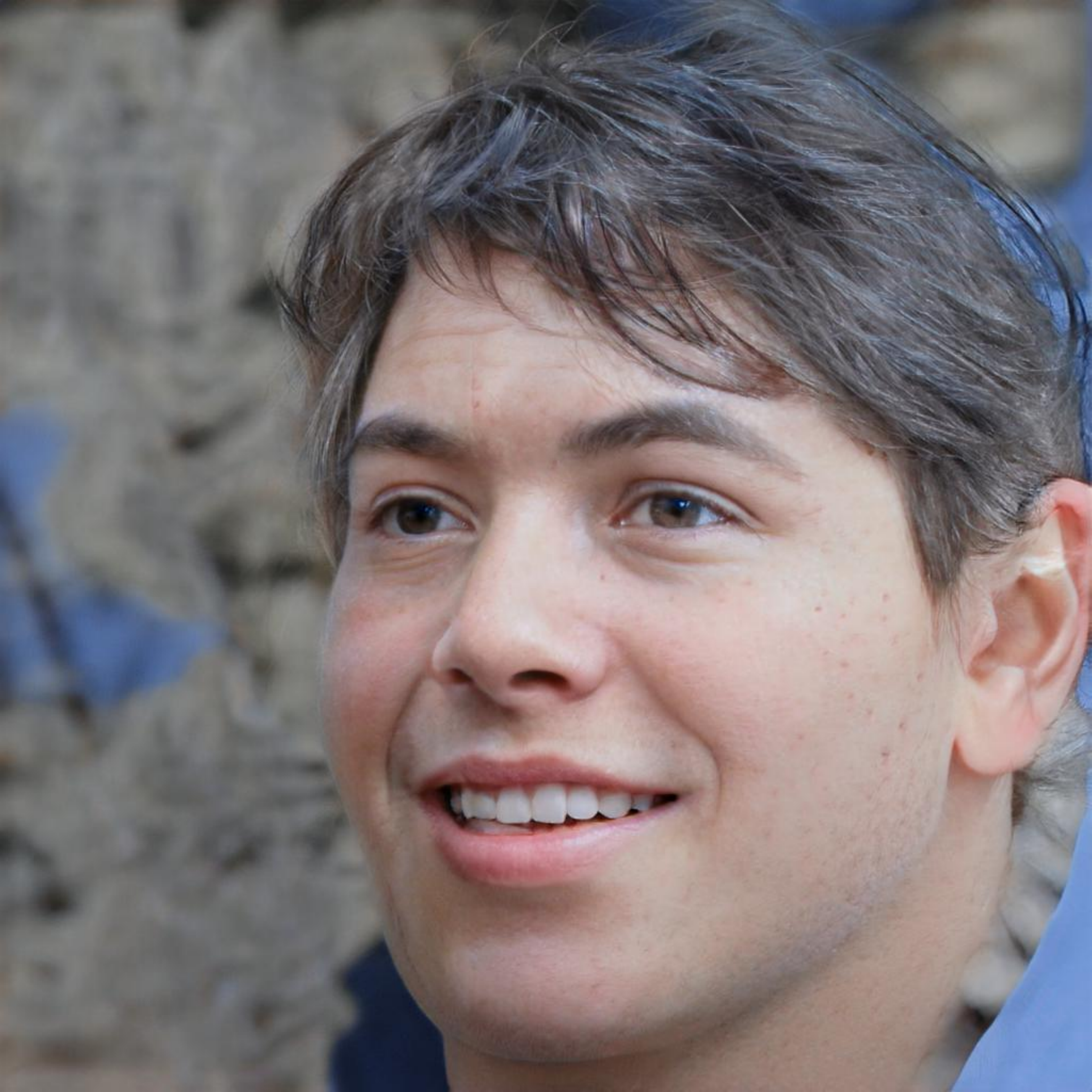}
    \includegraphics[width=0.24\textwidth]{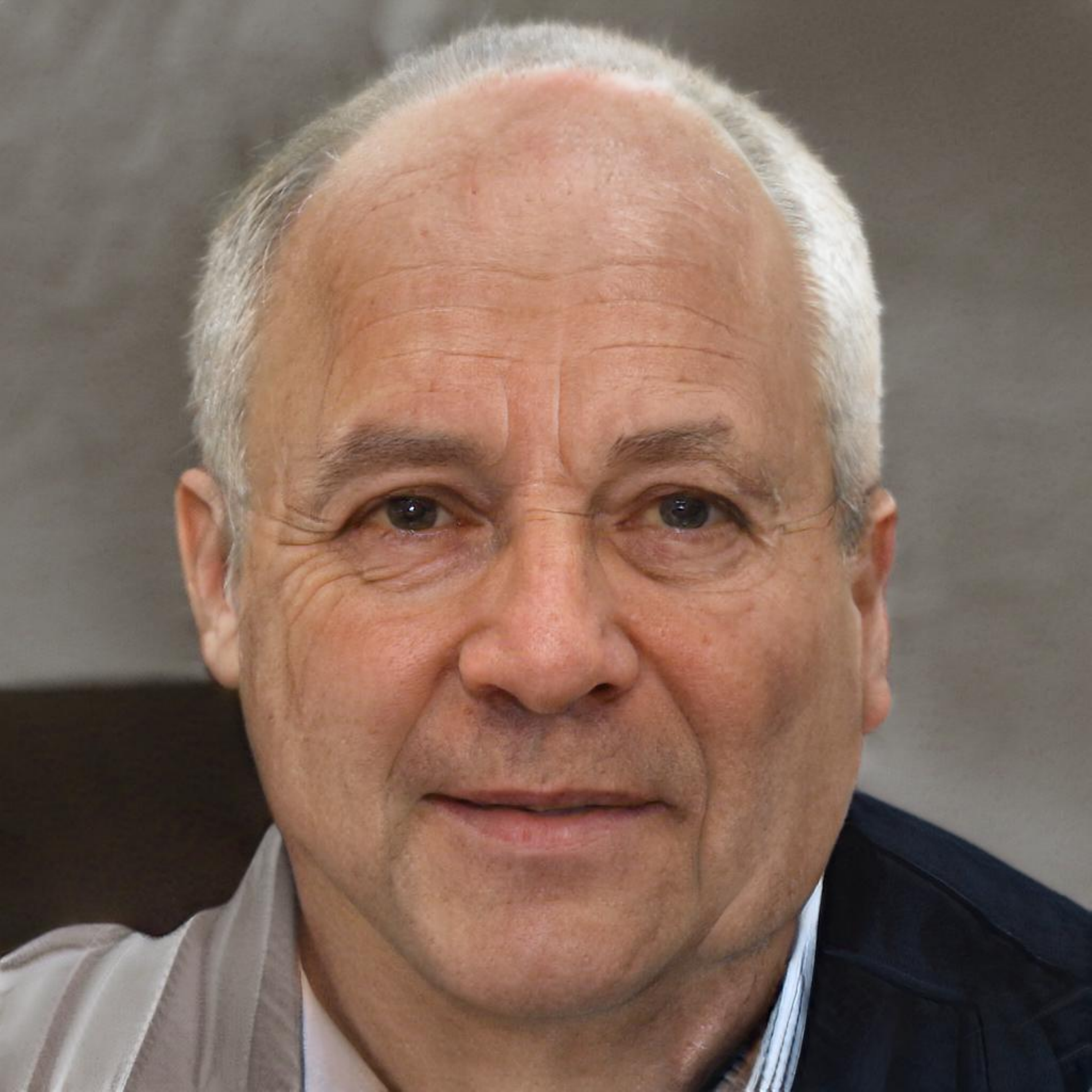}
\caption{\small FFHQ (1024x1024).
}
\end{subfigure}
\caption{\small Analogous plot to Fig.~\ref{fig:celebahq_sngan_vis}: additional high-resolution samples. }\label{fig:image_highresolution_appendix}
\end{figure}

\clearpage
\section{Architecture summary}

\begin{table}[!ht]
\centering
\caption{DCGAN architecture for the CIFAR-10 dataset. }
\label{tab:architecture-dcgan-cifar10}
\begin{subtable}[t]{.45\textwidth}
\caption{Generator $G_{\thetav}$}
\centering
\begin{tabular}{c}
\toprule 
$\epsilonv \in \mathbb{R}^{128} \sim \mathcal{N}(0,1)$ \\ \cmidrule(lr){1-1}
$128 \rightarrow 4\times4\times512$, dense, linear \\ \cmidrule(lr){1-1}
$4 \times 4$, stride=2 deconv. BN 256 ReLU \\ \cmidrule(lr){1-1}
$4 \times 4$, stride=2 deconv. BN 128 ReLU \\ \cmidrule(lr){1-1}
$4 \times 4$, stride=2 deconv. BN 64 ReLU \\ \cmidrule(lr){1-1}
$3 \times 3$, stride=1 conv. 3 Tanh \\
\bottomrule
\end{tabular}
\end{subtable}
\begin{subtable}[t]{.45\textwidth}
\caption{Feature encoder $\mathcal{T}_{\etav}$}
\centering
\begin{tabular}{c}
\toprule 
$\xv \in {[-1,1]}^{32\times 32 \times3}$ \\ \cmidrule(lr){1-1}
$3\times3$, stride=1 conv 64 lReLU \\
$4\times4$, stride=2 conv 64 lReLU \\ \cmidrule(lr){1-1}
$3\times3$, stride=1 conv 128 lReLU \\
$4\times4$, stride=2 conv 128 lReLU \\ \cmidrule(lr){1-1}
$3\times3$, stride=1 conv 256 lReLU \\
$4\times4$, stride=2 conv 256 lReLU \\ \cmidrule(lr){1-1} 
$3\times3$, stride=1 conv. 512 lReLU \\ \cmidrule(lr){1-1}
$ h \times w \times 512 \rightarrow m$, dense, linear \\ 
\bottomrule
\end{tabular}
\end{subtable}
\end{table}

\begin{table}[!ht]
\centering
\caption{DCGAN architecture for the CelebA and LSUN datasets. }
\label{tab:architecture-dcgan-celeba-lsun}
\begin{subtable}[t]{.45\textwidth}
\caption{Generator $G_{\thetav}$}
\centering
\begin{tabular}{c}
\toprule 
$\epsilonv \in \mathbb{R}^{128} \sim \mathcal{N}(0,1)$ \\ \cmidrule(lr){1-1}
$128 \rightarrow 4\times4\times1024$, dense, linear \\ \cmidrule(lr){1-1}
$4 \times 4$, stride=2 deconv. BN 512 ReLU \\ \cmidrule(lr){1-1}
$4 \times 4$, stride=2 deconv. BN 256 ReLU \\ \cmidrule(lr){1-1}
$4 \times 4$, stride=2 deconv. BN 128 ReLU \\ \cmidrule(lr){1-1}
$4 \times 4$, stride=2 deconv. BN 64 ReLU \\ \cmidrule(lr){1-1}
$3 \times 3$, stride=1 conv. 3 Tanh \\
\bottomrule
\end{tabular}
\end{subtable}
\begin{subtable}[t]{.45\textwidth}
\caption{Feature encoder $\mathcal{T}_{\etav}$}
\centering
\begin{tabular}{c}
\toprule 
$\xv \in {[-1,1]}^{64\times 64 \times3}$ \\ \cmidrule(lr){1-1}
$4\times 4$, stride=2 conv 64 lReLU \\
$4\times 4$, stride=2 conv BN 128 lReLU \\ \cmidrule(lr){1-1}
$4\times 4$, stride=2 conv BN 256 lReLU \\ \cmidrule(lr){1-1} 
$3\times 3$, stride=1 conv BN 512 lReLU \\ \cmidrule(lr){1-1}
$ h \times w \times 512 \rightarrow m$, dense, linear, Normalize \\ 
\bottomrule
\end{tabular}
\end{subtable}
\end{table}

\begin{table}[!ht]
\centering
\caption{ResNet architecture for the CIFAR-10 dataset. }
\label{tab:architecture-sngan-cifar10}
\begin{subtable}[t]{.45\textwidth}
\caption{Generator $G_{\thetav}$}
\centering
\begin{tabular}{c}
\toprule 
$\epsilonv \in \mathbb{R}^{128} \sim \mathcal{N}(0,1)$ \\ \cmidrule(lr){1-1}
$128 \rightarrow 4\times4\times 256$, dense, linear \\ \cmidrule(lr){1-1}
ResBlock up 256 \\ \cmidrule(lr){1-1}
ResBlock up 256 \\ \cmidrule(lr){1-1}
ResBlock up 256 \\ \cmidrule(lr){1-1}
BN, ReLU, $3\times3$ conv, 3 Tanh \\
\bottomrule
\end{tabular}
\end{subtable}
\begin{subtable}[t]{.45\textwidth}
\caption{Feature encoder $\mathcal{T}_{\etav}$}
\centering
\begin{tabular}{c}
\toprule 
$\xv \in {[-1,1]}^{32\times 32 \times3}$ \\ \cmidrule(lr){1-1}
ResBlock down 128 \\ \cmidrule(lr){1-1}
ResBlock down 128 \\ \cmidrule(lr){1-1}
ResBlock 128 \\ \cmidrule(lr){1-1}
ResBlock 128 \\ \cmidrule(lr){1-1}
ReLU \\ \cmidrule(lr){1-1}
Global sum pooling \\ \cmidrule(lr){1-1} 
$ h=128 \rightarrow m$, dense, linear, Normalize \\ 
\bottomrule
\end{tabular}
\end{subtable}
\end{table}

\begin{table}[!ht]
\centering
\caption{ResNet architecture for the CelebA and LSUN datasets. }
\label{tab:architecture-sngan-celeba-lsun}
\begin{subtable}[t]{.45\textwidth}
\caption{Generator $G_{\thetav}$}
\centering
\begin{tabular}{c}
\toprule 
$\epsilonv \in \mathbb{R}^{128} \sim \mathcal{N}(0,1)$ \\ \cmidrule(lr){1-1}
$128 \rightarrow 4\times4\times 1024$, dense, linear \\ \cmidrule(lr){1-1}
ResBlock up 512 \\ \cmidrule(lr){1-1}
ResBlock up 256 \\ \cmidrule(lr){1-1}
ResBlock up 128 \\ \cmidrule(lr){1-1}
ResBlock up 64 \\ \cmidrule(lr){1-1}
BN, ReLU, $3\times3$ conv, 3 Tanh \\
\bottomrule
\end{tabular}
\end{subtable}
\begin{subtable}[t]{.45\textwidth}
\caption{Feature encoder $\mathcal{T}_{\etav}$}
\centering
\begin{tabular}{c}
\toprule 
$\xv \in {[-1,1]}^{64\times 64 \times3}$ \\ \cmidrule(lr){1-1}
ResBlock down 128 \\ \cmidrule(lr){1-1}
ResBlock down 256 \\ \cmidrule(lr){1-1}
ResBlock down 512 \\ \cmidrule(lr){1-1}
ResBlock down 1024 \\ \cmidrule(lr){1-1}
ReLU \\
Global sum pooling \\ \cmidrule(lr){1-1} 
$ h=1024 \rightarrow m$, dense, linear, Normalize \\ 
\bottomrule
\end{tabular}
\end{subtable}
\end{table}

\begin{table}[!ht]
\centering
\caption{ResNet architecture for the LSUN-128 dataset. }
\label{tab:architecture-sngan-lsun-128}
\begin{subtable}[t]{.45\textwidth}
\caption{Generator $G_{\thetav}$}
\centering
\begin{tabular}{c}
\toprule 
$\epsilonv \in \mathbb{R}^{128} \sim \mathcal{N}(0,1)$ \\ \cmidrule(lr){1-1}
$128 \rightarrow 4\times4\times 1024$, dense, linear \\ \cmidrule(lr){1-1}
ResBlock up 1024 \\ \cmidrule(lr){1-1}
ResBlock up 512 \\ \cmidrule(lr){1-1}
ResBlock up 256 \\ \cmidrule(lr){1-1}
ResBlock up 128 \\ \cmidrule(lr){1-1}
ResBlock up 64 \\ \cmidrule(lr){1-1}
BN, ReLU, $3\times3$ conv, 3 Tanh \\
\bottomrule
\end{tabular}
\end{subtable}
\begin{subtable}[t]{.45\textwidth}
\caption{Feature encoder $\mathcal{T}_{\etav}$}
\centering
\begin{tabular}{c}
\toprule 
$\xv \in {[-1,1]}^{128\times 128 \times3}$ \\ \cmidrule(lr){1-1}
ResBlock down 128 \\ \cmidrule(lr){1-1}
ResBlock down 256 \\ \cmidrule(lr){1-1}
ResBlock down 512 \\ \cmidrule(lr){1-1}
ResBlock down 1024 \\ \cmidrule(lr){1-1}
ResBlock 1024 \\ \cmidrule(lr){1-1}
ReLU \\
Global sum pooling \\ \cmidrule(lr){1-1} 
$ h=1024 \rightarrow m$, dense, linear, Normalize \\ 
\bottomrule
\end{tabular}
\end{subtable}
\end{table}

\begin{table}[!ht]
\centering
\caption{ResNet architecture for the CelebA-HQ dataset. }
\label{tab:architecture-sngan-celeba-HQ}
\begin{subtable}[t]{.45\textwidth}
\caption{Generator $G_{\thetav}$}
\centering
\begin{tabular}{c}
\toprule 
$\epsilonv \in \mathbb{R}^{128} \sim \mathcal{N}(0,1)$ \\ \cmidrule(lr){1-1}
$128 \rightarrow 4\times4\times 1024$, dense, linear \\ \cmidrule(lr){1-1}
ResBlock up 1024 \\ \cmidrule(lr){1-1}
ResBlock up 512 \\ \cmidrule(lr){1-1}
ResBlock up 512 \\ \cmidrule(lr){1-1}
ResBlock up 256 \\ \cmidrule(lr){1-1}
ResBlock up 128 \\ \cmidrule(lr){1-1}
ResBlock up 64 \\ \cmidrule(lr){1-1}
BN, ReLU, $3\times3$ conv, 3 Tanh \\
\bottomrule
\end{tabular}
\end{subtable}
\begin{subtable}[t]{.45\textwidth}
\caption{Feature encoder $\mathcal{T}_{\etav}$}
\centering
\begin{tabular}{c}
\toprule 
$\xv \in {[-1,1]}^{256\times 256 \times3}$ \\ \cmidrule(lr){1-1}
ResBlock down 128 \\ \cmidrule(lr){1-1}
ResBlock down 256 \\ \cmidrule(lr){1-1}
ResBlock down 512 \\ \cmidrule(lr){1-1}
ResBlock down 512 \\ \cmidrule(lr){1-1}
ResBlock down 1024 \\ \cmidrule(lr){1-1}
ResBlock 1024 \\ \cmidrule(lr){1-1}
ReLU \\
Global sum pooling \\ \cmidrule(lr){1-1} 
$ h=1024 \rightarrow m$, dense, linear, Normalize \\ 
\bottomrule
\end{tabular}
\end{subtable}
\end{table}

\end{document}